\documentclass[10pt]{article}

\usepackage[T1]{fontenc}
\usepackage[utf8]{inputenc}
\usepackage{lmodern}        %
\usepackage{iftex}          %
\ifPDFTeX                   %
  \usepackage[activate={true,nocompatibility},final]{microtype}
\else                       %
  \usepackage[protrusion=true,final]{microtype}
\fi

\usepackage[margin=1in]{geometry}

\usepackage{authblk}                    %

\usepackage{amsmath}
\usepackage{amssymb,amsfonts}
\usepackage{mathtools}      %
\usepackage{bm}             %
\usepackage{nicefrac}
\allowdisplaybreaks         %

\usepackage{amsthm}

\usepackage{graphicx}
\graphicspath{{figures/}{./}}                       %
\usepackage{booktabs}                               %
\usepackage{multirow,makecell,array}
\usepackage[font=small,labelfont=bf]{caption}       %
\usepackage{subcaption}                             %
\usepackage{algorithm}
\usepackage{algpseudocode}                          %

\usepackage{enumitem}
\setlist[itemize]{leftmargin=2.2em,itemsep=2pt,topsep=2pt}
\setlist[enumerate]{leftmargin=2.2em,itemsep=2pt,topsep=2pt}

\usepackage{xcolor}
\definecolor{LinkColor}{rgb}{0.10,0.40,0.75}        %
\definecolor{CiteColor}{rgb}{0.70,0.25,0.20}        %
\definecolor{UrlColor} {rgb}{0.20,0.50,0.50}        %
\definecolor{TodoColor}{rgb}{0.80,0.30,0.10}        %

\usepackage{tikz}
\usetikzlibrary{positioning,calc,arrows.meta}
\definecolor{FigBlue}{RGB}{47,93,158}     %
\definecolor{FigOrange}{RGB}{217,130,43}
\definecolor{FigGreen}{RGB}{58,143,92}
\definecolor{FigRed}{RGB}{181,71,58}
\definecolor{FigGray}{RGB}{120,120,120}

\usepackage[round,sort&compress]{natbib}

\usepackage{url}
\usepackage{hyperref}
\hypersetup{
  colorlinks=true,
  linkcolor=LinkColor,
  citecolor=CiteColor,
  urlcolor=UrlColor,
  breaklinks=true,
  bookmarksnumbered=true,
}
\usepackage{bookmark}                               %

\numberwithin{equation}{section}

\newif\ifdraft \draftfalse
\ifdraft
  \usepackage{lineno}\linenumbers
  \newcommand{\todo}[1]{\textcolor{TodoColor}{\textbf{[TODO:}~#1\textbf{]}}}
\else
  \newcommand{\todo}[1]{}                            %
\fi
\newcommand{\R}{\mathbb{R}}
\newcommand{\N}{\mathbb{N}}
\newcommand{\Z}{\mathbb{Z}}
\newcommand{\Q}{\mathbb{Q}}
\newcommand{\E}{\mathbb{E}}                       %
\let\C\relax \newcommand{\C}{\mathbb{C}}          %
\renewcommand{\P}{\mathbb{P}}                     %

\newcommand{\eps}{\varepsilon}

\newcommand{\defeq}{\coloneqq}                    %

\DeclarePairedDelimiterX{\inner}[2]{\langle}{\rangle}{#1,#2}   %

\DeclareMathOperator{\Sym}{Sym}
\DeclareMathOperator{\Fix}{Fix}
\DeclareMathOperator{\Gal}{Gal}
\DeclareMathOperator{\cl}{cl}
\DeclareMathOperator{\dimDS}{DS}
\DeclareMathOperator{\dimN}{Nat}
\DeclareMathOperator{\Ldim}{Ldim}
\DeclareMathOperator{\TD}{TD}
\DeclareMathOperator{\er}{er}
\newcommand{\Hc}{\mathcal{H}}                     %
\newcommand{\HG}[2]{\mathcal{H}_{#1,#2}}          %
\newcommand{\hgt}{\mathsf{H}}                     %
\newcommand{\irr}{\mathsf{I}}                     %
\newcommand{\bsz}{\mathsf{b}}                     %
\newcommand{\Bsz}{\mathsf{B}}                     %
\newcommand{\stab}[2]{#1_{(#2)}}                  %
\newcommand{\VS}{V}                               %
\newcommand{\Wd}{W_d}                             %
\theoremstyle{plain}
\newtheorem{theorem}{Theorem}[section]

\newtheorem{lemma}[theorem]{Lemma}
\newtheorem{corollary}[theorem]{Corollary}
\newtheorem{conjecture}[theorem]{Conjecture}
\newtheorem{fact}[theorem]{Fact}

\theoremstyle{definition}
\newtheorem{definition}[theorem]{Definition}

\newtheorem{example}[theorem]{Example}

\theoremstyle{remark}
\newtheorem{remark}[theorem]{Remark}

\usepackage{thm-restate}  %
\usepackage[capitalise,nameinlink,noabbrev,sort&compress]{cleveref}  %
\crefname{theorem}{Theorem}{Theorems}          \Crefname{theorem}{Theorem}{Theorems}
\crefname{proposition}{Proposition}{Propositions}
\Crefname{proposition}{Proposition}{Propositions}
\crefname{lemma}{Lemma}{Lemmas}                \Crefname{lemma}{Lemma}{Lemmas}
\crefname{corollary}{Corollary}{Corollaries}   \Crefname{corollary}{Corollary}{Corollaries}
\crefname{conjecture}{Conjecture}{Conjectures} \Crefname{conjecture}{Conjecture}{Conjectures}
\crefname{fact}{Fact}{Facts}                   \Crefname{fact}{Fact}{Facts}
\crefname{definition}{Definition}{Definitions} \Crefname{definition}{Definition}{Definitions}
\crefname{assumption}{Assumption}{Assumptions} \Crefname{assumption}{Assumption}{Assumptions}
\crefname{example}{Example}{Examples}          \Crefname{example}{Example}{Examples}
\crefname{problem}{Problem}{Problems}          \Crefname{problem}{Problem}{Problems}
\crefname{remark}{Remark}{Remarks}             \Crefname{remark}{Remark}{Remarks}
\crefname{claim}{Claim}{Claims}                \Crefname{claim}{Claim}{Claims}
\crefname{algorithm}{Algorithm}{Algorithms}    \Crefname{algorithm}{Algorithm}{Algorithms}

\usepackage{etoolbox}
\makeatletter
\newcommand{\onepagetoc}{%
  \patchcmd{\l@section}{\addvspace{1.0em \@plus\p@}}{\addvspace{0.3em \@plus 1fill}}{}%
    {\PackageWarningNoLine{main}{TOC section-spacing patch failed}}%
  \patchcmd{\@dottedtocline}{\vskip \z@ \@plus.2\p@}{\vskip \z@ \@plus .3fill}{}%
    {\PackageWarningNoLine{main}{TOC subsection-spacing patch failed}}%
  \flushbottom}
\makeatother

\newcommand{\creftypefor}[1]{\AtBeginEnvironment{#1}{\crefalias{theorem}{#1}}}
\creftypefor{proposition}\creftypefor{lemma}\creftypefor{corollary}\creftypefor{conjecture}
\creftypefor{fact}\creftypefor{definition}\creftypefor{assumption}\creftypefor{example}
\creftypefor{problem}\creftypefor{remark}\creftypefor{claim}

\title{\fontsize{15.5}{19}\selectfont Learning a Group Element from Its Action: Galois Correspondences, Optimal ERM, and an Abel--Ruffini Threshold for the Natarajan Dimension}
\author{Xiaoyu Li}
\affil{University of New South Wales\\ \texttt{xiaoyulics@gmail.com}}
\date{}

\hypersetup{pdftitle={Learning a Group Element from Its Action: Galois Correspondences, Optimal ERM, and an Abel--Ruffini Threshold for the Natarajan Dimension},pdfauthor={Xiaoyu Li}}

\begin{document}
\maketitle

\begin{abstract}
We study learning an unknown element $g$ of a group $G$ acting with finite orbits on a countable set $\Omega$, from examples $(x,gx)$.
After a sample on a set $A$ of points, the consistent hypotheses are the restrictions of a coset of the pointwise stabilizer $\stab{G}{A}$, and the sample forces exactly the labels of the points that $\stab{G}{A}$ fixes; for a Galois group, these points form the field generated by the data.
Through this Galois correspondence, the multiclass dimensions become invariants of permutation groups.
The DS and graph dimensions are the height $\hgt$, the largest size of a set of points each of which can be moved while all the others stay fixed; the Littlestone and teaching dimensions are the length of the longest irredundant sequence and the minimal base size, two classical notions for permutation groups; and the list DS dimension is a list version of the height.
In the realizable model, every consistent learner, however it breaks ties, has expected error at most $\hgt/(m+1)$ after $m$ examples, within a factor $16$ of the minimax error when $2\le\hgt<\infty$ and $m\ge\max(2,\hgt-1)$.
On nondegenerate classes it attains the optimal deterministic sample complexity $\Theta((\hgt+\log(1/\delta))/\eps)$, and every consistent online learner attains the optimal deterministic mistake bound.
In the agnostic model, every ERM uses $O((\hgt+\log(1/\delta))/\eps^2)$ examples, at most a factor $O(\dimDS/\dimN)$, the ratio of the DS and Natarajan dimensions, above the known lower bound for deterministic learners.

Our main new results concern this ratio for Galois groups.
For the Galois group of a finite Galois extension of a countable field, the Natarajan dimension equals a group invariant, the coset-cube number: the largest $t$ for which some subgroups $V_1,\dots,V_t$ and cosets $k_iV_i\neq V_i$ have a common element for each of the $2^t$ ways of choosing $V_i$ or $k_iV_i$.
It is at least the number of complemented factors of any chief series of any subgroup.
Consequently the two dimensions coincide for every solvable Galois group, while for the groups $A_5$ and $S_5$ of the quintic they are $2<3$ and $3<4$, and $A_5$ is the smallest group with a gap.
For the Galois group $S_n$ of the general equation of degree $n$, the two dimensions coincide if and only if $n\le4$, that is, in characteristic zero, exactly when the equation is solvable by radicals.
Our proof of the gap for $n\ge5$ uses the independent generating sets of $S_n$, not its non-solvability, and for arbitrary groups the implication is one-way: the simple group $\mathrm{PSL}(2,8)$ has no gap.
For the generic quartic, a root, a root of the resolvent cubic and the square root of the discriminant, three quantities from the classical solution by radicals, form a Natarajan-shattered set.

We also show that for the Galois group of a finite Galois extension $L/K$ of a countable field with $L\neq K$ the teaching dimension is $1$, and the DS and Littlestone dimensions are the largest size of an independent subset and the length of the longest subgroup chain of the group; that for linear, orthogonal and unitary maps of $\mathbb R^d$ or $\mathbb C^d$ the Natarajan and DS dimensions both equal $d$; and that a two-fibered pseudo-cube, one in which every vector has a unique neighbor in each coordinate, is a disjoint union of Natarajan cubes exactly when the group generated by its neighbor maps is abelian.
For a permutation group of a finite set, learned on the whole set, the Natarajan dimension is at least half the DS dimension when the closure that sends a sample to its forced points is a matroid closure; we conjecture this bound for every such group.

\end{abstract}

\clearpage
{\small\onepagetoc\setcounter{tocdepth}{2}\tableofcontents\clearpage}

\section{Introduction}
\label{sec:intro}

Galois theory answers a question about information.
Given the coefficients of a polynomial, which quantities can be computed from them?
Galois's answer is that the computable quantities are exactly those fixed by the symmetries of the roots that the coefficients cannot see.
Learning theory asks a question of the same kind.
Given a labeled sample, which labels does it determine?
For a hypothesis class this is the region on which all hypotheses consistent with the sample agree.
In general that region has no structure we can name.
In this paper we identify a natural learning problem in which it does: the region is exactly the set fixed by a group of symmetries, and the resulting Galois correspondence controls the main dimensions of the learning problem.

Two features of multiclass learning theory have no structural explanation in general.
The first is the zoo of combinatorial dimensions (Natarajan, graph, DS, list-DS, Littlestone), each tied to a different algorithmic paradigm.
The second is the gap between the Natarajan and DS dimensions: the classes of \citet{brukhim2022characterization} with Natarajan dimension~$1$ and arbitrarily large DS dimension come from sophisticated geometric constructions.
For learning a symmetry, both acquire an algebraic description.
The DS, Littlestone and teaching dimensions become invariants of a single closure system, read through independent sets (DS), chains (Littlestone) and generating sets (teaching).
A Natarajan--DS gap needs symmetries that fail to commute (\cref{thm:natarajan}).
For Galois groups, chief factors, the structure behind solution by radicals, build Natarajan cubes when they are complemented: solvable groups have enough of them to close the gap, and for the general equation of degree~$n$ the gap opens exactly when, in characteristic zero, the equation stops being solvable by radicals.

\paragraph{Learning a symmetry.}
Let a group $G$ act on a set $\Omega$, and fix a domain $X\subseteq\Omega$.
A learner receives examples $(x,gx)$ for an unknown $g\in G$ and must predict $gx$ on new points.
Equivalently, the learner faces the multiclass hypothesis class
\[
  \HG{G}{X}=\{\,x\mapsto gx \;:\; g\in G\,\}\subseteq\Omega^X .
\]
Many problems have this form.
Learning a permutation from a partial matching is the case $G=S_n$.
Learning an invertible linear map from point correspondences is the case $G=\mathrm{GL}(V)$ acting on $V$.
Registration problems (projective, affine or rigid transformations) and learning an automorphism of a graph or of another structure are further cases.
The most classical instance is a field extension $L/K$ with Galois group $\Gamma$: learn an automorphism $\sigma\in\Gamma$ from pairs $(\alpha,\sigma\alpha)$.
For finite fields this means learning a power of the Frobenius map.

\begin{figure}[t]
\centering
\captionsetup{font=footnotesize}
\includegraphics[width=\textwidth]{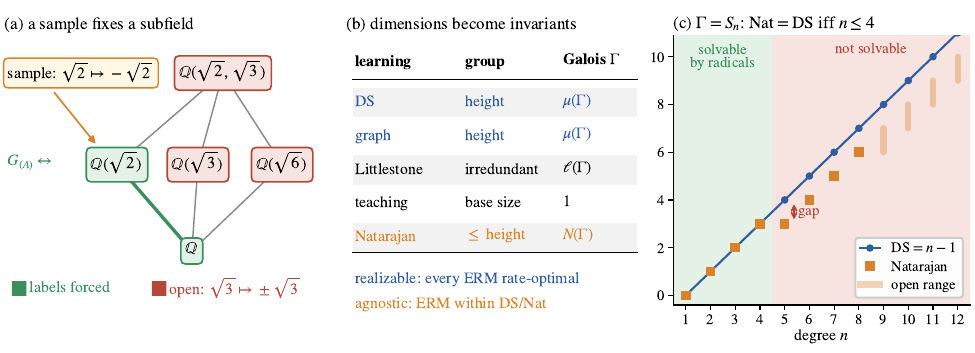}
\caption{Overview.
(a)~Learning an automorphism $\sigma$ of $L=\Q(\sqrt2,\sqrt3)$ over $\Q$ from examples $(\alpha,\sigma\alpha)$.
After the example $\sqrt2\mapsto-\sqrt2$, the consistent hypotheses form a coset of the pointwise stabilizer $G_{(A)}=\Gal(L/\Q(\sqrt2))$ of the sample set $A=\{\sqrt2\}$; the forced labels are exactly those of the elements of its fixed field $\Q(\sqrt2)$ (green), for instance $1+\sqrt2\mapsto1-\sqrt2$, while $\sqrt3$ keeps the $[\Q(\sqrt2)(\sqrt3):\Q(\sqrt2)]=2$ candidate labels $\pm\sqrt3$ (red; \cref{lem:coset,thm:galois}).
(b)~Learning dimensions as invariants of a group class with finite orbits and of the Galois group $\Gamma$ of a finite Galois extension $L/K$ of a countable field with $L\neq K$ (\cref{thm:dictionary,thm:galois,thm:natarajan,thm:cosetcube,prop:agnostic}).
The height, the irredundant length and the base size are defined in \cref{def:invariants}; $\mu(\Gamma)$ is the largest size of an independent subset of $\Gamma$, $\ell(\Gamma)$ the length of a longest subgroup chain, and $N(\Gamma)$ the coset-cube number (\cref{sec:fields,sec:solvable}).
On nondegenerate classes (some element fixes one domain point and moves another) every ERM is rate-optimal in the realizable model among deterministic learners (\cref{thm:erm}); its agnostic sample complexity is within a factor $O(\dimDS/\dimN)$, the ratio of the DS and Natarajan dimensions, of the lower bound of \citet{daniely2015multiclass} (\cref{prop:agnostic}).
(c)~For the Galois group $S_n$ of the general equation of degree~$n$ over a countable field, the DS dimension is $n-1$, and the Natarajan dimension equals it exactly when $n\le4$; in characteristic zero these are the degrees in which the equation is solvable by radicals (\cref{thm:solvable,thm:abelruffini}).
Squares are proved values; for $9\le n\le12$ the bars show the open range from $\lfloor3n/4\rfloor$ to $n-2$.}
\label{fig:hero}
\end{figure}

\paragraph{The Galois principle for version spaces.}
Our starting point is a simple observation (\cref{lem:coset}).
After the learner sees the labels $gx$ for $x$ in a finite set $A$, the hypotheses consistent with the sample are the restrictions of the coset $g\,\stab{G}{A}$, where $\stab{G}{A}$ is the pointwise stabilizer of $A$.
The sample therefore forces the label of a point $x$ exactly when every element of $\stab{G}{A}$ fixes $x$, that is, when
\[
  x\in\cl(A)\defeq\Fix\big(\stab{G}{A}\big),
\]
the set of points fixed by every element of $\stab{G}{A}$.
The map $A\mapsto\cl(A)$ is the closure operator of the Galois connection between subsets of $\Omega$ and subgroups of $G$.
This closure system is classical in permutation group theory, where it underlies the notions of bases and irredundant bases \citep{cameron1995bases}, which we recall below; our contribution is its learning-theoretic reading.
When $G=\Gal(L/K)$ acts on $L$, the fundamental theorem of Galois theory identifies $\cl(A)$ with the field $K(A)$ generated by the data.
In short: \emph{what a sample determines is exactly what its symmetries fix} (\cref{fig:hero}).
This paper develops the learning theory that follows from this principle.

\subsection{Our results}

Throughout, $G$ acts on a countable set $\Omega$, the domain is $X\subseteq\Omega$, and we assume that every orbit $Gx$ ($x\in X$) is finite (condition~\eqref{eq:F} of \cref{sec:prelim}).
This holds for every finite group and also for profinite Galois groups acting on separable closures of countable fields.
Linear, orthogonal and unitary maps of $\R^d$ and $\C^d$ fall outside these assumptions; for them we prove the dimension identities and the expected-error bounds separately (\cref{prop:linear}), and the other continuous cases remain open (\cref{sec:discussion}).
The relevant group invariants are the following.
A set $A\subseteq X$ is \emph{independent} if no point of $A$ lies in the closure of the others.
The \emph{height} $\hgt=\hgt_X(G)$ is the largest size of an independent set.
The quantity $\irr=\irr_X(G)$ is the largest length of an \emph{irredundant} sequence, that is, a sequence in which no point lies in the closure of the earlier ones.
The quantity $\bsz=\bsz_X(G)$ is the least size of a \emph{relative base}, that is, a set whose closure contains $X$.
For $X=\Omega$ these are the height, the maximal irredundant base length and the minimal base size of the permutation group, which are well studied in group theory \citep{cameron1995bases,gill2021height,kelsey2021relational}.

\paragraph{1. A dictionary between learning dimensions and Galois invariants (\cref{thm:dictionary}).}
For every group class, with $\dimDS$, $\Ldim$ and $\TD$ the DS, Littlestone and teaching dimensions (\cref{sec:prelim-learning}),
\[
  \dimDS(\HG{G}{X})=\hgt_X(G),\qquad
  \Ldim(\HG{G}{X})=\irr_X(G),\qquad
  \TD(h)=\bsz_X(G)\ \ \text{for every } h\in\HG{G}{X}.
\]
The list DS dimension of \citet{charikar2023list} becomes the $k$-height $\hgt^k_X(G)$ of \cref{def:invariants}, a list height defined by orbits of size larger than the list size~$k$.
\Cref{tab:dictionary} lists all of these correspondences.
So the classical chain $\bsz\le\hgt\le\irr$ of permutation group theory becomes, when $\irr<\infty$, the chain $\TD\le\dimDS\le\Ldim$ of learning-theoretic dimensions (\cref{rem:dictionary}).
Results proved about bases in group theory then transfer to learning.
For instance, by a theorem of \citet[Theorem~1.1]{kelsey2021relational} and \cref{thm:dictionary}, an element of a primitive group of degree $n$ that is not of ``large-base'' type (subgroups of the wreath products $S_m\wr S_r$ in product action that contain $A_m^r$, with $S_m$ acting on $k$-subsets of $[m]$; \cref{rem:transfer}) can be learned online with fewer than $5\log_2 n$ mistakes.

\paragraph{2. Every ERM is rate-optimal (\cref{thm:erm}).}
In general multiclass learning, empirical risk minimization can fail: optimal learners must sometimes be improper \citep{daniely2014optimal}.
Even the optimal sample complexity $\Theta((\dimDS+\log(1/\delta))/\eps)$ was settled only recently, and it is achieved by aggregating one-inclusion graph predictors \citep{brukhim2022characterization,pabbaraju2026optimal}.
For group classes the picture is much simpler.
\emph{Every} consistent learner, with arbitrary or even adversarial tie-breaking, has expected error at most $\hgt/(m+1)$.
The proof uses the Galois structure directly: a consistent hypothesis is correct on the closure of the sample, the sample points that do not lie in the closure of the other points form an independent set, and hence at most $\hgt$ of $m+1$ exchangeable points can be predicted wrongly in a leave-one-out experiment.
For $2\le\hgt<\infty$ and $m\ge\max(2,\hgt-1)$, no learner, randomized or not, can do better than $(\hgt-1)/(8m)$ in the worst case, so this bound is optimal up to a factor~$16$ (\cref{prop:expected-lower}).
With probability $1-\delta$ the error is $O((\hgt+\log(1/\delta))/m)$, and on every nondegenerate class (one in which the stabilizer of some point of $X$ moves some point of $X$) this rate is optimal among deterministic learners.
Every consistent online learner makes at most $\irr$ mistakes, which is optimal among deterministic online learners.
In the agnostic model, the graph dimension of a group class equals its height, so every ERM uses $O((\hgt+\log(1/\delta))/\eps^2)$ examples with no assumption on the labels (\cref{prop:agnostic}).

\paragraph{3. Galois groups (\cref{thm:galois}).}
Let $L/K$ be a finite Galois extension of a countable field with group $\Gamma\neq1$; we call $\HG{\Gamma}{L}$ its \emph{Galois class}.
By the primitive element theorem, every subgroup of $\Gamma$ is the stabilizer of some element of $L$.
Consequently the learning dimensions of the Galois class are invariants of the abstract group $\Gamma$.
The teaching dimension is $1$: a primitive element is a teaching set.
The DS dimension equals $\mu(\Gamma)$, the largest size of an \emph{independent} subset of $\Gamma$, that is, a subset none of whose elements lies in the subgroup generated by the others.
The Littlestone dimension equals the length $\ell(\Gamma)$, counted in strict inclusions, of a longest chain of subgroups.
After a sample on $A$, the number of labels consistent with the data at $\alpha$ equals the degree $[K(A)(\alpha):K(A)]$.
In other words, the number of candidate labels is a field degree, and the label of $\alpha$ is forced exactly when $\alpha\in K(A)$ (\cref{fig:hero}(a)).
For the generic case $\Gamma\cong S_n$, the theorems of \citet{whiston2000maximal} and \citet{cameron1989chains} give
\[
  \TD=1,\qquad \dimDS=n-1,\qquad \Ldim=\lceil 3n/2\rceil-b(n)-1,
\]
where $b(n)$ is the number of ones in the binary expansion of $n$.

\paragraph{4. Commuting symmetries give Natarajan shattering (\cref{thm:natarajan}).}
The distinction between the Natarajan and DS dimensions is central to multiclass learning theory.
The Natarajan dimension was long the leading candidate for characterizing multiclass learnability \citep{brukhim2022characterization}, and it governs the agnostic term of the sample complexity \citep{cohen2025agnostic,pabbaraju2026optimal}; \citet{brukhim2022characterization} showed, using hyperbolic pseudo-manifolds, that it does not characterize learnability in general.
For group classes the ratio $\dimDS/\dimN$ of the two dimensions also has a concrete meaning: the agnostic ERM bound of \cref{prop:agnostic} exceeds the lower bound of \citet{daniely2015multiclass} by at most this factor, up to absolute constants, and whether ERM actually pays it is open.
When is there a gap?
Natarajan shattering requires moves at different points that can be combined independently, where a \emph{move} at $a\in A$ is an element of $G$ that fixes the other points of $A$ and moves~$a$.
Commuting moves provide exactly this.
Hence $\dimN(\HG{G}{X})=\dimDS(\HG{G}{X})$ for abelian $G$, so a gap between the two dimensions requires non-commutativity.
In general, $\dimN$ is at least the height of every abelian subgroup (this lower bound can be strict).
The ratio $1/2$ occurs: $S_n$ acting on $n$ points has $\dimN=\lfloor n/2\rfloor$ and $\dimDS=n-1$.
We conjecture that $\dimN\ge\lceil\dimDS/2\rceil$ for every finite permutation group acting on its full domain $X=\Omega$ (\cref{conj:half}).
We prove it when the closure operator has the exchange property of a matroid closure (\cref{thm:matroid}), a class in which the constant $1/2$ is tight, and we support it by exhaustive computation.
The full-domain hypothesis is necessary: the coset complexes of \citet{brukhim2022characterization}, pseudo-cubes of the form $\{(gH_1,\dots,gH_d)\}$ (\cref{sec:natarajan}), are group classes restricted to a non-invariant set of points.

\paragraph{5. Solvability and an Abel--Ruffini threshold for the Natarajan dimension (\cref{thm:cosetcube,thm:solvable,thm:abelruffini}).}
For a Galois group the question becomes one about the abstract group, and we prove:
\begin{quote}
\emph{If $\Gamma$ is solvable, then $\dimN=\dimDS=\mu(\Gamma)$. If $\Gamma\cong S_n$, the Galois group of the general equation of degree~$n$, then $\dimN=\dimDS$ if and only if $n\le4$.}
\end{quote}
In characteristic zero the general equation of degree~$n$ is solvable by radicals exactly when $n\le4$, so its Galois class $\HG{S_n}{L}$ is \emph{Natarajan-saturated} (the two dimensions coincide) exactly when the equation is solvable by radicals.
The mechanism is a group invariant.
The Natarajan dimension of $\HG{\Gamma}{L}$ equals the \emph{coset-cube number} $N(\Gamma)$: the largest $t$ for which some subgroups $V_1,\dots,V_t$ and elements $k_i\notin V_i$ make all $2^t$ intersections $\bigcap_ik_i^{\varepsilon_i}V_i$ ($\varepsilon\in\{0,1\}^t$) nonempty, with $k^0V=V$ and $k^1V=kV$.
When $V_i$ is the stabilizer of a point $x_i$, the intersection indexed by $\varepsilon$ is the set of elements that send each $x_i$ to $x_i$ or to $k_ix_i$ as $\varepsilon$ prescribes, so a coset cube is a Natarajan cube in group-theoretic form; in a Galois class every subgroup is a stabilizer (\cref{fact:realization}), so every coset cube arises in this way (\cref{fig:cosetcube}).
A chief series of a group $\Delta$ is a maximal chain $1=\Delta_0<\dots<\Delta_r=\Delta$ of normal subgroups of $\Delta$, and its factor $\Delta_i/\Delta_{i-1}$ is complemented when $\Delta/\Delta_{i-1}$ splits as a semidirect product with normal factor $\Delta_i/\Delta_{i-1}$ (\cref{sec:solvable}).
Each complemented factor of a chief series contributes one dimension to a coset cube, through a semidirect-product lifting.
For the generic quartic the cube is visible in the classical solution by radicals: a root, a root of the resolvent cubic and the square root of the discriminant (\cref{ex:quartic}) form a Natarajan-shattered set (\cref{fig:quartic}).
Together with a bound on minimal generating sets of solvable groups (one half of a theorem of \citet{lucchini2013largest}), the chief-factor count proves the solvable case.
For $A_5$ the complemented chief factors of its subgroups fall short of $\mu(A_5)=3$, and a case analysis shows that $A_5$ has no coset cube of dimension~$3$ and $S_5$ none of dimension~$4$; $A_5$ is the smallest group with a gap (\cref{cor:a5smallest}).
For $n\ge5$ we get $\lfloor3n/4\rfloor\le\dimN\le n-2<n-1=\dimDS$, with $\dimN=n-2$ for $5\le n\le8$.
The upper bound is proved combinatorially, not from the non-solvability of $S_n$: it combines the classification of independent generating sets of $S_n$ of size $n-1$ \citep{whiston2000maximal,cameron2002independent} with an orthogonality relation that the cells of a coset cube impose on indicator vectors (\cref{fig:orthogonality}).
Indeed, for general groups the implication goes one way only: some non-solvable groups, such as $\mathrm{PSL}(2,8)$, have no gap (\cref{prop:psl28}).
The Natarajan dimension sees at least the complemented chief factors, solvable groups always have enough of them, and it can see more (\cref{prop:pgal}).

\paragraph{6. Monodromy of pseudo-cubes (\cref{thm:monodromy}).}
DS shattering is witnessed by pseudo-cubes \citep{daniely2014optimal,brukhim2022characterization}: finite sets of label vectors in which every vector has, in every coordinate, a neighbor that differs only there.
When these neighbors are unique, we call the pseudo-cube \emph{two-fibered}; the neighbor maps are then involutions, and they generate a \emph{monodromy group}~$M$.
We show that every orbit of $M$ on a two-fibered pseudo-cube is a coordinatewise image of the coset complex of its own monodromy group (the restriction of $M$ to the orbit) with respect to the parabolic subgroups (each generated by all involutions but one), and that, when $M$ acts regularly (transitively with trivial stabilizers), the orbit map is a bijection from that coset complex onto the orbit that preserves $i$-neighbors and is induced by coordinatewise maps of the labels (\cref{rem:regular}).
This is a converse to the coset-complex construction of \citet{brukhim2022characterization}: their pseudo-cubes are the Galois (regular) objects.
Moreover, a two-fibered pseudo-cube is a disjoint union of Natarajan cubes \emph{if and only if} its monodromy group is abelian; the $S_3$ hexagon of \cref{fig:hexagon} has monodromy group $S_3$ and contains no Natarajan square.
The neighbor graph of a two-fibered pseudo-cube is a thin chamber system, and the coset description is the pseudo-cube form of standard facts about chamber systems and abstract polytopes \citep{wilson2012maniplexes,calderbank2016parabolic,hartley1999all}; what is specific here is the coordinatewise labeling and the abelian criterion.

\subsection{Related work}

\paragraph{Multiclass learnability.}
\citet{daniely2015multiclass} related the ERM principle to the graph and Natarajan dimensions.
\citet{daniely2014optimal} introduced the DS dimension and showed that optimal learners must sometimes be improper.
\citet{brukhim2022characterization} proved that the DS dimension characterizes learnability.
Their proof that the Natarajan dimension does not characterize learnability builds pseudo-cubes from coset complexes of groups generated by involutions \citep{januszkiewicz2003hyperbolic}.
We use the same objects in the opposite direction: every two-fibered pseudo-cube arises, orbit by orbit, from its monodromy group, as thin chamber systems and polytopes do from theirs \citep{calderbank2016parabolic,hartley1999all,mcmullen2002abstract}.
List learning was characterized by \citet{charikar2023list}.
The dependence of the realizable, agnostic and list sample complexities on the DS dimension was sharpened by \citet{hanneke2024improved} and \citet{cohen2025agnostic}, and determined by \citet{pabbaraju2026optimal}.
Between the realizable and agnostic regimes, \citet{li2026optimistic} show that, over a finite label set, an improper compression-based learner attains excess risk $\widetilde O(\sqrt{L^\star\dimN/m}+\dimDS/m)$ over the best-in-class risk $L^\star$, with a matching lower bound, up to logarithmic factors, on a constructed class.
None of these works studies learning a group action as such, or relates the dimensions to permutation-group invariants.
Our optimality result for every ERM relies on the closure algorithm for intersection-closed classes \citep{auer2007new,darnstadt2015optimal}.
It is in the spirit of the closure and Helly-number analyses of proper learning \citep{bousquet2020proper}.
\citet{daniely2015multiclass} also bound the ERM sample complexity of classes that are symmetric under all permutations of the labels, with logarithmic factors.

\paragraph{Symmetry in learning.}
Invariance and equivariance have been studied mostly as a prior on the target function \citep{shao2022theory,elesedy2022group}, and in work on breaking symmetries with equivariant networks \citep{kaba2023symmetry}.
\citet{soleymani2026efficient} learn regression functions under exact invariances and recover an unknown subgroup of a known finite group from noisy labeled samples; there the group is a symmetry of the target. \citet{cipriani2025graphs} study PAC and online learnability of the copies of a countable graph under vertex permutations; there the hypotheses are edge indicators of copies, a binary class on pairs of vertices, and the target is a copy rather than a permutation.
Here the symmetry \emph{is} the target, observed through its action, and the questions concern multiclass learning dimensions.

\paragraph{Permutation groups.}
Bases, heights, irredundant bases and relational complexity are classical invariants.
They are studied by \citet{cameron1995bases}, \citet{gill2021height}, \citet{kelsey2021relational}, \citet{delvalle2025greedy} and \citet{duyan2018pyber}, among others.
Independent sets in groups and subgroup chains in symmetric groups were determined by \citet{whiston2000maximal}, \citet{cameron2002independent} and \citet{cameron1989chains}.
The Galois connection between subsets and pointwise stabilizers for actions with finite orbits, and its profinite completion, are studied by \citet{marmaridis2026finite}; we use only its finite, elementary part.
Our dictionary makes these results statements about learning.
Minimal generating sets of finite groups and chief factors are studied by \citet{lucchini2013largest} and \citet{lucchini2022independent}; families of subgroups in general position, by \citet{detomi2015general}; and the Helly dimension of coset systems, by \citet{domokos2005typical}.
The coset cubes of \cref{sec:solvable} add a sign-pattern condition that these notions do not impose.
In model theory and additive combinatorics, the VC dimension of the translates of a fixed subset of a group is a central notion \citep{conant2018structure}; there the family of sets is fixed, while coset cubes vary over subgroups.

\paragraph{Organization.}
\Cref{sec:prelim} sets up notation.
\Cref{sec:galois} develops the Galois correspondence and the dictionary.
\Cref{sec:erm} proves that every ERM is rate-optimal and that every consistent online learner is optimal among deterministic online learners.
\Cref{sec:fields} treats Galois groups.
\Cref{sec:natarajan} studies the Natarajan dimension of group classes, and \cref{sec:solvable} shows that solvable Galois groups have no Natarajan--DS gap, while the Galois group of the general equation of degree $n\ge5$ has one.
\Cref{sec:monodromy} studies the monodromy of pseudo-cubes.
\Cref{sec:discussion} discusses open problems.
\Cref{app:background} is a crash course on the Galois theory and group theory we use, \cref{app:external} restates the external results on which the proofs rely, and the remaining appendices contain the deferred proofs.

\section{Preliminaries}
\label{sec:prelim}

\subsection{Multiclass learning}
\label{sec:prelim-learning}

Let $X$ be a nonempty set (the domain) and $Y$ a set of labels.
A hypothesis class is a set $\Hc\subseteq Y^X$.
For a sequence $\mathbf{x}=(x_1,\dots,x_d)\in X^d$ we write
$\Hc|_{\mathbf{x}}=\{(h(x_1),\dots,h(x_d)) : h\in\Hc\}\subseteq Y^d$.
A labeled sample is a finite sequence $S=((x_1,y_1),\dots,(x_m,y_m))$ in $X\times Y$.
Its version space is $\VS(S)=\{h\in\Hc : h(x_i)=y_i\text{ for all }i\le m\}$, and $S$ is realizable if $\VS(S)\neq\emptyset$.

\paragraph{The PAC model.}
We work first in the realizable PAC model.
A distribution $D$ on $X$ and a target $h^\star\in\Hc$ are fixed and unknown.
The learner receives $S_m=((x_i,h^\star(x_i)))_{i\le m}$, where $x_1,\dots,x_m$ are drawn i.i.d.\ from $D$, and outputs a predictor $f\colon X\to Y$, which need not belong to $\Hc$; that is, a learner is a map from samples to predictors.
The error of $f$ is $\er_D(f)=\P_{x\sim D}[f(x)\neq h^\star(x)]$.
The sample complexity $M_{\Hc}(\eps,\delta)$ is the least $m$ such that some learner, given $m$ examples, satisfies $\P[\er_D(f)\le\eps]\ge1-\delta$ for every $D$ and every $h^\star\in\Hc$.
We call such a pair $(D,h^\star)$ a \emph{realizable pair}, and a learner that satisfies this condition for every realizable pair is \emph{$(\eps,\delta)$-correct on all realizable pairs} with $m$ examples.
The class is PAC learnable if $M_{\Hc}(\eps,\delta)<\infty$ for all $\eps,\delta\in(0,1)$.
A \emph{consistent learner} is any map that sends every realizable sample $S$ to some element of $\VS(S)$.
In the realizable case these are exactly the empirical risk minimizers (ERMs).
We allow arbitrary tie-breaking: the choice of the element of $\VS(S)$ may be adversarial.

In the agnostic model, $D$ is a distribution on $X\times Y$, the error is $\er_D(f)=\P_{(x,y)\sim D}[f(x)\neq y]$, and the learner must output $f$ with $\er_D(f)\le\inf_{h\in\Hc}\er_D(h)+\eps$ with probability at least $1-\delta$.
The agnostic sample complexity is defined as in the realizable case.

\paragraph{The online model.}
In round $t=1,2,\dots$ an adversary presents $x_t\in X$, the learner predicts $\hat y_t\in Y$, and the adversary reveals $y_t$.
The sequence is realizable if some $h\in\Hc$ satisfies $h(x_t)=y_t$ for all $t$.
A mistake is a round with $\hat y_t\neq y_t$.
A \emph{consistent online learner} predicts $\hat y_t=f_t(x_t)$ for some $f_t\in\Hc$ that is consistent with $(x_s,y_s)_{s<t}$.

\paragraph{Dimensions.}
We recall the combinatorial dimensions of multiclass learning.
For $d\ge1$, we call the elements of $Y^d$ \emph{words}.

\begin{definition}[Pseudo-cubes and the DS dimension]
\label{def:ds}
Let $d\ge1$ and $B\subseteq Y^d$.
Two words $b,b'\in B$ are \emph{$i$-neighbors} if $b'_j=b_j$ exactly for the indices $j\neq i$.
The \emph{$i$-panel} of $b\in B$ is $\{b'\in B : b'_j=b_j\text{ for all }j\neq i\}$.
For $k\ge1$, $B$ is a \emph{$k$-pseudo-cube} of dimension $d$ if $B$ is finite and nonempty and every $i$-panel of every $b\in B$ has at least $k+1$ elements.
A $1$-pseudo-cube is a \emph{pseudo-cube}: every $b\in B$ has an $i$-neighbor in $B$ for every $i\in[d]$.
A sequence $\mathbf{x}\in X^d$ is \emph{DS-shattered} (\emph{$k$-DS-shattered}) by $\Hc$ if $\Hc|_{\mathbf{x}}$ contains a $d$-dimensional pseudo-cube ($k$-pseudo-cube).
The DS dimension $\dimDS(\Hc)$ and the $k$-DS dimension $\dimDS_k(\Hc)$ are the suprema of the lengths of DS-shattered and $k$-DS-shattered sequences, respectively (the supremum of the empty set is $0$).
\end{definition}

The \emph{neighbor graph} of a set $B\subseteq Y^d$ has the words of $B$ as vertices and an edge between any two words that are $i$-neighbors for some $i$.
Pseudo-cubes and the DS dimension are due to \citet{daniely2014optimal}, and \citet{brukhim2022characterization} proved that $\Hc$ is PAC learnable if and only if $\dimDS(\Hc)<\infty$.
The $k$-DS dimension characterizes list learning with lists of size $k$ \citep[Definition~4]{charikar2023list}; the name $k$-pseudo-cube is from \citet[Definition~1.3]{hanneke2024improved}.

\begin{definition}[Natarajan dimension]
\label{def:natarajan}
A sequence $(x_1,\dots,x_d)\in X^d$ is \emph{N-shattered} by $\Hc$ if there are $f_0,f_1\colon X\to Y$ with $f_0(x_i)\neq f_1(x_i)$ for all $i$, such that for every $\epsilon\in\{0,1\}^d$ some $h\in\Hc$ satisfies $h(x_i)=f_{\epsilon_i}(x_i)$ for all $i$.
The Natarajan dimension $\dimN(\Hc)$ is the supremum of the lengths of N-shattered sequences.
\end{definition}

We also say \emph{Natarajan-shattered} for N-shattered, and we call $f_0$ and $f_1$ the \emph{witnesses}.
An N-shattered sequence has distinct entries, and then $\Hc|_{\mathbf{x}}$ contains the \emph{Natarajan cube} $\prod_i\{f_0(x_i),f_1(x_i)\}$.
A Natarajan cube is a pseudo-cube, so $\dimN(\Hc)\le\dimDS(\Hc)$ for every class \citep[see][]{brukhim2022characterization}.

\begin{definition}[Multiclass Littlestone dimension; \citealp{daniely2015multiclass}]
\label{def:ldim}
A \emph{mistake tree} of depth $d$ is a complete binary tree of depth $d$ in which every internal node $v$ carries a point $x_v\in X$ and the two edges from $v$ to its children carry two distinct labels.
The tree is shattered by $\Hc$ if for every root-to-leaf path, with nodes $v_1,\dots,v_d$ and edge labels $y_1,\dots,y_d$, some $h\in\Hc$ satisfies $h(x_{v_i})=y_i$ for all $i$.
$\Ldim(\Hc)$ is the supremum of the depths of shattered mistake trees.
\end{definition}

\begin{definition}[Teaching dimension]
\label{def:td}
A finite set $A\subseteq X$ is a \emph{teaching set} for $h\in\Hc$ if $h$ is the only element of $\Hc$ that agrees with $h$ on $A$.
$\TD(h)$ is the least size of a teaching set for $h$ ($\infty$ if there is none), and $\TD(\Hc)=\sup_{h\in\Hc}\TD(h)$.
\end{definition}

\subsection{Permutation groups and the Galois connection}
\label{sec:prelim-groups}

Throughout, $\Omega$ is a countable set, $G\le\Sym(\Omega)$ is a group of permutations of $\Omega$, and $g\omega$ denotes the image of $\omega$ under $g$.
Countability makes all events in the PAC model measurable.
For $A\subseteq\Omega$, the \emph{pointwise stabilizer} is $\stab{G}{A}=\{g\in G : ga=a\text{ for all }a\in A\}$, and $G_a=\stab{G}{\{a\}}$.
For a subgroup $K\le G$, $\Fix(K)=\{\omega\in\Omega : k\omega=\omega\text{ for all }k\in K\}$.
Both maps $A\mapsto\stab{G}{A}$ and $K\mapsto\Fix(K)$ reverse inclusions, and $A\subseteq\Fix(K)$ holds if and only if $K\le\stab{G}{A}$.
So the two maps form a Galois connection between subsets of $\Omega$ and subgroups of $G$.
The associated closure operator and its relative version on a set $X\subseteq\Omega$ are
\[
  \cl(A)=\Fix\big(\stab{G}{A}\big),\qquad \cl_X(A)=\cl(A)\cap X .
\]

\begin{lemma}[Closure]
\label{lem:closure}
For all $A,B\subseteq\Omega$:
\textup{(a)} $A\subseteq\cl(A)$;
\textup{(b)} if $A\subseteq B$ then $\cl(A)\subseteq\cl(B)$;
\textup{(c)} $\stab{G}{\cl(A)}=\stab{G}{A}$, and hence $\cl(\cl(A))=\cl(A)$.
\end{lemma}

\begin{proof}
(a) $\stab{G}{A}$ fixes every point of $A$.
(b) If $A\subseteq B$ then $\stab{G}{B}\le\stab{G}{A}$, and a point fixed by $\stab{G}{A}$ is fixed by $\stab{G}{B}$.
(c) By (a), $\stab{G}{\cl(A)}\le\stab{G}{A}$; by definition $\stab{G}{A}$ fixes $\cl(A)$ pointwise, so $\stab{G}{A}\le\stab{G}{\cl(A)}$.
Then $\cl(\cl(A))=\Fix(\stab{G}{\cl(A)})=\Fix(\stab{G}{A})=\cl(A)$.
\end{proof}

\begin{definition}[Galois invariants of $(G,X)$]
\label{def:invariants}
Let $X\subseteq\Omega$ be nonempty.
\begin{enumerate}[label=(\roman*)]
\item A set $A\subseteq X$ is \emph{independent} if $a\notin\cl(A\setminus\{a\})$ for every $a\in A$; equivalently, $\stab{G}{A\setminus\{a\}}$ moves $a$ for every $a\in A$.
  The \emph{height} is $\hgt_X(G)=\sup\{|A| : A\subseteq X\text{ finite and independent}\}$.
\item For $k\ge1$, the \emph{$k$-height} is $\hgt^k_X(G)=\sup\{|A| : A\subseteq X\text{ finite},\ |\stab{G}{A\setminus\{a\}}\,a|\ge k+1\text{ for all }a\in A\}$.
  Since $a$ is moved by $\stab{G}{A\setminus\{a\}}$ exactly when its orbit under this group has at least two points, $\hgt^1_X(G)=\hgt_X(G)$.
\item A sequence $(a_1,\dots,a_k)$ in $X$ is \emph{irredundant} if $a_i\notin\cl(\{a_1,\dots,a_{i-1}\})$ for every $i\le k$.
  $\irr_X(G)$ is the supremum of the lengths of irredundant sequences.
\item A finite set $A\subseteq X$ is a \emph{relative base} if $X\subseteq\cl(A)$, equivalently $\stab{G}{A}\le\stab{G}{X}$.
  $\bsz_X(G)$ is the least size of a relative base ($\infty$ if there is none), and $\Bsz_X(G)$ is the supremum of the sizes of the inclusion-minimal relative bases.
\end{enumerate}
\end{definition}

We also call $\hgt^k_X(G)$ the \emph{list height}; \cref{thm:dictionary}(b) identifies it with the $k$-DS dimension.
For $X=\Omega$ we have $\stab{G}{\Omega}=1$, so a relative base is a base in the usual sense ($\stab{G}{A}=1$).
Then $\hgt_\Omega(G)$, $\irr_\Omega(G)$, $\bsz_\Omega(G)$ and $\Bsz_\Omega(G)$ are the height, the maximal length of an irredundant base, the minimal base size and the maximal size of a minimal base of the permutation group $G$ \citep{cameron1995bases,gill2021height,kelsey2021relational}.
In that case we drop the subscript and write $\hgt(G)$, $\irr(G)$, $\bsz(G)$ and $\Bsz(G)$.
The next remark justifies this identification and records the order between the four invariants.

\begin{remark}[The chain $\bsz\le\Bsz\le\hgt\le\irr$]
\label{rem:chain}
(a) A finite independent set, listed in any order, is an irredundant sequence, because $\cl(\{a_1,\dots,a_{i-1}\})\subseteq\cl(A\setminus\{a_i\})$ by \cref{lem:closure}(b).
Hence $\hgt_X(G)\le\irr_X(G)$.
(b) An inclusion-minimal relative base $A$ is independent.
Indeed, if $a\in\cl(A\setminus\{a\})$ for some $a\in A$, then $A\subseteq\cl(A\setminus\{a\})$ and, by \cref{lem:closure}(b,c), $X\subseteq\cl(A)\subseteq\cl(A\setminus\{a\})$, so $A\setminus\{a\}$ is a smaller relative base.
Hence $\Bsz_X(G)\le\hgt_X(G)$.
(c) Suppose $\irr_X(G)<\infty$.
An irredundant sequence of maximal length cannot be extended, so every $x\in X$ lies in the closure of its points; that is, it is a relative base.
Conversely, an irredundant sequence whose points form a relative base cannot be extended.
So $\irr_X(G)$ is the maximal length of an \emph{irredundant relative base}; for $X=\Omega$ this is the classical maximal irredundant base length.
Relative bases then exist, and a relative base of minimum size is inclusion-minimal, so $\bsz_X(G)\le\Bsz_X(G)$.
In total, $\bsz_X(G)\le\Bsz_X(G)\le\hgt_X(G)\le\irr_X(G)$ whenever $\irr_X(G)<\infty$.
\end{remark}

\paragraph{The standing hypothesis.}
Several results need the following finiteness condition on $G$ and the domain $X$:
\begin{equation}
\label{eq:F}
\tag{F}
\text{every orbit } Gx=\{gx : g\in G\},\ x\in X,\ \text{is finite.}
\end{equation}
Condition~\eqref{eq:F} holds for every finite group.
It also holds for absolute Galois groups acting on separable closures (\cref{sec:fields}).
We state explicitly which results use it.

\subsection{Group classes}
\label{sec:prelim-classes}

For a nonempty $X\subseteq\Omega$, the \emph{group class} is
\[
  \HG{G}{X}=\{\,g|_X : g\in G\,\}\subseteq\Omega^X ,
\]
with label set $Y=\Omega$.
Learning $\HG{G}{X}$ means learning an unknown $g\in G$ from examples $(x,gx)$, as far as $g$ is visible on $X$.
For $g\in G$ and a finite $A\subseteq X$ we write $S_g(A)=\{(a,ga) : a\in A\}$.
A sample that is realizable by $\HG{G}{X}$ and labeled by $g|_X$ has the same version space as $S_g(A)$, where $A$ is its set of points; so version spaces are indexed by pairs $(g,A)$.
Distinct elements of $G$ may have the same restriction to $X$, which causes no difficulty.
We speak of the \emph{full domain} when $X=\Omega$.

\begin{remark}[\eqref{eq:F} cannot be dropped from $\dimDS=\hgt$]
\label{rem:F-necessary}
Let $G=\{x\mapsto\pm x+c : c\in\Z\}$, the infinite dihedral group, act on $\Omega=X=\Z$.
Its orbits are infinite.
The pair $\{0,1\}$ is independent: $x\mapsto-x$ fixes $0$ and moves $1$, and $x\mapsto2-x$ fixes $1$ and moves $0$.
The pointwise stabilizer of two distinct points is trivial, so no $3$-set is independent and $\hgt_X(G)=2$.
For $a\neq b$ and $\delta=b-a$, the restriction of the group class to $(a,b)$ is $\HG{G}{X}|_{(a,b)}=\{(c,c+\delta),(c,c-\delta) : c\in\Z\}$.
In this set the word $(c,c+\delta)$ has exactly one $1$-neighbor, $(c+2\delta,c+\delta)$, and exactly one $2$-neighbor, $(c,c-\delta)$, and the same holds for $(c,c-\delta)$ by symmetry.
So the neighbor graph is a union of bi-infinite paths, along which the first coordinate is unbounded, and no finite nonempty subset is a pseudo-cube.
Hence $\dimDS(\HG{G}{X})=1<2=\hgt_X(G)$.
\end{remark}

\section{The Galois correspondence of a group class}
\label{sec:galois}

Throughout this section $G\le\Sym(\Omega)$, $X\subseteq\Omega$ is nonempty, and $\Hc=\HG{G}{X}$.
We begin with the observation on which the paper rests.

\begin{lemma}[Version spaces are cosets]
\label{lem:coset}
Let $g\in G$, let $A\subseteq X$ be finite, and let $S=S_g(A)$.
\begin{enumerate}[label=(\roman*)]
\item $\VS(S)=\{g'|_X : g'\in g\,\stab{G}{A}\}$.
\item For every $x\in X$, the set of labels $\{h(x) : h\in\VS(S)\}$ equals $g\big(\stab{G}{A}\,x\big)$.
\item All $h\in\VS(S)$ agree at $x\in X$ if and only if $x\in\cl(A)$; in that case $h(x)=gx$ for every $h\in\VS(S)$.
\end{enumerate}
\end{lemma}

\begin{proof}
(i) For $g'\in G$, the restriction $g'|_X$ is consistent with $S$ if and only if $g'a=ga$ for all $a\in A$, that is, $g^{-1}g'\in\stab{G}{A}$, that is, $g'\in g\,\stab{G}{A}$.
Every element of $\VS(S)\subseteq\Hc$ has the form $g'|_X$ with $g'\in G$.
(ii) By (i), the set of labels is $\{g'x : g'\in g\,\stab{G}{A}\}=\{gkx : k\in\stab{G}{A}\}=g(\stab{G}{A}\,x)$.
(iii) Since $g$ is a bijection, the set in (ii) is a singleton if and only if $\stab{G}{A}\,x=\{x\}$, that is, $x\in\Fix(\stab{G}{A})=\cl(A)$; the singleton is then $\{gx\}$.
\end{proof}

So a consistent hypothesis is correct on $\cl_X(A)$, and nowhere else is the label determined: a sample forces the label exactly at the points fixed by the symmetries it cannot see.
In \cref{fig:hero}(a), the example $\sqrt2\mapsto-\sqrt2$ forces every label on $\Q(\sqrt2)$ and leaves $\sqrt3\mapsto\pm\sqrt3$ open.
At a point $x\notin\cl(A)$ the candidate labels form the orbit $g(\stab{G}{A}\,x)$, which is finite under~\eqref{eq:F}.

The second lemma identifies the points of a sample that carry information.
Call $x\in S$ \emph{essential} in a finite set $S\subseteq X$ if $x\notin\cl(S\setminus\{x\})$, and write $E(S)$ for the set of essential points.
By \cref{lem:coset}, these are the points whose label a consistent learner cannot infer from the other points of $S$.

\begin{lemma}[Essential points are independent]
\label{lem:essential}
Let $S\subseteq X$ be finite.
Then $E(S)$ is independent, so $|E(S)|\le\hgt_X(G)$.
Moreover, every inclusion-minimal $S'\subseteq S$ with $\cl(S')=\cl(S)$ is independent; hence $\cl(S)=\cl(S')$ for an independent $S'\subseteq S$ with $|S'|\le\hgt_X(G)$.
\end{lemma}

\begin{proof}
Let $e\in E(S)$.
Since $E(S)\setminus\{e\}\subseteq S\setminus\{e\}$, \cref{lem:closure}(b) gives $\cl(E(S)\setminus\{e\})\subseteq\cl(S\setminus\{e\})$, and $e\notin\cl(S\setminus\{e\})$.
So $E(S)$ is independent.
For the second claim, let $S'\subseteq S$ be inclusion-minimal with $\cl(S')=\cl(S)$; it exists because $S$ is finite.
If $s\in\cl(S'\setminus\{s\})$ for some $s\in S'$, then $S'\subseteq\cl(S'\setminus\{s\})$, and \cref{lem:closure}(b,c) gives $\cl(S')\subseteq\cl(S'\setminus\{s\})\subseteq\cl(S')$.
Then $\cl(S'\setminus\{s\})=\cl(S)$, which contradicts minimality.
\end{proof}

We can now state the dictionary.
Parts (a) and (b) use the standing hypothesis~\eqref{eq:F}; parts (c) and (d) hold for every permutation group.

\begin{restatable}[The dictionary]{theorem}{ThmDictionary}
\label{thm:dictionary}
Let $G\le\Sym(\Omega)$, let $X\subseteq\Omega$ be nonempty, and let $\Hc=\HG{G}{X}$.
\begin{enumerate}[label=(\alph*)]
\item \emph{(DS dimension $=$ height.)}
  Assume~\eqref{eq:F}.
  A sequence $\mathbf{a}=(a_1,\dots,a_d)$ in $X$ is DS-shattered by $\Hc$ if and only if its entries are distinct and $\{a_1,\dots,a_d\}$ is independent.
  In that case $\Hc|_{\mathbf{a}}$ is itself a $d$-dimensional pseudo-cube.
  Hence $\dimDS(\Hc)=\hgt_X(G)$, where both sides may be infinite.
\item \emph{(List DS dimension $=$ list height.)}
  Assume~\eqref{eq:F} and let $k\ge1$.
  A sequence $\mathbf{a}=(a_1,\dots,a_d)$ in $X$ with $A=\{a_1,\dots,a_d\}$ is $k$-DS-shattered by $\Hc$ if and only if its entries are distinct and $|\stab{G}{A\setminus\{a_i\}}\,a_i|\ge k+1$ for every $i$.
  In that case $\Hc|_{\mathbf{a}}$ is itself a $k$-pseudo-cube.
  Hence $\dimDS_k(\Hc)=\hgt^k_X(G)$.
\item \emph{(Littlestone dimension $=$ irredundant length.)}
  $\Ldim(\Hc)=\irr_X(G)$.
\item \emph{(Teaching $=$ bases.)}
  A finite $A\subseteq X$ is a teaching set for $h\in\Hc$ if and only if $A$ is a relative base.
  Consequently all $h\in\Hc$ have the same teaching sets, $\TD(h)=\bsz_X(G)$ for every $h\in\Hc$, and the supremum of the sizes of the inclusion-minimal teaching sets of any $h\in\Hc$ is $\Bsz_X(G)$.
\end{enumerate}
\end{restatable}

The proof is in \cref{app:galois}.

\paragraph{Idea of the proof.}
Each multiclass dimension is defined by local moves: a neighbor in one coordinate of a pseudo-cube, two distinct edge labels at a node of a mistake tree, or a second hypothesis that agrees with the target on a candidate teaching set.
Fix a sequence $\mathbf a=(a_1,\dots,a_d)$ in $X$ with point set $A$, and write $w(g)=(ga_1,\dots,ga_d)$.
By \cref{lem:coset}, a word that differs from $w(g)$ only in coordinate $i$ has the form $w(gk)$ with $k\in\stab{G}{A\setminus\{a_i\}}$ and $ka_i\neq a_i$.
So such a move exists at one word if and only if it exists at every word: left multiplication by $g$ permutes the words of $\Hc|_{\mathbf{a}}$.
This homogeneity is what general classes lack, where the existence of a neighbor depends on the word, and it is why one condition on the set of points (independence) decides DS shattering.
Condition~\eqref{eq:F} is used only to make $\Hc|_{\mathbf{a}}$ finite.
The Littlestone and teaching statements follow the same pattern, with moves along a chain (irredundant sequences) and with the absence of any move (relative bases).

\begin{figure}[t]
\centering
\begin{tikzpicture}[
  w/.style={draw, rounded corners=2pt, inner sep=3pt, fill=white, font=\small},
  e1/.style={line width=1.4pt, FigBlue},
  e2/.style={line width=1.4pt, FigOrange},
  lab/.style={font=\scriptsize}]
\node[w] (a) at (0,0)     {$(u_1,u_2)$};
\node[w] (b) at (2.8,0)   {$(v_1,u_2)$};
\node[w] (c) at (0,2.4)   {$(u_1,v_2)$};
\node[w] (d) at (2.8,2.4) {$(v_1,v_2)$};
\draw[e1] (a) -- node[below,lab]{$s_1$} (b);
\draw[e1] (c) -- node[above,lab]{$s_1$} (d);
\draw[e2] (a) -- node[left,lab]{$s_2$} (c);
\draw[e2] (b) -- node[right,lab]{$s_2$} (d);
\node[align=center,font=\small] at (1.4,-1.0) {Natarajan square\\[1pt] $s_1s_2=s_2s_1$,\ \ $\langle s_1,s_2\rangle\cong C_2\times C_2$};
\begin{scope}[shift={(8.4,1.2)}]
\def\R{1.85}
\node[w] (h12) at (90:\R)  {$(1,2)$};
\node[w] (h32) at (30:\R)  {$(3,2)$};
\node[w] (h31) at (-30:\R) {$(3,1)$};
\node[w] (h21) at (-90:\R) {$(2,1)$};
\node[w] (h23) at (-150:\R){$(2,3)$};
\node[w] (h13) at (150:\R) {$(1,3)$};
\draw[e1] (h12) -- node[above right,lab]{$s_1$} (h32);
\draw[e2] (h32) -- node[right,lab]{$s_2$} (h31);
\draw[e1] (h31) -- node[below right,lab]{$s_1$} (h21);
\draw[e2] (h21) -- node[below left,lab]{$s_2$} (h23);
\draw[e1] (h23) -- node[left,lab]{$s_1$} (h13);
\draw[e2] (h13) -- node[above left,lab]{$s_2$} (h12);
\end{scope}
\node[align=center,font=\small] at (8.4,-2.15) {$S_3$ on $\{1,2,3\}$, restricted to $(1,2)$:\\[1pt] a pseudo-cube ($\dimDS=2$) with no Natarajan square ($\dimN=1$);\\[1pt] $\langle s_1,s_2\rangle\cong S_3$ is not abelian};
\draw[e1] (-0.3,-2.0) -- (0.3,-2.0); \node[anchor=west,font=\scriptsize] at (0.35,-2.0) {changes coordinate 1};
\draw[e2] (-0.3,-2.45) -- (0.3,-2.45); \node[anchor=west,font=\scriptsize] at (0.35,-2.45) {changes coordinate 2};
\end{tikzpicture}
\caption{Commuting and non-commuting moves.
An edge labeled $s_i$ joins two words that differ only in coordinate~$i$; $s_i$ maps each word to its unique $i$-neighbor (\cref{def:monodromy}).
Left: on a Natarajan square the two moves commute and generate $C_2\times C_2$, where $C_n$ is the cyclic group of order~$n$.
Right: the restriction of the $S_3$ class to $(1,2)$ (\cref{sec:galois}).
Every word has exactly one neighbor in each coordinate (the pseudo-cube is \emph{two-fibered}), so $\dimDS=2$, but the moves generate $S_3$ and the hexagon contains no Natarajan square.
By \cref{thm:monodromy}(d), a two-fibered pseudo-cube is a union of Natarajan cubes exactly when its moves commute.}
\label{fig:hexagon}
\end{figure}
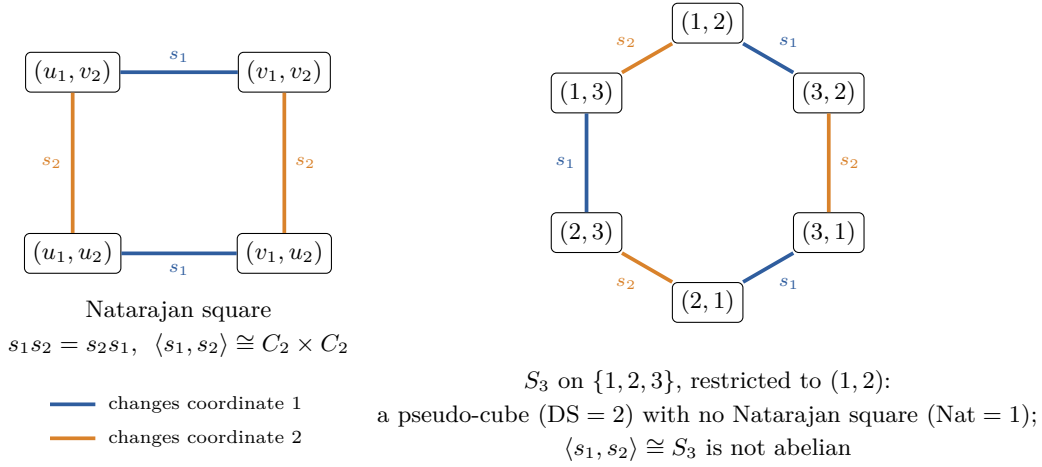

\paragraph{Why it is true.}
Let $G=S_3$ act on $\{1,2,3\}$.
The stabilizer of $1$ is $\{\mathrm{id},(2\,3)\}$, so $\cl(\{1\})=\{1\}$; the stabilizer of $\{1,2\}$ is trivial, so $\cl(\{1,2\})=\{1,2,3\}$.
Thus $\{1,2\}$ is independent and $\hgt=2$.
The restriction $\Hc|_{(1,2)}$ consists of the six ordered pairs of distinct labels.
They form a hexagon (\cref{fig:hexagon}),
\[
(1,2)\overset{1}{-}(3,2)\overset{2}{-}(3,1)\overset{1}{-}(2,1)\overset{2}{-}(2,3)\overset{1}{-}(1,3)\overset{2}{-}(1,2),
\]
in which every word has exactly one $1$-neighbor and one $2$-neighbor; so $\dimDS=2$.
The hexagon contains no Natarajan square $\{u_1,v_1\}\times\{u_2,v_2\}$: such a square needs four distinct labels (\cref{prop:symmetric}(a)), and only three exist.
We return to this hexagon in \cref{sec:monodromy}.

\paragraph{Takeaway.}
For group classes the DS, Littlestone and teaching dimensions are not new quantities: they are classical invariants of one closure system, read through independent sets (DS), chains (Littlestone) and generating sets (teaching).

\begin{table}[t]
\centering
\small
\setlength{\tabcolsep}{4pt}
\renewcommand{\arraystretch}{1.15}
\begin{tabular}{@{}>{\raggedright\arraybackslash}p{2.75cm}>{\raggedright\arraybackslash}p{3.45cm}>{\raggedright\arraybackslash}p{2.45cm}>{\raggedright\arraybackslash}p{2.05cm}>{\raggedright\arraybackslash}p{2.2cm}>{\raggedright\arraybackslash}p{1.35cm}@{}}
\toprule
Learning notion & Invariant of $(G,X)$ & $\Gamma=\Gal(L/K)$ on $L$ & $S_n$ on $[n]$ & $\Gamma\cong S_n$ & Result\\
\midrule
Labels forced by a sample on $A$ & $\cl_X(A)=\Fix(\stab{G}{A})\cap X$ & $K(A)$ & $A$ if $|A|\le n-2$; else $[n]$ & $K(A)$ & \ref{lem:coset}, \ref{thm:galois}\\
Labels at $x$ consistent with the sample & $|\stab{G}{A}\,x|$ & $[K(A)(x):K(A)]$ & $n-|A|$ if $x\notin A$ & $[K(A)(x):K(A)]$ & \ref{lem:coset}, \ref{thm:galois}\\
DS dimension & height $\hgt_X(G)$ & $\mu(\Gamma)$ & $n-1$ & $n-1$ & \ref{thm:dictionary}, \ref{thm:galois}\\
$k$-DS dimension & $k$-height $\hgt^k_X(G)$ & -- & $n-k$ & -- & \ref{thm:dictionary}\\
Littlestone dimension & $\irr_X(G)$ & $\ell(\Gamma)$ & $n-1$ & $\lceil 3n/2\rceil-b(n)-1$ & \ref{thm:dictionary}, \ref{thm:galois}\\
Teaching dimension & $\bsz_X(G)$ & $1$ & $n-1$ & $1$ & \ref{thm:dictionary}, \ref{thm:galois}\\
Largest minimal teaching set & $\Bsz_X(G)$ & -- & $n-1$ & -- & \ref{thm:dictionary}\\
Natarajan dimension & $\le\hgt_X(G)$; $=\hgt_X(G)$ if $G$ is abelian; $\ge\hgt_X(U)$ for abelian $U\le G$ & $\ge\mu(\Delta)$ for abelian $\Delta\le\Gamma$ & $\lfloor n/2\rfloor$ & $n-1$ if $n\le4$; between $\lfloor3n/4\rfloor$ and $n-2$ if $n\ge5$ & \ref{thm:natarajan}, \ref{prop:symmetric}, \ref{thm:galois}, \ref{thm:solvable}, \ref{thm:abelruffini}\\
\bottomrule
\end{tabular}
\caption{The dictionary between learning notions and Galois invariants.
The DS and $k$-DS rows assume~\eqref{eq:F}, which holds in the last three columns.
In the Galois columns, $L/K$ is a finite Galois extension of a countable field, $A\subseteq L$ is finite, the teaching row assumes $\Gamma\neq1$, $\mu$ is the largest size of an independent subset of $\Gamma$ (not necessarily generating) and $\ell$ the length of a longest subgroup chain (\cref{sec:fields}); $b(n)$ is the number of ones in the binary expansion of $n$.
The $S_n$ columns assume $n\ge2$, and the $k$-DS entry assumes $1\le k\le n-1$ (\cref{ex:symmetric}).
A dash marks a value we do not determine.
The last column points to the result that proves the row.}
\label{tab:dictionary}
\end{table}

\begin{remark}[Consequences of the dictionary]
\label{rem:dictionary}
(a) By \cref{rem:chain}, if $\Ldim(\Hc)<\infty$ and \eqref{eq:F} holds, then $\TD(\Hc)\le\dimDS(\Hc)\le\Ldim(\Hc)$.
For general classes the teaching dimension is not bounded by the DS dimension, even for binary classes: the singletons together with the empty set, on $n$ points, have VC dimension $1$ and teaching dimension $n$.
(b) The proof of \cref{thm:dictionary}(a) in \cref{app:galois} shows that a DS-shattered sequence consists of distinct points forming an independent set without using~\eqref{eq:F}.
So $\dimDS(\Hc)\le\hgt_X(G)$ for every group class, and \cref{rem:F-necessary} shows that the reverse inequality can fail without~\eqref{eq:F}.
\end{remark}

\begin{remark}[Transfer from permutation group theory]
\label{rem:transfer}
Results about bases in group theory now become learning bounds.
Let $G\le S_n$ be primitive on $\Omega=[n]$ and not a primitive large-base group (a subgroup of $S_m\wr S_r$ containing $A_m^r$, in product action, with $S_m$ acting on $k$-subsets of $[m]$; see \citealp[Definition~1.2]{gill2021height}).
\citet[Theorem~1.1]{kelsey2021relational} prove that $\irr(G)<5\log_2 n$.
By \cref{thm:dictionary}(c) and \cref{rem:chain}, the full-domain class $\HG{G}{[n]}$ has $\Ldim<5\log_2 n$, and also $\dimDS<5\log_2 n$ and $\TD<5\log_2 n$.
By \cref{thm:erm} below, \emph{every} consistent online learner makes fewer than $5\log_2 n$ mistakes on such a class.
For the height alone, the earlier bound $\hgt(G)<9\log_2 n$ is due to \citet[Theorem~1.3]{gill2021height}.
Large-base groups are genuine exceptions: $S_n$ in its natural action has $\irr=\hgt=n-1$ (\cref{ex:symmetric}).
\end{remark}

We close the section with three examples, each worked out from the definitions.

\begin{example}[Symmetric groups]
\label{ex:symmetric}
Let $G=S_n$ act on $[n]$, $n\ge2$, and $X=[n]$.
For $A\subseteq[n]$, $\stab{G}{A}=\Sym([n]\setminus A)$, so $\cl(A)=A$ if $|A|\le n-2$ and $\cl(A)=[n]$ if $|A|\ge n-1$.
Hence the independent sets are the sets of size at most $n-1$, the irredundant sequences are the sequences of at most $n-1$ distinct points, and the relative bases are the sets of size at least $n-1$.
So $\bsz=\Bsz=\hgt=\irr=n-1$.
For $a\in A$, the orbit $\stab{G}{A\setminus\{a\}}\,a$ is $[n]\setminus(A\setminus\{a\})$, of size $n-|A|+1$; so $\hgt^k=n-k$ for $1\le k\le n-1$.
By \cref{thm:dictionary}, learning a permutation from a partial matching has $\dimDS=\Ldim=\TD=n-1$ and $\dimDS_k=n-k$.
Its Natarajan dimension is $\lfloor n/2\rfloor$ (\cref{prop:symmetric}).
\end{example}

\begin{example}[Linear and affine maps over a finite field]
\label{ex:linear}
Let $G=\mathrm{GL}(d,q)$ act on $\Omega=X=\mathbb{F}_q^d$, with $(d,q)\neq(1,2)$.
Then $\cl(A)$ is the linear span $W$ of $A$.
Indeed, a linear map that fixes $A$ fixes $W$.
Conversely, let $v\notin W$ and choose a complement $U$ of $W$ that contains $v$.
If $q\ge3$, the map that is the identity on $W$ and multiplication by some $\lambda\notin\{0,1\}$ on $U$ fixes $W$ and moves $v$.
If $q=2$ and $\dim U\ge2$, extend $v$ by some $u$ to a basis of $U$ and use the map that fixes $W$ and the other basis vectors and sends $v\mapsto v+u$.
If $q=2$ and $\dim U=1$, then $W\neq0$ because $d\ge2$; pick $w\in W\setminus\{0\}$ and use the map that fixes $W$ and sends $v\mapsto v+w$.
Hence independent sets, irredundant sequences and relative bases are linearly independent sets, linearly independent sequences and spanning sets, and $\bsz=\Bsz=\hgt=\irr=d$.
Learning an invertible linear map from point correspondences therefore has $\dimDS=\Ldim=\TD=d$.
For the affine group $\mathrm{AGL}(d,q)$ acting on $\mathbb{F}_q^d$, conjugate by a translation that moves a point of a nonempty $A$ to $0$.
Affine maps that fix $0$ are linear, so the pointwise stabilizer of the translated set is its pointwise stabilizer in $\mathrm{GL}(d,q)$, whose fixed points form its linear span.
Translating back, $\cl(A)$ is the affine span of $A$; and $\cl(\emptyset)=\emptyset$ because translations fix no point.
So independent sets are affinely independent sets, relative bases are affinely spanning sets, and $\bsz=\Bsz=\hgt=\irr=d+1$.
\end{example}

\begin{example}[Projective transformations]
\label{ex:projective}
Let $G=\mathrm{PGL}(d+1,q)$ act on the points of the projective space $\Omega=X=\mathbb{P}^d(\mathbb{F}_q)$, with $d\ge1$ and $q\ge3$.
A frame, that is, $d+2$ points of which every $d+1$ have linearly independent representatives, is a base: in the basis given by $d+1$ of its points, an element that fixes these points is diagonal, and fixing the last point $e_0+\dots+e_d$ forces all diagonal entries to be equal.
No $d+1$ points form a base.
If their representatives are linearly independent, the diagonal map $\mathrm{diag}(\lambda,1,\dots,1)$ in that basis, with $\lambda\notin\{0,1\}$, fixes them and is not scalar.
Otherwise they lie in a hyperplane $U=\ker\varphi$, and the transvection $x\mapsto x+\varphi(x)u$ with $u\in U\setminus\{0\}$ fixes them and is not scalar.
Hence $\bsz=d+2$ and, by \cref{thm:dictionary}(d), $\TD=d+2$: a projective transformation is taught by the images of a frame and by no set of $d+1$ points.
The same linear algebra works over every field with at least three elements; over $\R$ and $d=2$ it is the familiar fact that four point correspondences in general position, and no three, determine a planar homography.
\end{example}

\section{Every ERM is rate-optimal}
\label{sec:erm}

For general multiclass classes, different ERMs can have different sample complexities \citep{daniely2015multiclass}, and no ERM need be optimal \citep{daniely2014optimal}.
For group classes, the choice among consistent hypotheses matters only up to absolute constants in the worst case.
Parts (a), (b) and (d) concern the realizable PAC model, part (c) is a lower bound for deterministic (possibly improper) learners, and part (e) concerns the online model.
Recall that $\hgt=\hgt_X(G)$ and $\irr=\irr_X(G)$.

\begin{restatable}[Every consistent learner is rate-optimal]{theorem}{ThmErm}
\label{thm:erm}
Let $G\le\Sym(\Omega)$, let $X\subseteq\Omega$ be nonempty, and let $\Hc=\HG{G}{X}$.
\begin{enumerate}[label=(\alph*)]
\item \emph{(Expected error.)}
  Let $\Omega$ be countable and $\hgt<\infty$.
  For every consistent learner $\mathcal{A}$, every distribution $D$ on $X$, every target $h^\star\in\Hc$ and every $m\ge0$,
  \[
    \E_{x_1,\dots,x_m\sim D}\big[\er_D(\mathcal{A}(S_m))\big]\le\frac{\hgt}{m+1}.
  \]
\item \emph{(High probability.)}
  Let $\Omega$ be countable, $\hgt<\infty$, $\delta\in(0,1)$ and $m\ge1$.
  With probability at least $1-\delta$ over $S_m\sim D^m$, \emph{every} $f\in\VS(S_m)$ satisfies
  \[
    \er_D(f)\le\frac{48}{m}\Big(\hgt\log_2880+\log_2\frac{24}{\delta}\Big).
  \]
  In particular, if $\hgt\ge1$ then $\er_D(f)\le c\,(\hgt+\log_2(1/\delta))/m$ with $c=690$.
\item \emph{(Lower bound.)}
  Assume~\eqref{eq:F}.
  Call $\Hc$ \emph{nondegenerate} if there are $x_0,x_1\in X$ and $k\in G_{x_0}$ with $kx_1\neq x_1$; this holds whenever $\hgt\ge2$.
  If $\Hc$ is nondegenerate and $\hgt<\infty$, let $\eps\in(0,1/16)$ and $\delta\in(0,1/8)$.
  Every learner (a deterministic map from samples to predictors, possibly improper) that uses $m$ examples and, for every distribution $D$ on $X$ and every target $h^\star\in\Hc$, outputs a predictor with $\er_D\le\eps$ with probability at least $1-\delta$, satisfies
  \[
    m\ge\frac{(\hgt-1)\ln2+4\ln(1/\delta)}{32\,\eps}.
  \]
  If $\hgt=\infty$, then $\Hc$ is not PAC learnable.
\item \emph{(Optimal rate.)}
  Consequently, if $\Omega$ is countable, \eqref{eq:F} holds, $\Hc$ is nondegenerate and $\hgt<\infty$, then for $\eps\in(0,1/16)$ and $\delta\in(0,1/8)$ the least number of examples with which a deterministic learner is $(\eps,\delta)$-correct on all realizable pairs is $\Theta\big((\hgt+\log_2(1/\delta))/\eps\big)$, with absolute constants: $\lceil 690(\hgt+\log_2(1/\delta))/\eps\rceil$ examples suffice for every consistent learner, and $(\ln 2/64)(\hgt+\log_2(1/\delta))/\eps$ examples are necessary.
\item \emph{(Online.)}
  Every consistent online learner makes at most $\irr$ mistakes on every sequence that is realizable by $\Hc$.
  If $\irr<\infty$, then every deterministic online learner errs in every round of some realizable sequence of length $\irr$; if $\irr=\infty$, then for every $n$ every deterministic online learner errs in every round of some realizable sequence of length $n$.
\end{enumerate}
\end{restatable}

Parts (a), (b) and (e) do not use~\eqref{eq:F}.
The full proof is in \cref{app:erm}; we sketch the key step here.

\begin{proof}[Proof sketch of \textup{(a)}]
Draw $x_1,\dots,x_{m+1}$ i.i.d.\ from $D$ and let $T_j$ be the set of these points other than $x_j$.
By \cref{lem:coset}(iii), the learner trained on the first $m$ points is correct at $x_{m+1}$ whenever $x_{m+1}\in\cl(T_{m+1})$.
So $\E[\er_D(\mathcal{A}(S_m))]\le\P[x_{m+1}\notin\cl(T_{m+1})]$.
By exchangeability, this probability equals $\frac{1}{m+1}\E[Z]$, where $Z$ is the number of indices $j$ with $x_j\notin\cl(T_j)$.
Each such $x_j$ occurs once in the sequence and is an essential point of the set $\{x_1,\dots,x_{m+1}\}$.
By \cref{lem:essential}, $Z\le\hgt$.
\end{proof}

\paragraph{Idea of the proof.}
For a general multiclass class, the error region of a consistent hypothesis depends on the hypothesis, which is why tie-breaking matters and ERM can fail.
For a group class, \cref{lem:coset} confines the error region of \emph{every} consistent hypothesis to one set, $X\setminus\cl_X(\{x_1,\dots,x_m\})$, which does not depend on the hypothesis.
The expected bound then follows from \cref{lem:essential}, as in the sketch above.
For the high-probability bound, leave-one-out arguments are too weak.
We use instead that the closed subsets of $X$ (the sets $C\subseteq X$ with $\cl_X(C)=C$) form an intersection-closed concept class whose VC dimension is exactly $\hgt$ (a finite set is shattered by closed sets if and only if it is independent), and we apply the analysis of the closure algorithm for such classes by \citet{darnstadt2015optimal}, which has no $\log(1/\eps)$ factor.
The lower bound is the DS lower bound of \citet{hanneke2024improved}, transported by \cref{thm:dictionary}(a); a short coupling argument passes from their definition of sample complexity, which allows distributions whose best error in the class is approached but not attained, to ours.
The online bound replaces independent sets by irredundant sequences: the mistake rounds of a consistent learner form an irredundant sequence.

\paragraph{Why it is true.}
Let $G=(\Z/2)^d$ act on $d$ disjoint pairs $\{x_i,x_i'\}$, the $i$-th factor swapping the $i$-th pair, so $\hgt=d$.
A sample determines the label of $x_i$ and of $x_i'$ exactly when it contains a point of the $i$-th pair, and a consistent learner, however adversarial, can err only on the pairs not yet seen.
In a leave-one-out experiment with $m+1$ draws, the last point is essential exactly when its pair is hit only once.
At most $d$ pairs can be hit only once, so by symmetry the last point is essential with probability at most $d/(m+1)$.

The lower bound of part~(c), like the bound of \citet{hanneke2024improved} from which it comes, is stated for deterministic learners.
Under~\eqref{eq:F}, against every learner, randomized ones included, the expected-error bound of \cref{thm:erm}(a) is optimal up to a constant factor when $2\le\hgt<\infty$ and $m\ge\max(2,\hgt-1)$.

\begin{restatable}[Expected-error lower bound]{proposition}{PropExpectedLower}
\label{prop:expected-lower}
Assume~\eqref{eq:F}, let $\Omega$ be countable, and let $d\ge1$ be an integer with $d\le\hgt_X(G)$.
For every $m\ge\max(2,d-1)$ and every learner $\mathcal{A}$ (possibly improper, and randomized through a probability kernel) there are a target $h^\star\in\Hc$ and a distribution $D$ supported on at most $d$ points of $X$ with
\[
  \E\big[\er_D(\mathcal{A}(S_m))\big]\ge\frac{d-1}{8m}.
\]
Hence, if $2\le\hgt<\infty$ and $m\ge\max(2,\hgt-1)$, the minimax expected error of $\Hc$ (the infimum over learners of the supremum over $D$ and $h^\star$) lies between $(\hgt-1)/(8m)$ and $\hgt/(m+1)$, and every consistent learner is within a factor $16$ of it.
If $\hgt=\infty$, then for every $m\ge2$ and every learner there are a target and a distribution forcing expected error at least $1/8$.
\end{restatable}

The proof (\cref{app:erm}) puts a uniform prior on the pseudo-cube $\Hc|_A$ of an independent set $A$.
At a point $a_i$ that the sample misses, the target's label is at least as likely to differ from any fixed guess as to equal it, because every word of a pseudo-cube has a neighbor in direction~$i$.

\paragraph{Takeaway.}
For learning a symmetry on a countable domain, the choice among consistent hypotheses matters only up to absolute constants in the worst case: under~\eqref{eq:F}, on nondegenerate classes with $\hgt<\infty$ and for $\eps\in(0,1/16)$ and $\delta\in(0,1/8)$, every consistent learner attains the optimal deterministic sample complexity (\cref{thm:erm}(d)), and every consistent online learner attains the optimal deterministic mistake bound (\cref{thm:erm}(e)).

\paragraph{Relation to the closure algorithm.}
The bounds of \cref{thm:erm}(a) and~(b) are known bounds for the closure algorithm on intersection-closed classes, transferred to group classes.
\citet{helmbold1990nested} show that the closure algorithm on an intersection-closed class of VC dimension~$d$ errs on the $t$-th random example with probability at most $d/t$; \citet{auer2007new} analyze it with high probability, and \citet{darnstadt2015optimal} gives the optimal high-probability bound.
In a group class, every consistent hypothesis is correct on the closure of the sample (\cref{lem:coset}), so every consistent learner errs only where the closure algorithm for the class of closed sets, with target concept $X$ (every example positive), errs, and that class has VC dimension~$\hgt$ (\cref{lem:closed-sets}).
This is why, for group classes, the choice among consistent hypotheses does not affect these worst-case bounds.

\paragraph{Classes with a Natarajan--DS gap.}
The pseudo-cube classes of \citet{brukhim2022characterization}, which separate the Natarajan and DS dimensions, are themselves group classes of finite groups restricted to a set of $d$ base cosets (\cref{sec:natarajan}, after \cref{conj:half}); they satisfy~\eqref{eq:F}, have height $d$, and are nondegenerate when $d\ge2$, so by \cref{thm:erm}(d) every ERM learns them with the optimal realizable sample complexity among deterministic learners, $\Theta((d+\log(1/\delta))/\eps)$ for $\eps<1/16$ and $\delta<1/8$, with absolute constants.

\paragraph{Relation to the star number.}
For binary classes, \citet{hanneke2016refined} studies the error of all consistent learners through the star number: the largest $s$ (or $\infty$ if there is no largest) for which some points $x_1,\dots,x_s$ and hypotheses $h_0,h_1,\dots,h_s$ satisfy $\{j : h_i(x_j)\neq h_0(x_j)\}=\{i\}$ for every $i$; for a class of finite VC dimension $d\ge1$, every consistent learner has rate $\Theta(1/m)$ in $m$ exactly when $s<\infty$ \citep[Corollary~14]{hanneke2016refined}, and his bounds lose no logarithmic factor when $s$ equals the VC dimension. In a group class the star sets, defined in the same way, are exactly the finite independent sets: an independent set $A$ is witnessed by $h_0=\mathrm{id}|_X$ and $h_a=k_a|_X$ with $k_a\in\stab{G}{A\setminus\{a\}}$ moving $a$, and conversely, if $h_i=g_i|_X$, then $g_0^{-1}g_i$ fixes every $x_j$ with $j\neq i$ and moves $x_i$. So the star number of a group class is its height, and \cref{thm:erm} is the multiclass counterpart of this binary phenomenon.

\paragraph{Agnostic learning.}
In the agnostic model the examples $(x,y)$ come from an arbitrary distribution on $X\times\Omega$, and no finiteness of the label set is needed.
The \emph{graph dimension} $d_G(\Hc)$ \citep{daniely2015multiclass} is the supremum of the sizes of sets $A\subseteq X$ for which some $f\colon A\to\Omega$ has the following property: for every $T\subseteq A$ some $h\in\Hc$ agrees with $f$ on $T$ and disagrees with $f$ at every point of $A\setminus T$.
Such a set $A$ is \emph{graph-shattered}, with \emph{witness} $f$.

\begin{restatable}[Agnostic ERM]{proposition}{PropAgnostic}
\label{prop:agnostic}
Let $\Omega$ be countable, $X\subseteq\Omega$, $\Hc=\HG{G}{X}$ and $\hgt=\hgt_X(G)$.
\begin{enumerate}[label=(\alph*)]
\item $\dimN(\Hc)\le d_G(\Hc)=\hgt$.
\item If $\hgt<\infty$, then for all $\eps,\delta\in(0,1)$ every ERM (every learner whose output minimizes the empirical error over $\Hc$, with arbitrary tie-breaking) learns $\Hc$ agnostically from $C_2(\hgt+\ln(1/\delta))/\eps^2$ examples, where $C_2$ is the absolute constant of \citet[Theorem~5]{daniely2015multiclass}.
\item If moreover $|\Hc|\ge2$, then $\dimN(\Hc)\ge1$ and, for all $\eps,\delta\in(0,1)$, this number is at most $(\hgt/\dimN(\Hc))\cdot C_2(\dimN(\Hc)+\ln(1/\delta))/\eps^2$.
\end{enumerate}
\end{restatable}

The graph dimension controls ERM in multiclass learning \citep{daniely2015multiclass}, and for group classes it is the height.
\Citet[Theorem~5]{daniely2015multiclass} also state that every learner in their model, which is deterministic, needs order $(\dimN+\log(1/\delta))/\eps^2$ examples.\footnote{Their statement gives no range of $\eps$ and $\delta$, and as stated it cannot hold for all $\eps,\delta\in(0,1)$: for a class of two hypotheses every ERM succeeds from one example whenever $\delta\ge1-\eps$, while the stated bound tends to infinity as $\eps\to1$ with $\delta=1-\eps$. We do not determine the range in which it holds.}
Wherever it holds, \cref{prop:agnostic}(c) shows that every agnostic ERM is optimal among deterministic learners up to the factor $\hgt/\dimN$ and absolute constants.
This ratio is $1$ for abelian groups (\cref{thm:natarajan}) and for solvable Galois groups (\cref{thm:solvable}), at most $4/3$ for the Galois groups $S_n$ with $n\ge2$ (\cref{thm:solvable}(c)), at most $2$ for $S_n$ acting on $n\ge2$ points (\cref{prop:symmetric}), and at most $2$ for every finite permutation group on its full domain if \cref{conj:half} holds.
The relation between the Natarajan and DS dimensions, studied in \cref{sec:natarajan,sec:solvable}, therefore measures how far this certificate of optimality is from tight.
On finite domains satisfying~\eqref{eq:F}, an improper learner of \citet[Corollary~1.2]{pabbaraju2026optimal} uses $\widetilde O(\dimDS/\eps+(\dimN+\log(1/\delta))/\eps^2)$ examples when the labels also lie in the finite set $G\cdot X$, so when $\hgt>\dimN$ it is open whether ERM really loses the factor $\hgt/\dimN$.

\paragraph{Beyond finite orbits: linear and orthogonal maps.}
\Cref{thm:dictionary,thm:erm} assume a countable domain and, for the DS identity, finite orbits.
The motivating continuous examples fall outside these assumptions, but the closure structure survives in the linear case.

\begin{restatable}[Learning a linear or orthogonal map]{proposition}{PropLinear}
\label{prop:linear}
Let $\mathbb F\in\{\mathbb R,\mathbb C\}$, $d\ge1$, $X=\mathbb F^d$, and let $G$ be $\mathrm{GL}(d,\mathbb F)$, or $\mathrm O(d)$ when $\mathbb F=\mathbb R$, or $\mathrm U(d)$ when $\mathbb F=\mathbb C$, acting by matrix multiplication; let $\Hc=\HG{G}{X}$.
\begin{enumerate}[label=(\alph*)]
\item $\cl(A)=\operatorname{span}(A)$ for every $A\subseteq X$, so independence is linear independence and the height is~$d$.
\item $\dimN(\Hc)=\dimDS(\Hc)=d$.
\item For every Borel distribution $D$ on $X$, every target and every $m\ge0$, every consistent learner has (outer) expected error at most $d/(m+1)$.
\item For every $m\ge\max(2,d-1)$ and every learner there are a target and a distribution on at most $d$ points with expected error at least $(d-1)/(8m)$.
\end{enumerate}
\end{restatable}

So learning an invertible linear map, an orthogonal map, or a unitary map from point correspondences has Natarajan and DS dimension $d$, and, for $d\ge2$ and $m\ge\max(2,d-1)$, every consistent learner is within a factor $16$ of the minimax expected error.
The proof (\cref{app:erm}) follows the finite-orbit arguments.
A consistent linear map agrees with the target on the span of the sample, so the leave-one-out argument applies verbatim, and the event that a sample point leaves the span of the others is Borel.
The error of an arbitrary consistent learner need not be a measurable function of the sample, which is why part~(c) bounds its outer expectation.
The sign matrices $\mathrm{diag}(\pm1)$ give a Natarajan cube on a basis.

\begin{remark}[Computation]
\label{rem:computation}
(a) Let $\Omega=[n]$ and let $G$ be given by a set of generating permutations.
The Schreier--Sims algorithm computes, in time polynomial in $n$ and the number of generators, a stabilizer chain $G=G^{(0)}\ge G^{(1)}\ge\dots\ge G^{(k)}$ with $G^{(i)}=\stab{G}{\{a_1,\dots,a_i\}}$ along any prescribed sequence of points, together with transversals of each $G^{(i)}$ in $G^{(i-1)}$ \citep[Chapter~4]{seress2003permutation}.
Given a sample $S_g(A)$ with $A=\{a_1,\dots,a_k\}$, one finds a consistent hypothesis level by level.
Suppose $g'_{i-1}$ agrees with $g$ on $a_1,\dots,a_{i-1}$.
Then $g_{i-1}'^{-1}g\in G^{(i-1)}$, so the point $y=g_{i-1}'^{-1}ga_i$ lies in the orbit $G^{(i-1)}a_i$.
If $t\in G^{(i-1)}$ is the transversal element with $ta_i=y$, then $g'_i=g'_{i-1}t$ agrees with $g$ on $a_1,\dots,a_i$.
So consistent learning, and hence by \cref{thm:erm} optimal PAC and online learning, is computationally efficient for permutation groups given by generators.
(b) Agnostic ERM for $S_n$ acting on $[n]$ is a maximum-weight bipartite matching problem.
Given a sample $((x_i,y_i))_{i\le m}$, put $c(x,y)=|\{i : (x_i,y_i)=(x,y)\}|$; a permutation $\pi$ maximizes the number of correctly labeled examples if and only if it maximizes $\sum_{x\in[n]}c(x,\pi(x))$, which is the assignment problem on the complete bipartite graph with vertex classes $[n]$ and $[n]$, solvable in polynomial time.
Whether agnostic ERM is tractable for other group classes is discussed in \cref{sec:discussion}.
\end{remark}

\section{Galois groups}
\label{sec:fields}

We now take the literal Galois setting, in which the DS, Littlestone and teaching dimensions become invariants of the abstract Galois group; \cref{app:background} reviews the Galois theory we use.
Let $L/K$ be a finite Galois extension of a countable field $K$, so that $L$ is countable as our standing convention requires, with group $\Gamma=\Gal(L/K)$ acting on $\Omega=X=L$.
The learner must identify an automorphism $\sigma\in\Gamma$ from examples $(\alpha,\sigma\alpha)$.
For a finite field extension $\mathbb{F}_{q^n}/\mathbb{F}_q$ this means learning the exponent $i$ of the map $\alpha\mapsto\alpha^{q^i}$, a power of the Frobenius map.

Two invariants of an abstract finite group $\Delta$ enter.
Write $\langle S\rangle$ for the subgroup generated by a subset $S$.
A subset $\{g_1,\dots,g_h\}\subseteq\Delta$ is \emph{independent} if $g_i\notin\langle g_j : j\neq i\rangle$ for every $i$ (independence of a set of group elements; independent sets of points are those of \cref{def:invariants}), and $\mu(\Delta)$ is the largest size of an independent subset \citep{whiston2000maximal,cameron2002independent}.
(\citet{cameron2002independent} write $\mu'$ for this quantity and reserve $\mu$ for the largest size of an independent generating set.)
The quantity $\ell(\Delta)$ is the largest length $\ell$ of a strictly decreasing chain of subgroups $\Delta=U_0>U_1>\dots>U_\ell=1$ \citep{cameron1989chains}.

The bridge between points and subgroups is the following consequence of the primitive element theorem.

\begin{fact}[Every subgroup is a point stabilizer]
\label{fact:realization}
For every subgroup $U\le\Gamma$ there is $\alpha\in L$ with $\Gamma_\alpha=U$.
\end{fact}

\begin{proof}
The fixed field $L^U=\Fix(U)$, the set of elements of $L$ fixed by every element of $U$, is a finite separable extension of $K$, so $L^U=K(\alpha)$ for some $\alpha$ by the primitive element theorem \citep[Chapter~V]{lang2002algebra}.
An automorphism fixes $\alpha$ if and only if it fixes $K(\alpha)$, so $\Gamma_\alpha=\Gal(L/K(\alpha))=\Gal(L/L^U)=U$ by the fundamental theorem of Galois theory \citep[Chapter~VI, Theorems~1.1 and~1.2]{lang2002algebra}.
\end{proof}

\begin{restatable}[Learning a Galois automorphism]{theorem}{ThmGalois}
\label{thm:galois}
Let $L/K$ be a finite Galois extension of a countable field $K$, with group $\Gamma=\Gal(L/K)$ acting on $\Omega=X=L$, and let $\Hc=\HG{\Gamma}{L}$.
Then~\eqref{eq:F} holds, and:
\begin{enumerate}[label=(\roman*)]
\item $\cl(A)=K(A)$ for every $A\subseteq L$: the learning closure is the field generated by the data;
\item if $\Gamma\neq1$, then $\TD(\Hc)=\bsz_L(\Gamma)=1$;
\item $\dimDS(\Hc)=\hgt_L(\Gamma)=\mu(\Gamma)$; more generally, $\hgt_L(\Delta)=\mu(\Delta)$ for every subgroup $\Delta\le\Gamma$;
\item $\Ldim(\Hc)=\irr_L(\Gamma)=\ell(\Gamma)$;
\item for a finite $A\subseteq L$ and $\alpha\in L$, the set of labels of $\alpha$ consistent with a sample on $A$ has exactly $[K(A)(\alpha):K(A)]$ elements;
\item $\dimN(\Hc)\ge\max\{\mu(\Delta) : \Delta\le\Gamma\text{ abelian}\}$, and for a finite abelian group $\Delta$, $\mu(\Delta)=\sum_p d(\Delta_p)$, where $\Delta_p$ is the Sylow $p$-subgroup and $d(\cdot)$ is the minimal number of generators.
\end{enumerate}
\end{restatable}

The proof is in \cref{app:fields}.
For three families of Galois groups the theorem, together with theorems of \citet{whiston2000maximal} and \citet{cameron1989chains}, gives explicit values.
We write $\omega(n)$ and $\omega^{+}(n)$ for the number of prime factors of $n$ counted without and with multiplicity, and $b(n)$ for the number of ones in the binary expansion of $n$.

\begin{restatable}[Explicit values]{corollary}{CorGaloisValues}
\label{cor:galois-values}
In the setting of \cref{thm:galois}:
\begin{enumerate}[label=(\alph*)]
\item if $\Gamma\cong S_n$ with $n\ge2$, then $\TD=1$, $\dimDS=n-1$, $\Ldim=\lceil3n/2\rceil-b(n)-1$ and $\lfloor n/2\rfloor\le\dimN\le n-1$;
\item if $\Gamma\cong\Z/n$ with $n\ge2$ (for example $\mathbb{F}_{q^n}/\mathbb{F}_q$), then $\TD=1$, $\dimDS=\dimN=\omega(n)$ and $\Ldim=\omega^{+}(n)$;
\item if $\Gamma\cong(\Z/2)^r$ with $r\ge1$, then $\TD=1$ and $\dimDS=\dimN=\Ldim=r$.
\end{enumerate}
\end{restatable}

Case (c) covers the multiquadratic fields $\Q(\sqrt{p_1},\dots,\sqrt{p_r})/\Q$ with distinct primes $p_i$, whose Galois group is $(\Z/2)^r$.\footnote{The subgroup of $\Q^\times/\Q^{\times2}$ generated by $p_1,\dots,p_r$ has order $2^r$, so the field has degree $2^r$ \citep[Theorem~5.30 and Example~5.31(b)]{milne2022fields}, and in particular $\sqrt{p_i}\notin\Q(\sqrt{p_j} : j\neq i)$.}
In case (b), learning a power of the Frobenius map of $\mathbb{F}_{q^{12}}/\mathbb{F}_q$ has $\dimDS=\dimN=2$, $\Ldim=3$ and $\TD=1$.

\paragraph{Idea of the proof.}
The pointwise stabilizer of a finite set is the intersection of the stabilizers of its points, and by \cref{fact:realization} these point stabilizers range over \emph{all} subgroups of $\Gamma$.
An independent set of points then yields an independent set of group elements: for each point $a$ pick $g_a$ in the stabilizer of the other points but not in $\Gamma_a$.
Conversely, an independent set $\{g_1,\dots,g_h\}$ yields the points whose stabilizers are the subgroups $\langle g_j : j\neq i\rangle$; these exist by \cref{fact:realization}.
In the same way, irredundant sequences correspond to strictly decreasing chains of pointwise stabilizers, which are arbitrary subgroup chains, and a base of size one is a primitive element of $L/K$.
The step that a general permutation group lacks is the converse direction: there, not every subgroup is a point stabilizer, and the height can be smaller than $\mu$ (in the regular action of $(\Z/2)^2$ on itself every point stabilizer is trivial, so the height is $1$, while $\mu((\Z/2)^2)=2$).

\paragraph{Why it is true.}
Let $L=\Q(\sqrt2,\sqrt3)$ and $K=\Q$, so $\Gamma=\{1,\sigma,\tau,\sigma\tau\}\cong(\Z/2)^2$, where $\sigma$ negates $\sqrt2$ and fixes $\sqrt3$, and $\tau$ negates $\sqrt3$ and fixes $\sqrt2$.
Then $\cl(\{\sqrt2\})=\Q(\sqrt2)\not\ni\sqrt3$ and $\cl(\{\sqrt3\})=\Q(\sqrt3)\not\ni\sqrt2$, so $\{\sqrt2,\sqrt3\}$ is independent and $\dimDS=2=\mu(\Gamma)$.
After the example $(\sqrt2,-\sqrt2)$ of \cref{fig:hero}(a), the label of $\sqrt3$ is still $\pm\sqrt3$, two values, as $[\Q(\sqrt2,\sqrt3):\Q(\sqrt2)]=2$ predicts.
The primitive element $\theta=\sqrt2+\sqrt3$ has trivial stabilizer, so the single example $(\theta,\sigma'\theta)$ identifies any $\sigma'\in\Gamma$: $\TD=1$.

\paragraph{Takeaway.}
For a Galois group $\Gamma\neq1$, one well-chosen example (a primitive element) identifies the automorphism, while the optimal deterministic sample complexity is $\Theta((\mu(\Gamma)+\log(1/\delta))/\eps)$ and online learning costs $\ell(\Gamma)$ mistakes, both invariants of the abstract group.
(\Cref{thm:erm}(d) applies, for $\eps\in(0,1/16)$ and $\delta\in(0,1/8)$, because the class is nondegenerate: every element of $\Gamma$ fixes $K$ pointwise and some element moves a point of $L$.)

\begin{remark}[Infinite Galois groups]
\label{rem:profinite}
Let $K^{\mathrm{sep}}$ be a separable closure of a countable field $K$ (so $K^{\mathrm{sep}}$ is countable) and let $G=\Gal(K^{\mathrm{sep}}/K)$ act on $\Omega=X=K^{\mathrm{sep}}$.
The orbit of $\alpha$ is its set of conjugates over $K$, which is finite, so~\eqref{eq:F} holds and \cref{thm:dictionary} applies.
For $K=\Q$ the class is not PAC learnable.
Indeed, let $P$ be a finite set of primes and $p\in P$.
By the classical fact quoted above, $\sqrt p\notin F=\Q(\sqrt r : r\in P\setminus\{p\})$, so $F(\sqrt p)/F$ has an automorphism that fixes $F$ and sends $\sqrt p$ to $-\sqrt p$.
This automorphism extends to an element of $G$ \citep[Chapter~V]{lang2002algebra}, which fixes every $\sqrt r$ with $r\in P\setminus\{p\}$ and moves $\sqrt p$.
So $\{\sqrt p : p\in P\}$ is independent for every finite $P$, and $\hgt_X(G)=\infty$.
By \cref{thm:dictionary}(a) and \cref{thm:erm}(c), $\dimDS=\infty$ and the class is not PAC learnable; by \cref{rem:chain}(a), also $\Ldim=\infty$.
Yet by \cref{lem:coset}(ii) and~\eqref{eq:F}, after every finite sample each point has only finitely many candidate labels.
\end{remark}

\section{Commuting symmetries and the Natarajan dimension}
\label{sec:natarajan}

The Natarajan dimension is always at most the DS dimension, and \citet{brukhim2022characterization} showed that the gap between them can be arbitrarily large.
For group classes the gap has a simple source.

\begin{restatable}[Commuting moves give Natarajan shattering]{theorem}{ThmNatarajan}
\label{thm:natarajan}
Let $G\le\Sym(\Omega)$, let $X\subseteq\Omega$ be nonempty, and let $\Hc=\HG{G}{X}$.
\begin{enumerate}[label=(\roman*)]
\item Every N-shattered sequence in $X$ consists of distinct points that form an independent set.
  Hence $\dimN(\Hc)\le\hgt_X(G)$; under~\eqref{eq:F}, $\dimN(\Hc)\le\dimDS(\Hc)=\hgt_X(G)$.
\item \emph{(Commuting moves.)}
  Let $A\subseteq X$ be finite, and suppose that there are pairwise commuting elements $k_a\in\stab{G}{A\setminus\{a\}}$ with $k_aa\neq a$, for $a\in A$.
  Then $A$, in any order, is N-shattered by $\Hc$, with $f_0(a)=a$ and $f_1(a)=k_aa$.
\item If $G$ is abelian, then $\dimN(\Hc)=\hgt_X(G)$.
  For every abelian subgroup $K\le G$, $\dimN(\Hc)\ge\hgt_X(K)$, the height of $K$ acting on $\Omega$, relative to $X$.
\end{enumerate}
\end{restatable}

The proof is in \cref{app:natarajan}.

\paragraph{Idea of the proof.}
Natarajan shattering needs moves at different points that can be switched on and off independently.
In a group class, a move at $a\in A$ is an element $k_a$ of the stabilizer of the other points, and a pattern $\epsilon\in\{0,1\}^A$ is realized by the product of the moves it switches on.
Each factor fixes every point of $A$ but its own; the difficulty is that a factor may move the \emph{image} $k_bb$ of another point, which need not lie in $A$.
If the moves commute, then $k_a(k_bb)=k_b(k_ab)=k_bb$, so the product sends each point $b$ to $b$ or to $k_bb$, as the pattern prescribes.
For (i), two patterns that differ in one coordinate give a move at that point, as in \cref{thm:dictionary}(a).

\paragraph{Why it is true.}
In $S_3$ acting on $\{1,2,3\}$ with $A=\{1,2\}$, the only moves are $k_1=(1\,3)$, which fixes $2$, and $k_2=(2\,3)$, which fixes $1$.
They do not commute: $k_1k_2$ sends $2$ to $k_1(3)=1$, a third label, so switching on both moves leaves the square $\{1,3\}\times\{2,3\}$, and the moves trace the hexagon of \cref{fig:hexagon} instead.
Indeed $\dimN=1<2=\hgt$ here.
For $(\Z/2)^2$ acting on two disjoint pairs, the two swaps commute and $\dimN=\hgt=2$.

\paragraph{Takeaway.}
Under~\eqref{eq:F}, a gap between the Natarajan and DS dimensions of a group class requires stabilizer moves that do not commute.

\begin{example}[Linear maps, continued]
\label{ex:linear-natarajan}
Let $G=\mathrm{GL}(d,q)$ act on $\mathbb{F}_q^d$ as in \cref{ex:linear}.
If $q\ge3$, let $K$ be the group of diagonal matrices with respect to a basis $e_1,\dots,e_d$.
The elements of $K$ that fix $e_j$ for $j\neq i$ are the matrices $\mathrm{diag}(1,\dots,\lambda,\dots,1)$, which move $e_i$ when $\lambda\neq1$.
So $\{e_1,\dots,e_d\}$ is independent for the action of $K$ (each $e_i$ is moved by an element of $K$ that fixes the others), and \cref{thm:natarajan}(iii) gives $\dimN=d=\dimDS$.
Over $\mathbb{F}_2$ the diagonal group is trivial, and the Natarajan dimension can drop: for $\mathrm{GL}(3,2)$ acting on the $7$ nonzero vectors of $\mathbb{F}_2^3$, exhaustive computation gives $\dimN=2<3=\hgt$.
\end{example}

\begin{remark}[The abelian lower bound is not an equality]
\label{rem:abelian-strict}
Let $S_3$ act on $\{0,1,2\}$ naturally and on $\{3,4\}$ through the sign, so that odd permutations swap $3$ and $4$.
The sequence $(0,3)$ is N-shattered with $f_0=(0,3)$ and $f_1=(1,4)$: the four patterns are realized by $\mathrm{id}$, by $(1\,2)$, by $(0\,1\,2)$ and by $(0\,1)$.
So $\dimN=\hgt=2$.
Every abelian subgroup is cyclic of order $2$ or $3$, and the stabilizer of a moved point in such a subgroup is trivial; so every abelian subgroup has height $1$.
Similarly, $\mathrm{PSL}(2,7)$ acting on the $8$ points of the projective line over $\mathbb{F}_7$ has $\hgt=3$ and $\dimN=2$, while all its abelian subgroups have height $1$ (exhaustive computation).
\end{remark}

Symmetric groups show that the factor between the two dimensions can be $2$.

\begin{restatable}[The factor $1/2$]{proposition}{PropSymmetric}
\label{prop:symmetric}
Let $\Omega$ be finite and nonempty with $|\Omega|=n$, and let $G\le\Sym(\Omega)$.
\begin{enumerate}[label=(\alph*)]
\item For every nonempty $X\subseteq\Omega$, $\dimN(\HG{G}{X})\le\lfloor n/2\rfloor$; and $\hgt(G)\le n-1$.
\item For $G=S_n$ acting on $[n]$ with $n\ge2$: $\hgt(S_n)=n-1$ and $\dimN(\HG{S_n}{[n]})=\lfloor n/2\rfloor$.
\item \emph{(Direct products.)}
  If $G_1\le\Sym(\Omega_1)$ and $G_2\le\Sym(\Omega_2)$ act on disjoint sets and $G=G_1\times G_2$ acts on $\Omega_1\sqcup\Omega_2$ componentwise, then $\hgt(G)=\hgt(G_1)+\hgt(G_2)$ and $\dimN(\HG{G}{\Omega_1\sqcup\Omega_2})=\dimN(\HG{G_1}{\Omega_1})+\dimN(\HG{G_2}{\Omega_2})$.
\item For every $k\ge1$, $G=S_3^k$ acting on $3k$ points has $\hgt(G)=2k$ and $\dimN(\HG{G}{\Omega})=k$.
\end{enumerate}
Hence $\inf\{\dimN(\HG{G}{\Omega})/\hgt(G) : \hgt(G)\ge1\}\le1/2$, and the ratio $1/2$ is attained at every even height.
\end{restatable}

The proof is in \cref{app:natarajan}.
For $S_n$ the lower bound $\lfloor n/2\rfloor$ comes from the abelian subgroup generated by $\lfloor n/2\rfloor$ disjoint transpositions, and the upper bound from counting labels: an N-shattered sequence of length $s\ge2$ uses $2s$ distinct labels.
The lower bound also follows from \citet[Lemma~16]{daniely2015multiclass}: in a class of Natarajan dimension $d$ that is closed under permutations of the labels, every hypothesis takes at most $2d+1$ values, and every permutation of $[n]$ takes $n$.

\paragraph{A conjecture.}
On the full domain, we conjecture that the symmetric group is the worst case.

\begin{conjecture}[Natarajan dimension is at least half the DS dimension]
\label{conj:half}
For every finite permutation group $G\le\Sym(\Omega)$ acting on its full domain $X=\Omega$,
\[
  \dimN(\HG{G}{\Omega})\ge\lceil\hgt(G)/2\rceil .
\]
\end{conjecture}

The following facts support the conjecture and delimit it.
\begin{itemize}
\item \emph{The constant is best possible.} By \cref{prop:symmetric}(d), $S_3^k$ attains $\dimN=\hgt/2$ at every even height.
\item \emph{Partial results.} By \cref{thm:natarajan}(iii) the conjecture holds for abelian groups and, more generally, whenever some abelian subgroup has height at least $\lceil\hgt(G)/2\rceil$; this covers $S_n$ (\cref{prop:symmetric}(b)) and $\mathrm{GL}(d,q)$ with $q\ge3$ (\cref{ex:linear-natarajan}).
  Abelian subgroups alone cannot prove it: for $\mathrm{PSL}(2,7)$ on $8$ points all abelian subgroups have height $1$, while $\dimN=2=\lceil3/2\rceil$ (\cref{rem:abelian-strict}).
\item \emph{Exhaustive computation.}
  For a list of permutation groups fixed before the run (\cref{app:natarajan}), we computed $\hgt$ and $\dimN$ exactly for every group reached within the time limit, all of degree at most $8$; every one of them satisfies $\lceil\hgt/2\rceil\le\dimN\le\hgt$, and the lower bound is attained by the symmetric groups, $S_3\times S_3$, $S_3\wr S_2$ and the dihedral groups $D_3$, $D_5$, $D_6$, $D_7$.
  A second, larger sweep by an independent script covered $7{,}292$ actions with heights up to $6$ and found no violation; its largest tight instance is a quotient of order $1920$ of the affine Coxeter group of type $\tilde A_4$, acting on $80$ points, with $\hgt=5$ and $\dimN=3$.
  This is evidence, not proof.
\item \emph{The full domain is necessary.}
  The coset complexes of \citet{brukhim2022characterization} are group classes restricted to a domain that is not invariant under the group.
  For a group $F$ with subgroups $H_1,\dots,H_d$, let $F$ act on the disjoint union $\Omega$ of the coset spaces $F/H_i$; the restriction of $\HG{F}{\Omega}$ to the $d$ base cosets $H_1,\dots,H_d$ is the coset complex $\{(gH_1,\dots,gH_d) : g\in F\}$.
  Their hyperbolic examples have Natarajan dimension $1$ on this domain, while the height of the domain is $d$, because the coset complex is a $d$-dimensional pseudo-cube (\cref{rem:dictionary}(b)).
  On the full domain the picture changes.
  For a small instance of the same construction, a quotient of order $54$ of the affine Coxeter group of type $\tilde A_2$ with $d=3$ and $27$ cosets, we computed height $3$ and Natarajan dimension $1$ on the three base cosets, but Natarajan dimension $2=\lceil3/2\rceil$ on all $27$ cosets.
\end{itemize}

\subsection{Partial results toward the conjecture}
\label{sec:natarajan-partial}

The commuting-moves construction of \cref{thm:natarajan}(ii) has an ordered generalization, in which a move only has to fix the images produced by the \emph{earlier} moves.

\begin{restatable}[Towers of moves]{lemma}{LemTower}
\label{lem:tower}
Let $G\le\Sym(\Omega)$, and let $x_1,\dots,x_t\in\Omega$ be distinct and $k_1,\dots,k_t\in G$ satisfy
\textup{(T1)} $k_jx_j\neq x_j$ for every $j$;
\textup{(T2)} $k_jx_i=x_i$ for all $i\neq j$;
\textup{(T3)} $k_j(k_ix_i)=k_ix_i$ for all $i<j$.
Then $(x_1,\dots,x_t)$ is N-shattered by $\HG{G}{X}$ for every $X\supseteq\{x_1,\dots,x_t\}$, with $f_0(x_i)=x_i$ and $f_1(x_i)=k_ix_i$.
\end{restatable}

Pairwise commuting moves satisfying (T1)--(T2) satisfy (T3), so \cref{lem:tower} contains \cref{thm:natarajan}(ii).
When the closure $\cl$ behaves like a linear span, towers of length $\lceil\hgt/2\rceil$ always exist.

\begin{restatable}[Matroidal closures]{theorem}{ThmMatroid}
\label{thm:matroid}
Let $\Omega$ be finite and $G\le\Sym(\Omega)$.
Suppose that the closure $\cl(S)=\Fix(\stab{G}{S})$ satisfies the exchange property: for all $S\subseteq\Omega$ and $p,q\in\Omega$, if $q\in\cl(S\cup\{p\})\setminus\cl(S)$, then $p\in\cl(S\cup\{q\})$.
Then $\dimN(\HG{G}{\Omega})\ge\lceil\hgt(G)/2\rceil$.
The constant $1/2$ cannot be improved even within this class: $S_{2k+1}$ acting on $2k+1$ points has matroidal closure, $\hgt=2k$ and $\dimN=k$.
\end{restatable}

\paragraph{Idea of the proof.}
A closure operator that is extensive, monotone, idempotent and satisfies exchange is the closure operator of a matroid \citep[Section~1.4]{oxley2011matroid}, and the independent sets of that matroid are exactly the independent sets of \cref{sec:prelim}.
We build a tower greedily from a maximum independent set $A$.
At step $j$ we take a point $x_j$ of the current independent remainder $R_j$ and a move $k_j$ that fixes all earlier points, all earlier images, and $R_j\setminus\{x_j\}$.
Adding the two new points $x_j$ and $k_jx_j$ to the fixed set raises the rank by at most $2$, so by exchange the remainder loses at most two points.
The tower therefore has length at least $\lceil\hgt/2\rceil$, and \cref{lem:tower} turns it into a Natarajan cube.

\paragraph{Takeaway.}
Matroidal closures satisfy the conjecture, with the constant $1/2$ tight; every open case has a closure that is not a matroid closure.
The theorem covers $S_n$, the linear groups $\mathrm{GL}(d,q)$ and the affine groups $\mathrm{AGL}(d,q)$ over finite fields (whose closures are linear and affine spans, by \cref{ex:linear}), and direct products of such groups.
For $\mathrm{GL}(d,2)$, where the diagonal argument of \cref{ex:linear-natarajan} is unavailable, it gives $\dimN\ge\lceil d/2\rceil$.
Closures that are not matroids exist already in small degree (for instance, a cyclic group of order $4$ acting on orbits of sizes $2$ and $4$), and the order-$54$ quotient of the affine Coxeter group of type $\tilde A_2$ mentioned above is another; the conjecture still holds in every such example we computed.

For the first nontrivial case of the conjecture, Natarajan dimension at least $2$, there is an exact criterion in terms of two point stabilizers.

\begin{restatable}[Natarajan squares]{proposition}{PropSquare}
\label{prop:square}
Let $G\le\Sym(\Omega)$, let $x\neq y$ be points of $\Omega$, and let $P=G_x$ and $Q=G_y$.
Then $(x,y)$ is N-shattered by $\HG{G}{\Omega}$ if and only if $PQ\cap QP\supsetneq P\cup Q$.
Consequently:
\begin{enumerate}[label=(\alph*)]
\item if $h\in G_v$ moves $u$ and $\stab{G}{\{u,hu\}}\not\subseteq G_v$, then $(u,v)$ is N-shattered;
\item if $G$ is finite, $\{x,y\}$ is independent and $PQ=QP$, then $(x,y)$ is N-shattered;
\item if some independent triple $\{a,b,c\}$ has $b\in G_c\,a$, then $\dimN(\HG{G}{\Omega})\ge2$; in particular, every $2$-transitive group of height at least $3$ has Natarajan dimension at least $2$.
\end{enumerate}
\end{restatable}

So \cref{conj:half} holds for every $2$-transitive group of height at most $4$.
Part (b) shows that permutable point stabilizers already suffice, which weakens the commutation hypothesis of \cref{thm:natarajan}(ii) to a statement about two subgroups.
Proofs are in \cref{app:natarajan}.

\section{Solvability and the Natarajan dimension of Galois actions}
\label{sec:solvable}

\Cref{thm:galois} identifies every dimension of the Galois class $\Hc=\HG{\Gamma}{L}$ with an invariant of the abstract group $\Gamma$, except the Natarajan dimension, for which it gives only the abelian lower bound~(vi).
For the natural action of $S_n$ on $n$ points the Natarajan dimension is $\lfloor n/2\rfloor$, about half the DS dimension $n-1$ (\cref{sec:natarajan}).
Does the Galois action of $S_n$ show the same gap?
For $n\le4$ it does not, since $S_n$ is then solvable and solvable groups never show a gap; for $n\ge5$ it does, although the gap is smaller (\cref{thm:abelruffini}).

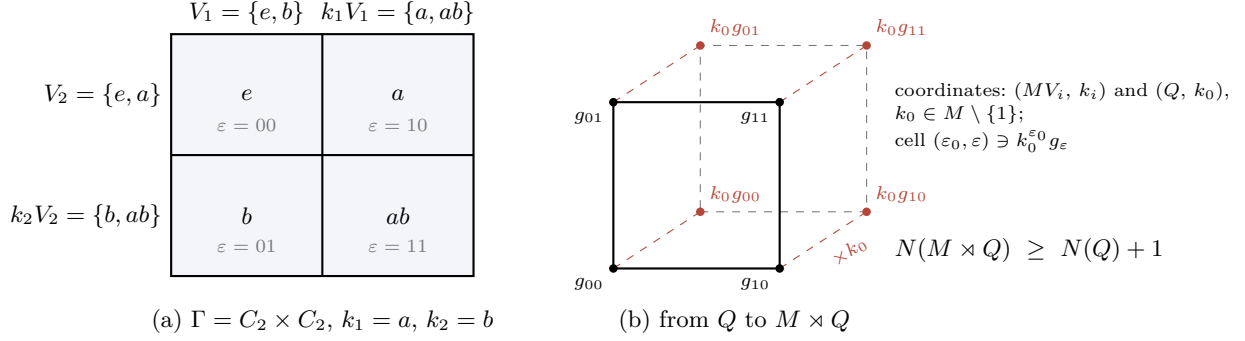
\begin{figure}[t]
\centering
\begin{tikzpicture}[font=\small]
\begin{scope}
\fill[FigBlue!6] (0,0) rectangle (4,3.2);
\draw[thick] (0,0) rectangle (4,3.2);
\draw[thick] (2,0) -- (2,3.2);  \draw[thick] (0,1.6) -- (4,1.6);
\node at (1,2.4) {$e$};   \node[FigGray,font=\scriptsize] at (1,2.0) {$\varepsilon=00$};
\node at (3,2.4) {$a$};   \node[FigGray,font=\scriptsize] at (3,2.0) {$\varepsilon=10$};
\node at (1,0.8) {$b$};   \node[FigGray,font=\scriptsize] at (1,0.4) {$\varepsilon=01$};
\node at (3,0.8) {$ab$};  \node[FigGray,font=\scriptsize] at (3,0.4) {$\varepsilon=11$};
\node[above] at (1,3.2) {$V_1=\{e,b\}$};
\node[above] at (3,3.2) {$k_1V_1=\{a,ab\}$};
\node[left] at (0,2.4) {$V_2=\{e,a\}$};
\node[left] at (0,0.8) {$k_2V_2=\{b,ab\}$};
\node at (2,-0.6) {(a) $\Gamma=C_2\times C_2$, $k_1=a$, $k_2=b$};
\end{scope}
\begin{scope}[shift={(5.85,0.1)}]
\coordinate (p00) at (0,0);   \coordinate (p10) at (2.2,0);
\coordinate (p01) at (0,2.2); \coordinate (p11) at (2.2,2.2);
\coordinate (o) at (1.15,0.75);
\foreach \x in {p00,p10,p01,p11} { \coordinate (q\x) at ($(\x)+(o)$); }
\draw[FigGray,dashed] (qp00) -- (qp10) -- (qp11) -- (qp01) -- cycle;
\foreach \x in {p00,p10,p01,p11} { \draw[FigRed,dashed] (\x) -- (q\x); }
\draw[thick] (p00) -- (p10) -- (p11) -- (p01) -- cycle;
\foreach \x/\l in {p00/{g_{00}},p10/{g_{10}},p01/{g_{01}},p11/{g_{11}}} {
  \fill (\x) circle (1.6pt); \node[below left,font=\scriptsize] at (\x) {$\l$}; }
\foreach \x/\l in {p00/{k_0g_{00}},p10/{k_0g_{10}},p01/{k_0g_{01}},p11/{k_0g_{11}}} {
  \fill[FigRed] (q\x) circle (1.6pt); \node[above right,font=\scriptsize,FigRed] at (q\x) {$\l$}; }
\node[FigRed,font=\scriptsize,rotate=33] at ($(p10)!0.5!(qp10)+(0.35,-0.15)$) {$\times k_0$};
\node at (1.6,-0.7) {(b) from $Q$ to $M\rtimes Q$};
\node[align=left,font=\scriptsize,anchor=west] at (3.6,2.0) {coordinates: $(MV_i,\,k_i)$ and $(Q,\,k_0)$,\\ $k_0\in M\setminus\{1\}$;\\[2pt] cell $(\varepsilon_0,\varepsilon)\ni k_0^{\varepsilon_0}g_\varepsilon$};
\node[font=\small,anchor=west] at (3.6,0.25) {$N(M\rtimes Q)\ \ge\ N(Q)+1$};
\end{scope}
\end{tikzpicture}
\caption{Coset cubes.
(a)~A coset cube of dimension~$2$ in $C_2\times C_2=\{e,a,b,ab\}$: the cosets of $V_1$ (columns) and of $V_2$ (rows) cut the group into four nonempty cells $k_1^{\varepsilon_1}V_1\cap k_2^{\varepsilon_2}V_2$.
(b)~The lifting step of \cref{thm:cosetcube}(iii).
Let $(V_i,k_i)$ be a cube in a complement $Q$ of $M$ (front face; the cell $\varepsilon$ contains $g_\varepsilon\in Q$).
The coordinates $(MV_i,k_i)$ together with $(Q,k_0)$, where $k_0\in M\setminus\{1\}$, form a cube in $M\rtimes Q$ of one more dimension.
The back face consists of the translates $k_0g_\varepsilon$, which lie in $k_0Q$ and, since $M$ is normal, in every $k_i^{\varepsilon_i}MV_i$.
Inducting down a chief series gives \cref{thm:cosetcube}(iv).}
\label{fig:cosetcube}
\end{figure}

\paragraph{Coset cubes.}
For a finite group $\Gamma$, call a family of subgroups $V_1,\dots,V_t\le\Gamma$ together with elements $k_i\in\Gamma\setminus V_i$ a \emph{coset cube} of dimension $t$ if
\[
\bigcap_{i=1}^t k_i^{\varepsilon_i}V_i\neq\emptyset\qquad\text{for every }\varepsilon\in\{0,1\}^t,
\]
where $k^0V=V$ and $k^1V=kV$.
If $V_i=\Gamma_{x_i}$ is the stabilizer of a point $x_i$, the intersection indexed by $\varepsilon$ is the set of elements that send each $x_i$ to $x_i$ or to $k_ix_i$ as $\varepsilon$ prescribes; this is why coset cubes measure Natarajan shattering (\cref{thm:cosetcube}(i)).
\Cref{fig:cosetcube}(a) shows a coset cube of dimension~$2$.
The \emph{coset-cube number} $N(\Gamma)$ is the largest dimension of a coset cube.
It is finite: the $2^t$ intersections, the \emph{cells} of the cube, are pairwise disjoint, so $2^t\le|\Gamma|$.
The \emph{unit cells} are those whose pattern $\varepsilon$ has exactly one coordinate equal to~$1$.

\paragraph{Chief series.}
A \emph{chief series} of a finite group $\Delta$ is a chain $1=\Delta_0<\Delta_1<\dots<\Delta_r=\Delta$ of normal subgroups of $\Delta$ with no normal subgroup of $\Delta$ strictly between consecutive terms; its \emph{chief factors} are the quotients $\Delta_i/\Delta_{i-1}$.
A \emph{complement} of a normal subgroup $M$ of $\Gamma$ is a subgroup $Q$ with $MQ=\Gamma$ and $M\cap Q=1$; then $\Gamma$ is the semidirect product $M\rtimes Q$.
The factor $\Delta_i/\Delta_{i-1}$ is \emph{complemented} if some subgroup $C$ with $\Delta_{i-1}\le C\le\Delta$ satisfies $C\Delta_i=\Delta$ and $C\cap\Delta_i=\Delta_{i-1}$, that is, if $C/\Delta_{i-1}$ is a complement of $\Delta_i/\Delta_{i-1}$ in $\Delta/\Delta_{i-1}$.
For $S_4$, all three factors of the chief series $1<V_4<A_4<S_4$ are complemented, where $V_4=\{1,(1\,2)(3\,4),(1\,3)(2\,4),(1\,4)(2\,3)\}$ is the Klein four-group (\cref{fig:quartic}).

\begin{restatable}[Natarajan dimension as a group invariant]{theorem}{ThmCosetCube}
\label{thm:cosetcube}
Let $L/K$ be a finite Galois extension of a countable field $K$, with group $\Gamma$, and $\Hc=\HG{\Gamma}{L}$.
\begin{enumerate}[label=(\roman*)]
\item $\dimN(\Hc)=N(\Gamma)$.
\item $N(\Delta)\le N(\Gamma)$ for every subgroup $\Delta\le\Gamma$, and $N(\Gamma/M)\le N(\Gamma)$ for every normal subgroup $M$.
\item If a normal subgroup $M\neq1$ has a complement $Q$, then $N(\Gamma)\ge N(Q)+1$.
\item For every subgroup $\Delta\le\Gamma$ and every chief series of $\Delta$, $\dimN(\Hc)$ is at least the number of complemented factors of the series.
\end{enumerate}
\end{restatable}

\begin{restatable}[Solvability and the Natarajan dimension]{theorem}{ThmSolvable}
\label{thm:solvable}
In the setting of \cref{thm:cosetcube}:
\begin{enumerate}[label=(\alph*)]
\item if $\Gamma$ is solvable, then $\dimN(\Hc)=\dimDS(\Hc)=\mu(\Gamma)$;
\item if $\Gamma\cong A_5$, then $\dimN(\Hc)=2<3=\dimDS(\Hc)$; if $\Gamma\cong S_5$, then $\dimN(\Hc)=3<4=\dimDS(\Hc)$;
\item if $\Gamma\cong S_n$ with $n\ge1$, then $\lfloor 3n/4\rfloor\le\dimN(\Hc)\le n-1=\dimDS(\Hc)$;
\item if $X_0\subseteq L$ is finite and every subgroup of $\Gamma$ is the stabilizer of a point of $X_0$, then $\HG{\Gamma}{X_0}$ has finitely many labels, $\dimDS(\HG{\Gamma}{X_0})=\mu(\Gamma)$ and $\dimN(\HG{\Gamma}{X_0})=N(\Gamma)$, so (a)--(c) hold verbatim for $\HG{\Gamma}{X_0}$.
\end{enumerate}
\end{restatable}

The upper bounds in (b) follow from a short case analysis (\cref{lem:quintic}), confirmed by exhaustive computer search.

So solvable Galois groups are \emph{Natarajan-saturated}: the Natarajan and DS dimensions coincide, as they do for abelian groups.
The groups of the quintic are not.
The implication goes one way only: some non-solvable groups are saturated as well (\cref{prop:psl28}).
All these statements concern the class on the whole field~$L$, or on a finite domain on which every subgroup of $\Gamma$ occurs as a stabilizer.
On the roots of a polynomial alone the picture is different: there $S_3$ already has Natarajan dimension~$1$ and DS dimension~$2$ (\cref{sec:natarajan}); \cref{ex:cubic} contrasts the two domains for the splitting field of $x^3-2$.
Part~(c) improves the lower bound $\lfloor n/2\rfloor$ of \cref{cor:galois-values}(a); in the Galois action the Natarajan dimension of $S_n$ is at least three quarters of the DS dimension, not one half.

\begin{figure}[t]
\centering
\begin{tikzpicture}[font=\small,
  g/.style={draw, rounded corners=2pt, inner sep=3pt, fill=white},
  box/.style={draw, rounded corners=3pt, inner sep=4pt, line width=1pt, fill=white, align=center},
  ar/.style={-{Stealth[length=5pt]}, line width=0.9pt}]
\node[g] (S4) at (0,4.5) {$S_4$};
\node[g] (A4) at (0,3.0) {$A_4$};
\node[g] (V4) at (0,1.5) {$V_4$};
\node[g] (one) at (0,0) {$1$};
\draw[line width=1.3pt] (S4) -- (A4) -- (V4) -- (one);
\node[font=\scriptsize,FigGray] at (0,-0.6) {chief series of $S_4$};
\node[anchor=west,font=\scriptsize] at (0.25,3.75) {$S_4/A_4\cong C_2$};
\node[anchor=west,font=\scriptsize] at (0.25,2.25) {$A_4/V_4\cong C_3$};
\node[anchor=west,font=\scriptsize] at (0.25,0.75) {$V_4\cong C_2^2$};
\node[box,draw=FigBlue] (c3) at (4.2,3.75) {complement $A_4$};
\node[box,draw=FigOrange] (c2) at (4.2,2.25) {complement $D_4$};
\node[box,draw=FigGreen] (c1) at (4.2,0.75) {complement $S_3$};
\draw[ar,FigBlue] (2.0,3.75) -- (c3); \draw[ar,FigOrange] (2.0,2.25) -- (c2); \draw[ar,FigGreen] (2.0,0.75) -- (c1);
\node[box,draw=FigBlue,anchor=west] (q3) at (7.4,3.75) {$\sqrt{\mathrm{disc}}=\prod_{i<j}(x_i-x_j)$\\[-1pt] {\scriptsize degree $2$ over $K$}};
\node[box,draw=FigOrange,anchor=west] (q2) at (7.4,2.25) {$\theta=x_1x_2+x_3x_4$\\[-1pt] {\scriptsize root of the resolvent cubic, degree $3$}};
\node[box,draw=FigGreen,anchor=west] (q1) at (7.4,0.75) {$x_4$\\[-1pt] {\scriptsize a root, degree $4$}};
\draw[ar,FigBlue] (c3) -- node[above,font=\scriptsize]{stabilizer of} (q3);
\draw[ar,FigOrange] (c2) -- node[above,font=\scriptsize]{stabilizer of} (q2);
\draw[ar,FigGreen] (c1) -- node[above,font=\scriptsize]{stabilizer of} (q1);
\node[font=\scriptsize,FigGray] at (4.2,-0.15) {$C\,\Delta_i=\Delta$,\ \ $C\cap\Delta_i=\Delta_{i-1}$};
\node[font=\scriptsize,FigGray,anchor=west] at (7.4,-0.15) {quantities in the classical solution};
\node[align=center] at (5.6,-1.1) {three complemented factors $\Rightarrow$ $\{x_4,\theta,\sqrt{\mathrm{disc}}\}$ is Natarajan-shattered:\ \ $\dimN=3=\mu(S_4)=\dimDS$};
\end{tikzpicture}
\caption{The quartic.
Each factor of the chief series $1<V_4<A_4<S_4$, where $V_4$ is the Klein four-group, is complemented, in the sense of \cref{sec:solvable}, by the subgroup $C=S_3$, $D_4$ (dihedral of order $8$) or $A_4$, and each complement is the stabilizer of a quantity in the classical solution of the quartic by radicals.
Each complemented factor contributes one coordinate of a coset cube (\cref{thm:cosetcube}(iii),(iv)), so the three quantities are Natarajan-shattered.
For the groups of the quintic the two dimensions separate (\cref{thm:solvable}(b)); every subgroup of $A_5$ has at most two complemented chief factors, fewer than $\mu(A_5)=3$.}
\label{fig:quartic}
\end{figure}

\paragraph{Why it is true: the quartic.}
Let $\Gamma=S_4$ act on the roots $x_1,\dots,x_4$ of a generic quartic, so that $L=\Q(x_1,\dots,x_4)$ is Galois over the field of symmetric rational functions with group $S_4$ \citep[Corollary~5.39]{milne2022fields}.
Take three points of $L$: the root $x_4$, the root $\theta=x_1x_2+x_3x_4$ of the resolvent cubic, and $\sqrt{\mathrm{disc}}=\prod_{i<j}(x_i-x_j)$.
These are three quantities that appear in the classical solution of the quartic by radicals (\cref{ex:quartic}).
Their stabilizers are $S_3$, the dihedral group $D_4=\langle(1\,2),(3\,4),(1\,3)(2\,4)\rangle$ of order~$8$ and $A_4$ (we write $D_n$ for the dihedral group of order $2n$).
Let the second witness $f_1$ send $x_4\mapsto x_3$, $\theta\mapsto x_2x_3+x_1x_4$ and $\sqrt{\mathrm{disc}}\mapsto-\sqrt{\mathrm{disc}}$, which are the images under $(1\,2)(3\,4)$, $(1\,2\,3)$ and $(1\,2)$.
The first witness, $f_0$, fixes all three points.
Every one of the $2^3$ patterns is realized by some permutation, so the three points are Natarajan-shattered, and $\dimN=3=\mu(S_4)=\dimDS$.
The three moves come from the chief series $1<V_4<A_4<S_4$: $V_4$ is complemented by $S_3$, $A_4/V_4$ by $D_4/V_4$, and $S_4/A_4$ by $A_4/A_4$.
Each complemented chief factor contributes one coordinate of the cube (\cref{fig:quartic}).

\paragraph{Idea of the proofs of \cref{thm:cosetcube,thm:solvable}.}
Part~(i) of \cref{thm:cosetcube} is the observation that Natarajan shattering by a group is a statement about cosets of stabilizers: after translating the witness $f_0$ to the identity, the pattern $\varepsilon$ is realized exactly by the elements of $\bigcap_i k_i^{\varepsilon_i}\Gamma_{x_i}$.
By \cref{fact:realization} every subgroup is a stabilizer, so the Galois class shatters every coset cube.
The heart is the lifting step~(iii).
If $\Gamma=M\rtimes Q$ and $(V_i,k_i)$ is a cube in $Q$, then the subgroups $MV_i$ with the same $k_i$, together with $V_0=Q$ and any $k_0\in M\setminus\{1\}$, form a cube in $\Gamma$.
The pattern $(\varepsilon_0,\varepsilon)$ is realized by $k_0^{\varepsilon_0}g$, where $g$ realizes $\varepsilon$ in $Q$ (\cref{fig:cosetcube}(b)).
Indeed $Mk_i^{\varepsilon_i}V_i$ absorbs the factor $k_0^{\varepsilon_0}\in M$, and $Q\cap MV_i=V_i$ by Dedekind's rule ($MV\cap Q=(M\cap Q)V$ for $V\le Q$; \cref{app:background}).
Inducting down a chief series gives~(iv).
For~(a), an independent set $S$ of size $\mu(\Gamma)$ is a minimal generating set of $\Delta=\langle S\rangle$.
A short induction on a chief series (\cref{lem:mingen}, the inequality half of a theorem of \citet[Theorem~2]{lucchini2013largest}) shows that a solvable $\Delta$ has no minimal generating set larger than its number of complemented chief factors.
So the chief-factor bound reaches $\mu(\Gamma)$, which is also the upper bound $\dimDS=\mu$.
For $A_5$ the argument breaks: $A_5$ is simple, its only chief factor is itself, and every subgroup of $A_5$ has at most two complemented chief factors, while $\mu(A_5)=3$.
A case analysis of the subgroups of $A_5$ shows that no coset cube of dimension~$3$ exists (\cref{lem:quintic}).

\paragraph{The general equation.}
The general equation of degree~$n$ has Galois group $S_n$ over the field generated by its coefficients \citep[Theorem~5.40]{milne2022fields}.
In characteristic zero it is therefore solvable by radicals if and only if $S_n$ is solvable \citep[Theorem~5.34]{milne2022fields}, that is, if and only if $n\le4$ \citep[Aside~4.39 and Chapter~6]{milne2025group}.
For $n\ge5$, \cref{thm:solvable}(c) leaves room between $\lfloor3n/4\rfloor$ and $n-1$.
The upper end is never attained.

\begin{restatable}[An Abel--Ruffini threshold for the Natarajan dimension]{theorem}{ThmAbelRuffini}
\label{thm:abelruffini}
\begin{enumerate}[label=(\alph*)]
\item Let $n\ge5$. There are no subgroups $V_1,\dots,V_{n-1}\le S_n$ and elements $k_i\in S_n\setminus V_i$ such that $\bigcap_{i}k_i^{\varepsilon_i}V_i\neq\emptyset$ for every $\varepsilon\in\{0,1\}^{n-1}$ with at most two coordinates equal to~$1$. In particular $N(S_n)\le n-2$.
\item $N(S_n)=n-2$ for $5\le n\le8$.
\item Let $L/K$ be a finite Galois extension of a countable field $K$ with group $\Gamma\cong S_n$, and $\Hc=\HG{\Gamma}{L}$. Then $\dimN(\Hc)=\dimDS(\Hc)$ if and only if $n\le4$; for $n\ge5$, $\dimN(\Hc)\le n-2<n-1=\dimDS(\Hc)$. The same holds for $\HG{\Gamma}{X_0}$, with $X_0$ as in \cref{thm:solvable}(d).
\end{enumerate}
\end{restatable}

So the Galois class of the general equation of degree~$n$ is Natarajan-saturated exactly in the degrees in which, in characteristic zero, the equation is solvable by radicals.
For symmetric groups, the one-way implication of \cref{thm:solvable}(a) is therefore an equivalence.
The coincidence concerns the threshold, not the mechanism: our proof of the gap for $n\ge5$ uses the structure of the independent generating sets of $S_n$, not non-solvability.
Together with \cref{thm:solvable}(c), part~(b) gives $N(S_n)=\lfloor3n/4\rfloor$ for every $n\le8$.

\paragraph{Idea of the proof of \cref{thm:abelruffini}.}
The proof uses only the cells with at most two coordinates equal to~$1$, and it has two ingredients.
First, the unit cells give an independent subset of $S_n$ of size $n-1$ (\cref{lem:unitcells}): an element $g_i$ of the unit cell of coordinate~$i$ lies in every $V_j$ with $j\neq i$ but not in $V_i$, so $g_i\notin\langle g_j : j\neq i\rangle\le V_i$.
By theorems of \citet{whiston2000maximal} and \citet{cameron2002independent}, such a set is built from the transpositions of a tree on $[n]$ (for $n=6$, after an automorphism of $S_6$): it consists of one edge transposition $s$ and, for every other edge $t$, one of $t$, $st$, $ts$.
Second, the cells with two coordinates equal to~$1$ give an orthogonality relation (\cref{lem:orthogonality}).
If $i\neq j$, a set $A\subseteq[n]$ is invariant under $V_i$ and $A'$ under $V_j$, and $g_i,g_j$ are the elements chosen from the unit cells, then the vectors $\mathbf 1_{g_iA}-\mathbf 1_A$ and $\mathbf 1_{g_jA'}-\mathbf 1_{A'}$ are orthogonal.
For an edge $t\neq s$ that does not split $[n]$ into two halves, \cref{lem:young} lets us take for $A$ the side of $t$ away from $s$, and the vector becomes $\mathbf 1_y-\mathbf 1_x$, where $x$ is the endpoint of $t$ away from $s$ and $y$ is its other endpoint or, when $t$ meets $s$, an endpoint of~$s$.
Orthogonality then forbids an edge that does not meet $s$ together with the next edge on its path to~$s$, and two edges that leave the same vertex away from~$s$; it allows at most two edges that meet~$s$ (\cref{fig:orthogonality}).
Once $n\ge5$, every tree violates these rules (a balanced edge, which splits $[n]$ into two halves and of which there is at most one, is excluded by a short count).
Both cited theorems rely on the classification of finite simple groups, and the cases $n=5,6$ of the second rest on a computation reported by \citet{cameron2002independent}; we have also confirmed part~(a) for $n=5,6$ directly, by an exhaustive computer search that uses neither theorem.

\begin{corollary}
\label{cor:a5smallest}
Among finite groups of order at most $60$, all of which occur as Galois groups of finite extensions of countable fields, $A_5$ is, up to isomorphism, the only one whose Galois classes are not Natarajan-saturated.
\end{corollary}

\begin{proof}
Every group of order less than $60$ is solvable, and a non-solvable group of order $60$ is simple and hence isomorphic to $A_5$ \citep[5.19 and Chapter~6]{milne2025group}.
Every finite group $G$ occurs: it acts faithfully on $\Q(x_1,\dots,x_n)$ by permuting the indeterminates, for some $n$, and Artin's theorem \citep[Corollary~3.11]{milne2022fields} makes this field Galois over the fixed field of $G$ with group $G$.
Now apply \cref{thm:solvable}(a) and~(b).
\end{proof}

\begin{restatable}[Saturation without solvability]{proposition}{PropPSL}
\label{prop:psl28}
The simple group $\mathrm{PSL}(2,8)$ has $N=\mu=3$.
Consequently, for every finite Galois extension $L/K$ of a countable field with group $\Gamma\cong\mathrm{PSL}(2,8)$, the class $\HG{\Gamma}{L}$ has $\dimN=\dimDS=3$.
\end{restatable}

\paragraph{What it means for learning.}
For a Galois class, $\hgt=\mu(\Gamma)$ and $\dimN=N(\Gamma)$, so the factor $\hgt/\dimN$ by which the agnostic ERM bound of \cref{prop:agnostic} can exceed the lower bound of \citet[Theorem~5]{daniely2015multiclass} is $\mu(\Gamma)/N(\Gamma)$.
It is $1$ for every solvable group, so on solvable Galois classes every agnostic ERM is optimal among deterministic learners up to absolute constants, wherever that lower bound holds (see the discussion after \cref{prop:agnostic}).
It is also $1$ for $\mathrm{PSL}(2,8)$ (\cref{prop:psl28}); it is $3/2$ for $A_5$, $4/3$ for $S_5$, at most $4/3$ for every $S_n$ with $n\ge2$, and larger than $1$ for every $S_n$ with $n\ge5$.
These are constant factors, and the constants of \citet{daniely2015multiclass} are not explicit, so \cref{thm:solvable,thm:abelruffini} do not separate learning rates; they determine, for the groups above, whether the ratio in this certificate of optimality equals $1$.

\paragraph{Beyond chief factors.}
The chief-factor bound of \cref{thm:cosetcube}(iv) is attained by every solvable group, by $\mathrm{PSL}(2,8)$ (see the proof of \cref{prop:psl28}), and by each of the roughly $130$ groups of order at most $240$ for which we computed both quantities by exhaustive computer search (\cref{tab:cosetcube} lists some of them).
It is not attained in general.

\begin{restatable}[The chief-factor bound is not tight]{proposition}{PropPGaL}
\label{prop:pgal}
Let $\Gamma=\mathrm{P\Gamma L}(2,8)=\mathrm{PSL}(2,8)\rtimes\langle\varphi\rangle$, of order $1512$, where $\varphi$ is induced by $x\mapsto x^2$ on $\mathbb F_8$.
Then $N(\Gamma)\ge4$, whereas every chief series of every subgroup of $\Gamma$ has at most $3$ complemented factors; the last statement is computer-assisted.
\end{restatable}

The four coordinates of the cube are the stabilizers of three points of the projective line and the subgroup $\mathrm{PSL}(2,8)$ itself.
The cube exists because $\mathrm{PSL}(2,8)$ is sharply $3$-transitive and the stabilizer of three points is a complement to it (\cref{app:solvable}).
So the Natarajan dimension of a Galois action is not determined by the chief-factor count, and characterizing it is open (\cref{sec:discussion}).

\paragraph{Takeaway.}
For Galois classes the Natarajan dimension is the coset-cube number, which is at least the number of complemented factors of any chief series of any subgroup: it equals the DS dimension for every solvable group, and for the general equation of degree~$n$ it equals the DS dimension exactly in the degrees $n\le4$ in which, in characteristic zero, the equation is solvable by radicals.

\section{Monodromy of pseudo-cubes}
\label{sec:monodromy}

The previous sections explain DS and Natarajan shattering of group classes through their symmetries.
Does a pseudo-cube that comes from an arbitrary class carry a group structure as well?
We show that it does whenever neighbors are unique.

\begin{definition}[Two-fibered pseudo-cubes and their monodromy]
\label{def:monodromy}
A $d$-dimensional pseudo-cube $B\subseteq Y^d$ is \emph{two-fibered} if every $b\in B$ has exactly one $i$-neighbor in $B$ for each $i\in[d]$; we denote it by $s_i(b)$.
The \emph{monodromy group} of $B$ is the permutation group $M=M(B)=\langle s_1,\dots,s_d\rangle\le\Sym(B)$.
Its \emph{parabolic subgroups} are $K_i=\langle s_j : j\neq i\rangle$, for $i\in[d]$.
We write $\pi_j\colon B\to Y$, $b\mapsto b_j$, for the coordinate maps.
\end{definition}

Since each $s_i$ is an involution (\cref{thm:monodromy}(a)), the free Coxeter group $\Wd=\langle s_1,\dots,s_d\mid s_1^2,\dots,s_d^2\rangle$ acts on $B$, and $M$ is the image of this action.
For a group $F$ and subgroups $H_1,\dots,H_d$, the \emph{coset complex} of \citet{brukhim2022characterization} is
\[
  \mathcal{C}_F(H_1,\dots,H_d)=\{(gH_1,\dots,gH_d) : g\in F\}.
\]
When $F$ is finite, it is a $d$-dimensional pseudo-cube as soon as $\big(\bigcap_{j\neq i}H_j\big)\setminus H_i\neq\emptyset$ for every $i$: an element $z$ of this set turns the word $(gH_1,\dots,gH_d)$ into its $i$-neighbor $(gzH_1,\dots,gzH_d)$.
This is condition~(1) of \citet[Theorem~45]{brukhim2022characterization}, and the argument is the one in the proof of their Proposition~46.
We regard a pseudo-cube $B\subseteq Y^d$ as a hypothesis class on the domain $[d]$, so that $\dimN(B)$ is defined.
Right cosets $K_jm$ are subsets of $M$, while orbits $K_j\cdot b$ are subsets of $B$; we keep the two notations apart.

\begin{restatable}[Monodromy of two-fibered pseudo-cubes]{theorem}{ThmMonodromy}
\label{thm:monodromy}
Let $B\subseteq Y^d$ be a two-fibered $d$-dimensional pseudo-cube with monodromy group $M$ and parabolic subgroups $K_1,\dots,K_d$.
\begin{enumerate}[label=(\alph*)]
\item Each $s_i$ is a fixed-point-free involution of $B$.
  Each $\pi_j$ is constant on $K_j$-orbits, and $\pi_j(s_jb)\neq\pi_j(b)$.
  In particular $s_jb\notin K_j\cdot b$ \textup{(separation)}, and $s_j\notin K_j$.
\item \emph{(Orbit labels.)}
  The map $\iota\colon b\mapsto(K_1\cdot b,\dots,K_d\cdot b)$ is injective.
  Its image $B^\sharp$ is a $d$-dimensional pseudo-cube in which $\iota(s_ib)$ is an $i$-neighbor of $\iota(b)$, and $B$ is the image of $B^\sharp$ under the coordinatewise maps $K_j\cdot b\mapsto\pi_j(b)$.
\item \emph{(Galois closure.)}
  Assume that $M$ is transitive on $B$, and fix $b_0\in B$.
  Let $C=\{(K_1m,\dots,K_dm) : m\in M\}$, with right cosets.
  Then $C$ is a $d$-dimensional pseudo-cube; after the relabeling $m\mapsto m^{-1}$ it is the coset complex $\mathcal{C}_M(K_1,\dots,K_d)$.
  The coordinatewise maps $K_jm\mapsto K_j\cdot(mb_0)$ are well defined and map $C$ onto $B^\sharp$, sending the word $(K_jm)_{j}$ to $\iota(mb_0)$.
  If moreover $M$ acts regularly on $B$, these maps are bijections, and $C\cong B^\sharp$.
\item \emph{(Abelian monodromy.)}
  $M$ is abelian if and only if $B$ is a disjoint union of sets of the form $\prod_{j=1}^d\{u_j,v_j\}$ with $u_j\neq v_j$ (full Natarajan cubes).
  In that case these sets are exactly the $M$-orbits, and $M$ is an elementary abelian $2$-group.
\item \emph{(Local version.)}
  If $T\subseteq[d]$, $b\in B$, and the involutions $s_i$ $(i\in T)$ pairwise commute on the orbit $O=\langle s_i : i\in T\rangle\cdot b$, then the projection of $O$ to the coordinates in $T$ is the Natarajan cube $\prod_{j\in T}\{\pi_j(b),\pi_j(s_jb)\}$.
  Hence $\dimN(B)\ge|T|$.
\end{enumerate}
\end{restatable}

The proof is in \cref{app:monodromy}.
In~(c), $C\cong B^\sharp$ means that coordinatewise bijections of the labels carry $C$ onto $B^\sharp$; such maps send $i$-neighbors to $i$-neighbors.

\begin{remark}[Transitivity is no restriction]
\label{rem:orbits}
Each $M$-orbit $O\subseteq B$ is closed under every $s_i$, so every word of $O$ has its unique $i$-neighbor inside $O$.
Thus $O$ is itself a two-fibered $d$-dimensional pseudo-cube, whose monodromy group is the restriction of $M$ to $O$ and is transitive.
Since $B$ is the disjoint union of its $M$-orbits, part (c) describes every two-fibered pseudo-cube, one orbit at a time.
\end{remark}

\begin{remark}[The regular case]
\label{rem:regular}
By (b), $\iota$ is a bijection from $B$ onto $B^\sharp$ that maps $i$-neighbors to $i$-neighbors, and $B$ is obtained from $B^\sharp$ by a coordinatewise relabeling.
So when $M$ acts regularly, (c) and (b) together show that the map $(K_jm)_j\mapsto mb_0$ is a bijection from $C$ onto $B$ that maps $i$-neighbors to $i$-neighbors for every $i$, and that is induced by coordinatewise relabelings.
In this sense $B$ is the coset complex of its monodromy group, up to coordinatewise relabeling of the labels.
\end{remark}

\paragraph{Relation to chamber systems.}
The neighbor graph of a two-fibered pseudo-cube, with each edge colored by the coordinate in which its endpoints differ, is a properly $d$-edge-colored graph whose color classes are perfect matchings.
Each connected component is a $(d-1)$-complex in the sense of \citet[Section~5]{wilson2012maniplexes}, with $s_i$ in the role of his $r_{i-1}$ (such complexes are essentially combinatorial maps), equivalently a connected thin chamber system over $[d]$ \citep{tits1981local,calderbank2016parabolic}.
In that language our monodromy group is the connection group, and the orbits of the parabolic subgroup $K_i$ are the $(i-1)$-faces.
Part~(c) of \cref{thm:monodromy} is the pseudo-cube form of a standard fact: the structure group of a connected thin chamber system acts transitively, so the orbit map of a chamber is a surjective morphism from the Cayley chamber system of that group, and it is an isomorphism exactly when the group acts freely \citep[Proposition~4.15]{calderbank2016parabolic}; for abstract polytopes, compare \citet{hartley1999all} and \citet[Theorem~2E11]{mcmullen2002abstract}.
What is specific to pseudo-cubes is that the labels are coordinatewise images of the orbit labels (parts~(a) and~(b)), and the relation between abelian monodromy and Natarajan cubes (parts~(d) and~(e)).

\paragraph{Idea of the proof.}
Coordinate $j$ of a word changes only under $s_j$, so it is constant on the orbits of the parabolic subgroup $K_j$; these orbits form a canonical labeling $\iota$ that refines the given one.
For a transitive action, the orbits of $K_j$ are the images of the right cosets $K_jm$ under $m\mapsto mb_0$, which is the coset complex.
If $M$ is abelian, it is generated by commuting involutions, so the orbit of $b$ is $\{s^\epsilon b : \epsilon\in\{0,1\}^d\}$ with $s^\epsilon=\prod_is_i^{\epsilon_i}$.
The separation property $s_jb\notin K_j\cdot b$ forces these $2^d$ words to be distinct and their coordinates to vary independently, which is a full Natarajan cube.
The only delicate point is (e), where the involutions commute on one orbit only, so all products must be evaluated as maps on that orbit.

\paragraph{Why it is true.}
Take $d=2$.
Starting from a word and applying $s_1,s_2,s_1,\dots$ alternately traces a cycle whose edges alternate between $1$-neighbors and $2$-neighbors.
Its length is $2k$ with $k\ge2$, because $s_1b$ and $s_2b$ differ from $b$ in different coordinates.
On such a cycle $M$ acts as the dihedral group of order $2k$, which is abelian exactly when $k=2$.
So (d) says that a two-dimensional two-fibered pseudo-cube is a union of Natarajan squares if and only if all its cycles are $4$-cycles.
The hexagon of \cref{sec:galois}, the restriction of the $S_3$ class to $(1,2)$, is a $6$-cycle: $s_1$ replaces the first label by the third unused one, $s_2$ does the same for the second label, and $s_1s_2$ rotates the hexagon by two steps.
Its monodromy group is dihedral of order $6$, isomorphic to $S_3$, and acts regularly; so by (c) and \cref{rem:regular} the hexagon is the coset complex of $S_3$ with respect to two subgroups of order $2$.
Since $M$ is not abelian, by (d) the hexagon is not a union of Natarajan squares; in fact it contains none (\cref{prop:symmetric}(a)).
\Cref{fig:hexagon} contrasts it with a Natarajan square, on which $s_1$ and $s_2$ commute.

\paragraph{Takeaway.}
Every two-fibered pseudo-cube is a group action in disguise: each of its monodromy orbits is a coordinatewise image of the coset complex of its own monodromy group, and the pseudo-cube is a union of Natarajan cubes exactly when its monodromy group is abelian.

\paragraph{Interpretation.}
The analogy with Galois theory is literal.
The coset complex $C$ plays the role of a Galois closure: it is built from the monodromy group alone, and $B$ (more precisely, each $M$-orbit of $B$, by \cref{rem:orbits}) is recovered from it by passing from cosets to orbits, much as an intermediate field is recovered from a Galois extension as the fixed field of a subgroup.
When $M$ acts regularly no word is lost: the orbit map is a neighbor-preserving bijection from $C$ onto $B$, induced by coordinatewise maps of the labels (\cref{rem:regular}).
In this sense the coset complexes of \citet{brukhim2022characterization} are the Galois objects among two-fibered pseudo-cubes.
At the other extreme sit the two-fibered pseudo-cubes with abelian monodromy, which are exactly the two-fibered disjoint unions of Natarajan cubes (\cref{thm:monodromy}(d)).
Part (e) is the local form of this statement and parallels \cref{thm:natarajan}(ii): commuting neighbor moves on part of a pseudo-cube produce Natarajan shattering there.
In particular, if a two-fibered $d$-dimensional pseudo-cube $B$ has $\dimN(B)<d$, then on every $M$-orbit some two of the involutions $s_1,\dots,s_d$ fail to commute.

\section{Discussion and open problems}
\label{sec:discussion}

For the problem of learning a symmetry with finite orbits on a countable domain, the Galois correspondence between point sets and pointwise stabilizers determines the realizable PAC, online, teaching and list dimensions exactly, and it makes every consistent learner rate-optimal among deterministic learners on nondegenerate classes.
For Galois groups the Natarajan dimension is the coset-cube number; it equals the DS dimension for every solvable group and is strictly smaller for the groups of the quintic (\cref{thm:cosetcube,thm:solvable}).
For general group classes the Natarajan dimension, and with it the exact agnostic sample complexity, remains undetermined; the agnostic ERM bound of \cref{prop:agnostic}, the lower bound of \citet{daniely2015multiclass} and the improper bound of \citet{pabbaraju2026optimal} on finite domains are collected after \cref{prop:agnostic}.
The following problems are open.

\paragraph{1. The factor one half.}
Prove or refute \cref{conj:half}: $\dimN(\HG{G}{\Omega})\ge\lceil\hgt(G)/2\rceil$ for every finite permutation group acting on its full domain.
Two obstacles are known.
First, the closure operator of a permutation group need not be a matroid closure; a cyclic group of order $4$ acting on orbits of sizes $2$ and $4$ is an example (\cref{sec:natarajan-partial}).
Without the exchange property, the induction behind \cref{thm:matroid}, which uses up at most two points of an independent set for each coordinate of the cube, breaks down: adding a single point can pull many independent points into the closure (a point with trivial stabilizer has all of $\Omega$ in its closure).
Second, commuting moves are not enough: in $\mathrm{PSL}(2,7)$ acting on $8$ points every abelian subgroup has height $1$, yet $\dimN=2$ (by exhaustive computation; \cref{rem:abelian-strict}).
A proof must therefore produce Natarajan squares from moves that do not commute in $G$.

\paragraph{2. The coset-cube number.}
Which group invariant is $N(\Gamma)$, and hence the Natarajan dimension of every Galois class?
It is at least the number of complemented factors of every chief series of every subgroup (\cref{thm:cosetcube}(iv)).
This bound is exact for solvable groups, for $\mathrm{PSL}(2,8)$, and for each of the roughly $130$ groups of order at most $240$ for which we computed it by exhaustive computer search, but it fails for $\mathrm{P\Gamma L}(2,8)$ (\cref{prop:pgal}).
There, sharp $3$-transitivity produces an extra coordinate.
A natural refinement replaces each complemented chief factor $M$ (a minimal normal subgroup of $\Delta/\Delta_{i-1}$, for a chief series $(\Delta_i)$ of a subgroup $\Delta\le\Gamma$; by \cref{thm:cosetcube}(ii) we may work in that quotient, which we again call $\Gamma$) by the largest dimension of a cube all of whose cells meet every coset of $M$.
Such a cube can be stacked on a cube of the quotient by $M$,\footnote{Lift the cube of $\Gamma/M$ to $\Gamma$ by taking preimages of its subgroups, as in the proof of \cref{thm:cosetcube}(ii); each cell of the lifted cube is a nonempty union of cosets of $M$, and each cell of the given cube meets every coset of $M$, so every pattern of the combined cube is realized.} which gives a lower bound at least as good as the chief-factor bound, because for a complemented $M$ with complement $Q$ and any $k\in M\setminus\{1\}$, $(Q,k)$ is such a cube of dimension~$1$ ($Q$ and $kQ$ meet every coset of $M$, as $QM=\Gamma$).
For $S_n$ the question includes the exact value of $N(S_n)$.
It lies between $\lfloor3n/4\rfloor$ and $n-2$ for $n\ge5$, and it equals $\lfloor3n/4\rfloor$ for every $n\le8$ (\cref{thm:solvable,thm:abelruffini}).
Is $N(S_n)=\lfloor3n/4\rfloor$ for all $n$?
If so, the ratio of the DS and Natarajan dimensions of the generic Galois class of degree~$n$ tends to $4/3$.
More generally, is $\mu(\Gamma)/N(\Gamma)$ bounded over all finite groups?
Realize $\Gamma$ as the Galois group of a finite Galois extension $L/K$ of a countable field (every finite group is such a Galois group; see the proof of \cref{cor:a5smallest}), let $X_0\subseteq L$ be as in \cref{thm:solvable}(d), and let $\Omega_0$ be the union of the $\Gamma$-orbits of its points.
Then $\Gamma$ acts faithfully on the finite set $\Omega_0$, every subgroup is a stabilizer of a point of $\Omega_0$, and \cref{thm:solvable}(d) gives $\dimDS=\mu(\Gamma)$ and $\dimN=N(\Gamma)$ on $\Omega_0$; so \cref{conj:half} would give $N(\Gamma)\ge\lceil\mu(\Gamma)/2\rceil$.
Is even $N(\Gamma)\ge\lceil2\mu(\Gamma)/3\rceil$?
The group $A_5$ shows that $2/3$ could not be improved.
Finally, which non-solvable groups are saturated?
$\mathrm{PSL}(2,8)$ is (\cref{prop:psl28}); $A_5$ and $S_5$ are not (\cref{thm:solvable}(b)), and neither is $\mathrm{PSL}(2,7)$ (by exhaustive computer search, \cref{tab:cosetcube}).

\paragraph{3. A group-theoretic formula for the Natarajan dimension of general group classes.}
We know that $\max_U\hgt_X(U)\le\dimN(\HG{G}{X})\le\hgt_X(G)$, where $U$ ranges over abelian subgroups, and both inequalities can be strict (\cref{rem:abelian-strict} and \cref{prop:symmetric}).
Is there an invariant of the permutation group that equals $\dimN(\HG{G}{X})$?
\Cref{thm:monodromy}(e) suggests that the relevant condition is commutation of moves on one orbit of words rather than commutation in $G$.

\paragraph{4. DS dimension without finite orbits.}
Without~\eqref{eq:F} we still have $\dimDS(\HG{G}{X})\le\hgt_X(G)$ (\cref{rem:dictionary}(b)), and the inequality can be strict (\cref{rem:F-necessary}).
There is also a lower bound.
Let $U\le G$ and let $A\subseteq X$ be finite and independent for the action of $U$ (\cref{def:invariants} with $U$ in place of $G$), with $Ua$ finite for every $a\in A$.
Applying \cref{thm:dictionary}(a) to $U$ with domain $A$ shows that $A$ is DS-shattered by $\HG{U}{A}$, hence by $\HG{G}{X}$; so $\dimDS(\HG{G}{X})\ge|A|$.
Is $\dimDS(\HG{G}{X})$ equal to the supremum of $|A|$ over such pairs $(U,A)$?
A positive answer would characterize PAC learnability for group classes with infinite orbits, such as discrete groups of isometries.

\paragraph{5. Pseudo-cubes with larger panels.}
\Cref{thm:monodromy} and \cref{rem:orbits} show that every monodromy orbit of a two-fibered pseudo-cube is a coordinatewise image of a coset complex, that is, of a group class restricted to a set of base cosets (\cref{sec:natarajan}).
In a general pseudo-cube an $i$-panel may have more than two elements, and the single involution $s_i$ must be replaced, for instance, by the group of all permutations of $B$ that preserve every $i$-panel.
Does every pseudo-cube arise, up to coordinatewise relabeling, from a group class in this way?
Do the analogues of \cref{thm:monodromy}(c) and~(d) hold for $k$-pseudo-cubes, which govern list learning?

\paragraph{6. Local symmetry of version spaces.}
Group classes are homogeneous: all version spaces of samples on the same points are translates of each other (\cref{lem:coset}).
For a general class one can still ask for a ``local Galois group'' attached to a version space, for example the monodromy group of the pseudo-cubes it contains.
Is there such a notion for which the gap between the Natarajan and DS dimensions of a general class is controlled by non-commutativity, as it is for two-fibered pseudo-cubes?

\paragraph{7. Computational complexity of agnostic learning.}
For $S_n$, agnostic ERM is an assignment problem and is solvable in polynomial time (\cref{rem:computation}).
For $\mathrm{GL}(d,2)$ acting on $\mathbb{F}_2^d$, agnostic ERM asks for an invertible matrix $A$ that satisfies as many of the equations $Ax_i=y_i$ as possible, a problem of the same flavor as decoding noisy linear equations over $\mathbb{F}_2$.
Is agnostic ERM for this class computationally hard?
More generally, which group classes admit efficient agnostic learners?
We do not know a reduction in either direction.

\paragraph{AI Disclosure.}
This work was carried out with extensive use of AI tools.
The author posed the research question, the connection between Galois theory and multiclass learnability.
Claude models (Anthropic), run through Claude Code in a multi-agent workflow, were used to search the literature, to propose and refine conjectures, to write and check proofs, to run the computer searches reported in the paper, and to draft and revise the text.
In this workflow, separate model instances tried to refute each proof, and every citation was checked against a retrieved copy of its source.
The author reviewed the results and the manuscript and takes full responsibility for the content of the paper.

\bibliographystyle{plainnat}   %
\bibliography{refs}

\clearpage
\appendix
\section{A crash course on Galois theory}
\label{app:background}

This appendix collects the field theory and group theory used in the paper, for readers from learning theory.
All of it is standard; we follow \citet{milne2022fields} for fields and \citet{milne2025group} for groups.
The external theorems on which our proofs rely are restated, with precise sources, in \cref{app:external}.

\subsection{Field extensions}

A \emph{field extension} $L/K$ is a field $L$ that contains $K$ as a subfield.
Then $L$ is a vector space over $K$, and its dimension $[L:K]$ is the \emph{degree} of the extension, which is \emph{finite} if $[L:K]<\infty$.
Degrees multiply in towers: if $K\subseteq F\subseteq L$, then $[L:K]=[L:F]\,[F:K]$.
For a subset $A\subseteq L$, the field $K(A)$ is the smallest subfield of $L$ that contains $K$ and $A$, the field \emph{generated} by $A$ over $K$; we write $K(\alpha_1,\dots,\alpha_n)$ for $K(\{\alpha_1,\dots,\alpha_n\})$.

An element $\alpha\in L$ is \emph{algebraic} over $K$ if $f(\alpha)=0$ for some nonzero $f\in K[x]$.
The monic polynomial of least degree with this property is the \emph{minimal polynomial} of $\alpha$; it is irreducible, and $[K(\alpha):K]$ equals its degree.
Every element of a finite extension is algebraic.
A polynomial is \emph{separable} if it has no repeated root in a field over which it splits into linear factors, and an algebraic extension $L/K$ is \emph{separable} if the minimal polynomial of every element of $L$ is separable.
In characteristic zero, and over finite fields, every finite extension is separable.

A \emph{$K$-automorphism} of $L$ is a field automorphism $\sigma$ of $L$ with $\sigma c=c$ for every $c\in K$; these form the group $\mathrm{Aut}(L/K)$.
For a group $U$ of automorphisms of $L$, the \emph{fixed field} is $L^U=\{\alpha\in L : \sigma\alpha=\alpha\text{ for all }\sigma\in U\}$.
In the notation of \cref{sec:prelim-groups}, with $\Omega=L$ and $G=\mathrm{Aut}(L/K)$, we have $L^U=\Fix(U)$.
Moreover, an automorphism fixes $A$ pointwise if and only if it fixes $K(A)$ pointwise, because its fixed points form a subfield; so $\stab{G}{A}=\mathrm{Aut}(L/K(A))$.

\subsection{Galois extensions and the fundamental theorem}

A finite extension $L/K$ is \emph{Galois} if the only elements of $L$ fixed by every $K$-automorphism are those of $K$, that is, $L^{\mathrm{Aut}(L/K)}=K$.
Equivalently, $L$ is the splitting field over $K$ of a separable polynomial: the field generated over $K$ by its roots \citep[Theorem~3.10]{milne2022fields}.
The group $\Gamma=\Gal(L/K)\defeq\mathrm{Aut}(L/K)$ is then the \emph{Galois group}.
The fundamental theorem (\cref{ext:ftgt}) says that $F\mapsto\Gal(L/F)$ is an inclusion-reversing bijection from the intermediate fields $K\subseteq F\subseteq L$ onto the subgroups of $\Gamma$, with inverse $U\mapsto L^U$, and that $[L:L^U]=|U|$; in particular $|\Gamma|=[L:K]$.
In the language of \cref{sec:prelim-groups}, the Galois connection between subsets of $L$ and subgroups of $\Gamma$ has as closed sets exactly the intermediate fields, and $\cl(A)=K(A)$ (\cref{thm:galois}(i)).

\begin{example}[A biquadratic field]
\label{ex:biquadratic}
Let $K=\Q$ and $L=\Q(\sqrt2,\sqrt3)$, of degree $4$.
The group $\Gamma=\{1,\sigma,\tau,\sigma\tau\}\cong(\Z/2)^2$ consists of the identity, $\sigma\colon\sqrt2\mapsto-\sqrt2,\ \sqrt3\mapsto\sqrt3$, $\tau\colon\sqrt2\mapsto\sqrt2,\ \sqrt3\mapsto-\sqrt3$, and $\sigma\tau$.
Its five subgroups $1$, $\langle\sigma\rangle$, $\langle\tau\rangle$, $\langle\sigma\tau\rangle$, $\Gamma$ correspond to the five intermediate fields $L$, $\Q(\sqrt3)$, $\Q(\sqrt2)$, $\Q(\sqrt6)$, $\Q$.
This is the field of \cref{fig:hero}(a) and of the example in \cref{sec:fields}.
More generally, for distinct primes $p_1,\dots,p_r$ the field $\Q(\sqrt{p_1},\dots,\sqrt{p_r})$ has degree $2^r$ over $\Q$ \citep[Theorem~5.30 and Example~5.31(b)]{milne2022fields}, and its Galois group is $(\Z/2)^r$.
\end{example}

\begin{example}[Finite fields]
\label{ex:finitefields}
For a prime power $q$ and $n\ge1$, the extension $\mathbb F_{q^n}/\mathbb F_q$ is Galois, and its group is cyclic of order $n$, generated by the Frobenius map $\alpha\mapsto\alpha^q$.
Its intermediate fields are the fields $\mathbb F_{q^d}$ with $d\mid n$.
Learning an element of this group means learning an exponent $i$ modulo $n$ from examples $(\alpha,\alpha^{q^i})$.
\end{example}

\begin{example}[A cubic]
\label{ex:cubic}
Let $L=\Q(\sqrt[3]2,\omega)$ with $\omega=e^{2\pi i/3}$, the splitting field of $x^3-2$ over $\Q$.
Then $[L:\Q]=6$, and $\Gamma\cong S_3$ acts faithfully on the three roots $\sqrt[3]2$, $\omega\sqrt[3]2$, $\omega^2\sqrt[3]2$.
The stabilizer of the root $\sqrt[3]2$ has order $2$ and fixed field $\Q(\sqrt[3]2)$, of degree $3$; the subgroup $A_3$ has fixed field $\Q(\omega)=\Q(\sqrt{-3})$, of degree $2$.
On the three roots alone, the class is $S_3$ acting on three points, which has Natarajan dimension $1$ and DS dimension $2$ (\cref{prop:symmetric}(b), \cref{thm:dictionary}(a), and the hexagon of \cref{fig:hexagon}); on the whole field $L$, $\dimN=\dimDS=\mu(S_3)=2$ (\cref{thm:solvable}(a)).
\end{example}

\subsection{Conjugates, labels and primitive elements}

The following computation explains \cref{thm:galois}(v).

\begin{fact}[Conjugates]
\label{fact:conjugates}
Let $L/K$ be a finite Galois extension with group $\Gamma$, let $\alpha\in L$, and let $f$ be the minimal polynomial of $\alpha$ over $K$.
Then the orbit $\Gamma\alpha$ is the set of roots of $f$, $f$ is separable, $|\Gamma\alpha|=[K(\alpha):K]$, and $\Gamma_\alpha=\Gal(L/K(\alpha))$.
\end{fact}

\begin{proof}
Each $\sigma\in\Gamma$ maps roots of $f$ to roots of $f$, since $f$ has coefficients in $K$.
The polynomial $g=\prod_{\beta\in\Gamma\alpha}(x-\beta)$ has distinct roots, and its coefficients are fixed by every $\sigma\in\Gamma$, since $\sigma$ permutes $\Gamma\alpha$; so $g\in L^\Gamma[x]=K[x]$.
As $g(\alpha)=0$, $f$ divides $g$; so $f$ has distinct roots, and they lie in $\Gamma\alpha$.
Conversely every point of $\Gamma\alpha$ is a root of $f$.
Hence $f=g$ and $|\Gamma\alpha|=\deg f=[K(\alpha):K]$.
Finally, $\sigma$ fixes $\alpha$ if and only if it fixes $K(\alpha)$ pointwise.
\end{proof}

In particular, the minimal polynomial over $K$ of every element of $L$ is separable, so $L/K$ and every intermediate extension $F/K$ are separable; this is the separability used in \cref{fact:realization}.

Applied to the Galois extension $L/K(A)$, whose group is the pointwise stabilizer of $A$, \cref{fact:conjugates} and \cref{lem:coset}(ii) say that after a sample on $A$ labeled by $\sigma$, the labels of $\alpha$ that remain possible are the images $\sigma\beta$ of the conjugates $\beta$ of $\alpha$ over $K(A)$; there are $[K(A)(\alpha):K(A)]$ of them.
In particular the label is forced exactly when $\alpha\in K(A)$.
The primitive element theorem (\cref{ext:primitive}) gives $\theta$ with $L=K(\theta)$; then $\Gamma_\theta=\Gal(L/L)=1$, and the single example $(\theta,\sigma\theta)$ identifies $\sigma$.

\subsection{The Galois group of a polynomial, and solvability by radicals}

Let $f\in K[x]$ be separable of degree $n$, with roots $x_1,\dots,x_n$ in a splitting field $L$.
Every $\sigma\in\Gal(L/K)$ permutes the roots, and $\sigma$ is determined by this permutation because $L=K(x_1,\dots,x_n)$; so $\Gal(L/K)$ is a subgroup of $S_n$, the \emph{Galois group of $f$}.
For indeterminates $x_1,\dots,x_n$ the group is all of $S_n$ (\cref{ext:general}): the field $\Q(x_1,\dots,x_n)$ is Galois over the field of symmetric rational functions, with group $S_n$ permuting the $x_i$, and this is the Galois group of the general equation of degree $n$.

A polynomial is \emph{solvable by radicals} if its roots lie in a field obtained from $K$ by successively adjoining roots of elements, $\beta^{1/m}$.
In characteristic zero this happens exactly when its Galois group is solvable (\cref{ext:radicals}).
The group $S_n$ is solvable exactly when $n\le4$ (\cref{ext:small}), which gives the Abel--Ruffini theorem: in characteristic zero, the general equation of degree $n$ is solvable by radicals exactly when $n\le4$.

\begin{example}[The quartic]
\label{ex:quartic}
Let $L=\Q(x_1,\dots,x_4)$, $\Gamma=S_4$ and $K=L^{S_4}$.
The group $S_4$ permutes the three elements $\theta_1=x_1x_2+x_3x_4$, $\theta_2=x_1x_3+x_2x_4$ and $\theta_3=x_1x_4+x_2x_3$, which are the roots of the resolvent cubic, and the kernel of this action is $V_4=\{1,(1\,2)(3\,4),(1\,3)(2\,4),(1\,4)(2\,3)\}$.
The element $\sqrt{\mathrm{disc}}=\prod_{i<j}(x_i-x_j)$ has stabilizer $A_4$.
By the fundamental theorem, the tower
\[
K\ \subset\ K(\sqrt{\mathrm{disc}})\ \subset\ K(\theta_1,\theta_2,\theta_3)\ \subset\ L
\]
corresponds to the chief series (defined below) $S_4>A_4>V_4>1$, and its steps have degrees $2$, $3$ and $4$.
The classical solution follows this tower: a square root, then the roots of the resolvent cubic, then square roots.
The three points used in \cref{sec:solvable} and \cref{fig:quartic} are $\sqrt{\mathrm{disc}}$, $\theta_1$ and $x_4$; their stabilizers $A_4$, $D_4$ and $S_3$ are the subgroups $C$ that complement the three chief factors, whereas the tower above uses the members of the series.
\end{example}

\subsection{Group-theoretic background}

For a subset $S$ of a group, $\langle S\rangle$ is the subgroup it generates; cosets are left cosets $gU$, except in \cref{sec:monodromy}, which uses right cosets; $N\trianglelefteq G$ means that $N$ is a normal subgroup.
We write $C_n$ for the cyclic group of order $n$, $D_n$ for the dihedral group of order $2n$, $V_4\cong C_2^2$ for the Klein four-group, and $A_n\le S_n$ for the alternating group.

\paragraph{Independent and minimal generating sets.}
A subset $S$ of a group $\Delta$ is \emph{independent} if $s\notin\langle S\setminus\{s\}\rangle$ for every $s\in S$, and $\mu(\Delta)$ is the largest size of an independent subset.
A generating set of $\Delta$ is \emph{minimal} if no proper subset of it generates $\Delta$; we write $m(\Delta)$ for the largest size of a minimal generating set.
A minimal generating set is independent, and an independent set $S$ is a minimal generating set of $\langle S\rangle$; since independence of $S$ does not depend on the ambient group, $\mu(\Delta)=\max_{U\le\Delta}m(U)$.
For example, the transpositions $(1\,2),(2\,3),\dots,(n-1\;n)$ are independent in $S_n$, since removing $(i\;i+1)$ leaves a set that generates a subgroup preserving $\{1,\dots,i\}$; and $\mu(S_n)=n-1$ (\cref{ext:whiston}).
\citet{cameron2002independent} write $\mu'$ for our $\mu$ and $\mu$ for the largest size of an independent generating set; \citet{whiston2002psl} use the same convention.

\paragraph{Chief series and complements.}
A \emph{complement} of a normal subgroup $M\trianglelefteq\Gamma$ is a subgroup $Q$ with $MQ=\Gamma$ and $M\cap Q=1$; then every element of $\Gamma$ is uniquely a product $mq$, and $\Gamma$ is the semidirect product $M\rtimes Q$.
A \emph{chief series} of a finite group $\Delta$ is a chain $1=\Delta_0<\Delta_1<\dots<\Delta_r=\Delta$ of normal subgroups of $\Delta$ such that no normal subgroup of $\Delta$ lies strictly between $\Delta_{i-1}$ and $\Delta_i$; the quotients $\Delta_i/\Delta_{i-1}$ are its \emph{chief factors}.
The factor $\Delta_i/\Delta_{i-1}$ is \emph{complemented} if some subgroup $C$ with $\Delta_{i-1}\le C\le\Delta$ satisfies $C\Delta_i=\Delta$ and $C\cap\Delta_i=\Delta_{i-1}$; equivalently, $C/\Delta_{i-1}$ is a complement of $\Delta_i/\Delta_{i-1}$ in $\Delta/\Delta_{i-1}$.
For $S_4$ the chief series is $1<V_4<A_4<S_4$, with factors $C_2^2$, $C_3$ and $C_2$, complemented by $C=S_3$, $D_4$ and $A_4$ (\cref{fig:quartic}).

\paragraph{Dedekind's rule.}
If $V\le Q$ and $M$ are subgroups of a group, then $MV\cap Q=(M\cap Q)V$.
Indeed, $(M\cap Q)V\subseteq MV\cap Q$ because $V\le Q$; and if $mv\in Q$ with $m\in M$ and $v\in V$, then $m=(mv)v^{-1}\in M\cap Q$.
This is the step $Q\cap MV_i=V_i$ in the lifting argument of \cref{thm:cosetcube}(iii), where $M\cap Q=1$.

\paragraph{Solvable groups.}
A group $\Delta$ is \emph{solvable} if it has a chain of subgroups $1=U_0\trianglelefteq U_1\trianglelefteq\dots\trianglelefteq U_s=\Delta$, each normal in the next, with abelian quotients $U_j/U_{j-1}$ \citep[Chapter~6]{milne2025group}.
Subgroups and quotients of solvable groups are solvable, and an extension of a solvable group by a solvable group is solvable \citep[Proposition~6.6]{milne2025group}.
Hence a finite group $\Delta$ is solvable if and only if all its chief factors are abelian.
Indeed, if they are, a chief series is a chain as in the definition; conversely, a chief factor $\Delta_i/\Delta_{i-1}$ of a solvable $\Delta$ is solvable and is a minimal normal subgroup of $\Delta/\Delta_{i-1}$, and its derived subgroup is normal in $\Delta/\Delta_{i-1}$ and proper, hence trivial.
A chief factor that is abelian is elementary abelian, that is, isomorphic to $C_p^k$ for a prime $p$, since its elements of order dividing $p$ form a nontrivial subgroup that is normal in $\Delta/\Delta_{i-1}$.
Abelian groups, groups of prime-power order and $S_4$ are solvable; $A_5$ is simple and not solvable (\cref{ext:small}).

\subsection{A dictionary}

\Cref{tab:galoisdictionary} summarizes how the classical notions read for the class $\HG{\Gamma}{L}$ of a finite Galois extension $L/K$ of a countable field $K$, with group $\Gamma$.
\Cref{tab:dictionary} gives the corresponding dictionary for general group classes.

\begin{table}[h]
\centering
\small
\begin{tabular}{p{0.36\linewidth}p{0.31\linewidth}>{\raggedright\arraybackslash}p{0.23\linewidth}}
\toprule
Galois theory & learning & where\\
\midrule
the field $K(A)$ generated by the sample points & the points whose label is forced, $\cl(A)$ & \cref{thm:galois}(i), \cref{lem:coset}(iii)\\
the group $\Gal(L/K(A))$ & the version space is a coset of it & \cref{lem:coset}(i)\\
the conjugates $\beta$ of $\alpha$ over $K(A)$ & the possible labels $\sigma\beta$ of $\alpha$ & \cref{thm:galois}(v)\\
a primitive element $\theta$, $\Gamma_\theta=1$ & a teaching set of size one & \cref{thm:galois}(ii)\\
independent subsets of $\Gamma$, $\mu(\Gamma)$ & DS and graph dimension & \cref{thm:galois}(iii), \cref{prop:agnostic}(a)\\
longest chain of subgroups, $\ell(\Gamma)$ & Littlestone dimension & \cref{thm:galois}(iv)\\
coset cubes, $N(\Gamma)$ & Natarajan dimension & \cref{thm:cosetcube}(i)\\
$\Gamma$ solvable & implies $\dimN=\dimDS$ (not conversely) & \cref{thm:solvable}(a), \cref{prop:psl28}\\
\bottomrule
\end{tabular}
\caption{The Galois dictionary.}
\label{tab:galoisdictionary}
\end{table}

\section{External results used in the proofs}
\label{app:external}

For the reader's convenience we restate the external results on which our proofs rely, in the form in which we use them, and say where each is used.
Results that enter only remarks and comparisons \citep[for example][]{gill2021height,kelsey2021relational,hanneke2016refined,pabbaraju2026optimal} are stated where they are used.

\subsection{Fields}

\begin{fact}[{Fundamental theorem of Galois theory; \citealp[Corollary~3.13 and Theorem~3.17]{milne2022fields}}]
\label{ext:ftgt}
Let $L/K$ be a finite Galois extension with group $\Gamma$.
For every intermediate field $K\subseteq F\subseteq L$, the extension $L/F$ is Galois.
The maps $F\mapsto\Gal(L/F)$ and $U\mapsto L^U$ are mutually inverse, inclusion-reversing bijections between the intermediate fields of $L/K$ and the subgroups of $\Gamma$, and $[L:L^U]=|U|$ for every subgroup $U\le\Gamma$.
\end{fact}
\noindent That $L/F$ is Galois and that the two maps are mutually inverse bijections is also in \citet[Chapter~VI, Theorems~1.1 and~1.2]{lang2002algebra}, which is what \cref{sec:fields,app:fields} cite for these two statements.
\emph{Used in} \cref{fact:realization}, \cref{thm:galois}(i),(v) and \cref{app:background}.

\begin{fact}[{Primitive element theorem; \citealp[Theorem~5.1]{milne2022fields}}]
\label{ext:primitive}
Every finite separable extension $L/K$ is simple: $L=K(\theta)$ for some $\theta\in L$.
\end{fact}
\noindent See also \citet[Chapter~V]{lang2002algebra}, which is what \cref{fact:realization} cites.
\emph{Used in} \cref{fact:realization} and \cref{thm:galois}(ii).

\begin{fact}[{Artin's theorem; \citealp[Corollary~3.11]{milne2022fields}}]
\label{ext:artin}
Let $G$ be a finite group of automorphisms of a field $E$.
Then $E/E^G$ is a finite Galois extension of degree $|G|$, with Galois group $G$.
\end{fact}
\noindent\emph{Used in} \cref{cor:a5smallest} and \cref{thm:abelruffini}(b), to realize every finite group, and $S_n$ in particular, as a Galois group over a countable field.

\begin{fact}[{The general equation; \citealp[Corollary~5.39 and Theorem~5.40]{milne2022fields}}]
\label{ext:general}
Let $F$ be a field, let $X_1,\dots,X_n$ be indeterminates, and let $p_1,\dots,p_n$ be the elementary symmetric polynomials in them.
Then $F(X_1,\dots,X_n)$ is Galois over $F(p_1,\dots,p_n)$, with Galois group $S_n$ acting by permuting the $X_i$.
The Galois group of the general polynomial of degree $n$ is $S_n$.
\end{fact}
\noindent\emph{Used in} \cref{sec:solvable} (the general equation, and the quartic) and \cref{ex:quartic}.

\begin{fact}[{Galois's criterion; \citealp[Theorem~5.34]{milne2022fields}}]
\label{ext:radicals}
Let $F$ be a field of characteristic zero.
A polynomial $f\in F[X]$ is solvable in radicals if and only if its Galois group is solvable.
\end{fact}
\noindent\emph{Used in} \cref{sec:solvable}, to read \cref{thm:abelruffini} as a statement about equations.

\begin{fact}[{Small groups; \citealp{milne2025group}}]
\label{ext:small}
\leavevmode
\begin{enumerate}[label=(\alph*)]
\item For $n\ge5$, the group $A_n$ is simple, and $A_n$ and $S_n$ are not solvable \textup{(Theorem~4.33 and Aside~4.39)}.
\item Every group of order less than $60$ is solvable; in particular $S_n$ is solvable for $n\le4$ \textup{(Chapter~6)}.
\item A simple group of order $60$ is isomorphic to $A_5$ \textup{(5.19)}.
\item If $N\trianglelefteq G$ and both $N$ and $G/N$ are solvable, then $G$ is solvable \textup{(Proposition~6.6(b))}.
\end{enumerate}
\end{fact}
\noindent By (b) and (d), a non-solvable group of order $60$ has no proper nontrivial normal subgroup, so it is simple, and by (c) it is isomorphic to $A_5$.
\emph{Used in} \cref{thm:abelruffini}(c), \cref{cor:a5smallest} and \cref{sec:solvable}.

\subsection{Groups}

\begin{fact}[{\citealp{whiston2000maximal}; see also \citealp[Section~2]{cameron2002independent}}]
\label{ext:whiston}
Let $n\ge2$.
Every independent subset of $S_n$ has at most $n-1$ elements, and every independent subset of $S_n$ with $n-1$ elements generates $S_n$.
\end{fact}
\noindent The proof uses the classification of finite simple groups.
\emph{Used in} \cref{cor:galois-values}(a), \cref{thm:solvable}(b),(c) and the proof of \cref{thm:abelruffini}(a).

\begin{fact}[{\citealp[Theorem~2.1 and the remark after Corollary~2.4]{cameron2002independent}}]
\label{ext:cameroncara}
For a tree $T$ on $[n]$, let $S(T)$ be the set of transpositions along its edges.
Let $n\ge5$ with $n\neq6$, and let $S$ be an independent generating set of $S_n$ with $n-1$ elements.
Then there is a tree $T$ on $[n]$ such that either $S=S(T)$, or, for some $s\in S(T)$, $S=\{s\}\cup\{(st)^{\epsilon(t)} : t\in S(T)\setminus\{s\}\}$ with $\epsilon(t)\in\{1,-1\}$.
For $n=6$ the same conclusion holds after applying a suitable automorphism of $S_6$ to $S$.
\end{fact}
\noindent Theorem~2.1 of \citet{cameron2002independent} is stated for $n\ge7$; the remark after their Corollary~2.4 reports that the independent generating sets of size $n-1$ for $n\le6$ were determined by hand and by computer, and that the theorem holds for $n=5$, and for $n=6$ up to the outer automorphism of $S_6$.
\emph{Used in} the proof of \cref{thm:abelruffini}(a); we also confirmed \cref{thm:abelruffini}(a) for $n=5,6$ by an exhaustive search that uses neither this fact nor \cref{ext:whiston}.

\begin{fact}[{\citealp{cameron1989chains}}]
\label{ext:chains}
For $n\ge2$, the longest chain of subgroups of $S_n$ has length $\ell(S_n)=\lceil3n/2\rceil-b(n)-1$, where $b(n)$ is the number of ones in the binary expansion of $n$.
\end{fact}
\noindent\emph{Used in} \cref{cor:galois-values}(a).

\begin{fact}[{\citealp[Theorem~2]{lucchini2013largest}}]
\label{ext:lucchini}
Let $G$ be a finite solvable group.
The largest size of a minimal generating set of $G$ equals the number of complemented factors in any chief series of $G$.
\end{fact}
\noindent\emph{Used in} \cref{thm:solvable}(a), through the inequality ``$\le$'', which we prove directly as \cref{lem:mingen}.

\begin{fact}[{Dickson's theorem for $\mathrm{PSL}(2,8)$; \citealp[Theorem~2.1]{giudici2007maximal}}]
\label{ext:dickson}
Every maximal subgroup of $\mathrm{PSL}(2,8)$ is isomorphic to $C_2^3\rtimes C_7$ (the stabilizer of a point of the projective line over $\mathbb F_8$), to $D_7$ or to $D_9$, of orders $56$, $14$ and $18$.
\end{fact}
\noindent This is the case $q=8$ of the list of maximal subgroups of $\mathrm{PSL}(2,q)$ for $q=2^f\ge4$ in the cited theorem.
The proof of \cref{prop:psl28} shows that every subgroup of order $56$ isomorphic to $C_2^3\rtimes C_7$ is a point stabilizer.
\emph{Used in} the proof of \cref{prop:psl28}.

\begin{fact}[{\citealp[Section~2]{cameron2006designs}; \citealp[p.~3]{leemans2009pgl}}]
\label{ext:pgl}
For every prime power $q$, the group $\mathrm{PGL}(2,q)$ acts $3$-transitively on the $q+1$ points of the projective line over $\mathbb F_q$ and has order $q^3-q$, the number of ordered triples of distinct points; hence the action is sharply $3$-transitive, and a nonidentity element fixes at most two points.
If $q$ is even, every element of $\mathbb F_q^\times$ is a square, so $\mathrm{PSL}(2,q)$, the group of M\"obius maps with $ad-bc$ a nonzero square, equals $\mathrm{PGL}(2,q)$; \citet[p.~3]{leemans2009pgl} record the isomorphism.
\end{fact}
\noindent Here a M\"obius map of the projective line is a map $x\mapsto(ax+b)/(cx+d)$ with $a,b,c,d\in\mathbb F_q$ and $ad-bc\neq0$.
\emph{Used in} the proofs of \cref{prop:pgal,prop:psl28}.

\subsection{Matroids and learning}

\begin{fact}[{Closure axioms of matroids; \citealp[Section~1.4]{oxley2011matroid}}]
\label{ext:matroid}
Let $E$ be a finite set and let $\cl$ be an operator on the subsets of $E$ such that, for all $A,B\subseteq E$ and $x,y\in E$:
(CL1) $A\subseteq\cl(A)$;
(CL2) if $A\subseteq B$, then $\cl(A)\subseteq\cl(B)$;
(CL3) $\cl(\cl(A))=\cl(A)$;
(CL4) if $y\in\cl(A\cup\{x\})\setminus\cl(A)$, then $x\in\cl(A\cup\{y\})$.
Then the sets $I\subseteq E$ with $x\notin\cl(I\setminus\{x\})$ for every $x\in I$ are the independent sets of a matroid on $E$ whose closure operator is $\cl$.
\end{fact}
\noindent\emph{Used in} the proof of \cref{thm:matroid}, together with the standard properties of the rank function listed there \citep[Sections~1.3--1.4]{oxley2011matroid}.

\begin{fact}[{\citealp{brukhim2022characterization}}]
\label{ext:bcdmy}
\begin{enumerate}[label=(\alph*)]
\item \textup{(Theorem~A.)} A class $\Hc\subseteq Y^X$ is PAC learnable if and only if $\dimDS(\Hc)<\infty$.
\item Let $F$ be a finite group and $H_1,\dots,H_d\le F$ subgroups with $\big(\bigcap_{j\neq i}H_j\big)\setminus H_i\neq\emptyset$ for every $i$. Then the coset complex $\{(gH_1,\dots,gH_d) : g\in F\}$ is a $d$-dimensional pseudo-cube.
\end{enumerate}
\end{fact}
\noindent Part~(b) is not a numbered statement of \citet{brukhim2022characterization}: the condition is hypothesis~(1) of their Theorem~45, and the implication is the argument of the proof of their Proposition~46.
It takes one line: the set is finite and nonempty, and if $z\in\big(\bigcap_{j\neq i}H_j\big)\setminus H_i$, then $(gzH_1,\dots,gzH_d)$ agrees with $(gH_1,\dots,gH_d)$ exactly in the coordinates $j\neq i$.
\emph{Used in} \cref{thm:erm}(c), for classes of infinite height, and in \cref{sec:monodromy} and \cref{thm:monodromy}(c), where the condition in~(b) is identified.

\begin{fact}[{\citealp[Theorem~5]{daniely2015multiclass}}]
\label{ext:dssbds}
There is an absolute constant $C_2>0$ such that, for every class $\Hc\subseteq Y^X$ with $X$ and $Y$ countable and all $\eps,\delta\in(0,1)$, every ERM (with arbitrary tie-breaking) learns $\Hc$ agnostically from $C_2(d_G(\Hc)+\ln(1/\delta))/\eps^2$ examples, where $d_G$ is the graph dimension.
\end{fact}
\noindent The same theorem also states a lower bound $C_1(\dimN(\Hc)+\ln(1/\delta))/\eps^2$ on the agnostic sample complexity of deterministic learners.
It is stated without a range of $\eps$ and $\delta$ and cannot hold for all $\eps,\delta\in(0,1)$ (see the discussion after \cref{prop:agnostic}); no proof in this paper uses it.
\emph{Used in} \cref{prop:agnostic}(b),(c).

\begin{fact}[{\citealp{darnstadt2015optimal}; Theorem~8 of the author's preprint}]
\label{ext:darnstadt}
Let $\mathcal K$ be a family of subsets of a nonempty set $Z$ that is closed under intersections of arbitrary subfamilies and has VC dimension $d$.
For every target $K^\star\in\mathcal K$, every distribution on $Z$ and every $m>0$, with probability at least $1-\delta$ over $m$ i.i.d.\ examples labeled by $K^\star$, the closure hypothesis (the intersection of all members of $\mathcal K$ consistent with the sample, which is the smallest member of $\mathcal K$ that contains the positive examples) has error at most $\frac{48}{m}\big(d\log880+\log(24/\delta)\big)$.
\end{fact}
\noindent The constants are those of the proof of the cited theorem, whose statement gives the sample complexity $O((d+\log(1/\delta))/\eps)$.
\emph{Used in} \cref{thm:erm}(b).

\begin{fact}[{\citealp[Theorem~2.5, with $k=1$]{hanneke2024improved}}]
\label{ext:hmz}
Call $\Hc\subseteq Y^X$ \emph{nondegenerate} if there are $h_1,h_2\in\Hc$ and $x_0,x_1\in X$ with $h_1(x_0)=h_2(x_0)$ and $h_1(x_1)\neq h_2(x_1)$.
For every nondegenerate class of DS dimension $d\in\N$ and all $\eps\in(0,1/16)$ and $\delta\in(0,1/8)$,
\[
M^{\mathrm{HMZ}}_{\Hc}(\eps,\delta)\ \ge\ \frac{(d-1)\ln2+4\ln(1/\delta)}{32\,\eps},
\]
where $M^{\mathrm{HMZ}}_{\Hc}$ is the sample complexity of deterministic learners in their model, defined in \cref{app:erm}.
\end{fact}
\noindent For a group class this is the nondegeneracy of \cref{thm:erm}(c); the proof of \cref{thm:erm}(c) checks the equivalence.
\emph{Used in} \cref{thm:erm}(c),(d).

\section{Proofs for Section~\ref{sec:galois}: the Galois correspondence}
\label{app:galois}

\ThmDictionary*

\begin{proof}
For a sequence $\mathbf{a}=(a_1,\dots,a_d)$ in $X$ and $g\in G$, write $w(g)=(ga_1,\dots,ga_d)$, so that $\Hc|_{\mathbf{a}}=\{w(g) : g\in G\}$.
Let $A=\{a_1,\dots,a_d\}$.

\emph{(a), shattered implies distinct and independent.}
Let $B\subseteq\Hc|_{\mathbf{a}}$ be a $d$-dimensional pseudo-cube.
For $d=1$ there is nothing to prove about distinctness.
If $d\ge2$ and $a_i=a_j$ for some $i\neq j$, then every word of $\Hc|_{\mathbf{a}}$ has equal $i$-th and $j$-th coordinates; an $i$-neighbor $b'$ of $b\in B$ changes coordinate $i$ but not coordinate $j$, which is impossible.
For independence, fix $i$, pick $b=w(g)\in B$, and pick an $i$-neighbor $b'=w(g')\in B$.
Then $g'a_j=ga_j$ for $j\neq i$ and $g'a_i\neq ga_i$.
So $k\defeq g^{-1}g'\in\stab{G}{A\setminus\{a_i\}}$ and $ka_i\neq a_i$, that is, $a_i\notin\Fix(\stab{G}{A\setminus\{a_i\}})=\cl(A\setminus\{a_i\})$.
This direction does not use~\eqref{eq:F}.

\emph{(a), distinct and independent implies that $\Hc|_{\mathbf{a}}$ is a pseudo-cube.}
$\Hc|_{\mathbf{a}}$ is nonempty, and it is finite because it is contained in $\prod_iGa_i$, which is finite by~\eqref{eq:F}.
Let $b=w(g)$ and $i\in[d]$.
By independence there is $k\in\stab{G}{A\setminus\{a_i\}}$ with $ka_i\neq a_i$.
The word $w(gk)$ agrees with $w(g)$ at every $j\neq i$, since $gka_j=ga_j$, and differs from it at $i$, since $g$ is injective and $ka_i\neq a_i$.
So $w(gk)$ is an $i$-neighbor of $b$ in $\Hc|_{\mathbf{a}}$.

\emph{(a), the dimension.}
By the first step, every DS-shattered sequence of length $d$ yields an independent $d$-subset of $X$, so $\dimDS(\Hc)\le\hgt_X(G)$.
By the second step, every finite independent subset of $X$, listed in any order, is DS-shattered, so $\dimDS(\Hc)\ge\hgt_X(G)$.
Both sides are suprema over the same set of lengths, so they agree, including the value $+\infty$.

\emph{(b).}
Assume the entries of $\mathbf{a}$ are distinct.
The $i$-panel of $w(g)$ in $\Hc|_{\mathbf{a}}$ is $\{w(g') : g'a_j=ga_j\text{ for all }j\neq i\}=\{w(gu) : u\in\stab{G}{A\setminus\{a_i\}}\}$.
Its elements agree outside coordinate $i$ and are distinguished by their $i$-th coordinates $gua_i$; since $g$ is injective, the panel has exactly $|\stab{G}{A\setminus\{a_i\}}\,a_i|$ elements.
($\Leftarrow$) If all these orbits have at least $k+1$ points, then $\Hc|_{\mathbf{a}}$, which is finite by~\eqref{eq:F} and nonempty, is a $k$-pseudo-cube.
($\Rightarrow$) Let $B\subseteq\Hc|_{\mathbf{a}}$ be a $k$-pseudo-cube.
Every panel of $B$ has at least two elements, so $B$ is a pseudo-cube and the entries of $\mathbf{a}$ are distinct by part (a).
For $b=w(g)\in B$, the $i$-panel of $b$ in $B$ is contained in its $i$-panel in $\Hc|_{\mathbf{a}}$, so $|\stab{G}{A\setminus\{a_i\}}\,a_i|\ge k+1$.
The equality $\dimDS_k(\Hc)=\hgt^k_X(G)$ follows as in (a).

\emph{(c), $\Ldim(\Hc)\ge\irr_X(G)$.}
Let $(a_1,\dots,a_r)$ be irredundant.
Consider the complete binary tree of depth $r$ in which every node at depth $i-1$ carries the point $a_i$.
We label its edges so that every node at depth $i-1$ is reached by a path with labels $(y_1,\dots,y_{i-1})$ that is realized by some $g\in G$, that is, $y_j=ga_j$ for $j<i$.
At the root the path is empty, and every $g\in G$ realizes it.
At a node reached by a path realized by $g$, the labels of $a_i$ that are consistent with the path form the set $g(\stab{G}{\{a_1,\dots,a_{i-1}\}}\,a_i)$, by \cref{lem:coset}(ii).
This set has at least two elements because $a_i\notin\cl(\{a_1,\dots,a_{i-1}\})$.
Choose two distinct elements $gka_i$ and $gk'a_i$, with $k,k'\in\stab{G}{\{a_1,\dots,a_{i-1}\}}$, as the labels of the two edges leaving the node.
The two children are then reached by paths realized by $gk$ and by $gk'$, because $k$ and $k'$ fix $a_1,\dots,a_{i-1}$.
By induction every root-to-leaf path is realized, so the tree is shattered and $\Ldim(\Hc)\ge r$.

\emph{(c), $\Ldim(\Hc)\le\irr_X(G)$.}
Let a shattered mistake tree of depth $r$ be given, and fix a root-to-leaf path with points $z_1,\dots,z_r$, whose edge labels are realized by some $g\in G$, that is, they are $gz_1,\dots,gz_r$.
At the $i$-th node, each of the two out-edge labels, together with the labels of the path prefix, is realized by some hypothesis (take any leaf below that edge).
The prefix labels are those of $g$, so by \cref{lem:coset}(ii) the labels of $z_i$ that are consistent with the prefix form the set $g(\stab{G}{\{z_1,\dots,z_{i-1}\}}\,z_i)$.
This set contains the two distinct out-edge labels, so it has at least two elements, and $z_i\notin\cl(\{z_1,\dots,z_{i-1}\})$.
Hence $(z_1,\dots,z_r)$ is irredundant and $r\le\irr_X(G)$.

\emph{(d).}
Let $h=g|_X$.
By \cref{lem:coset}(i), $\VS(S_g(A))=\{g'|_X : g'\in g\,\stab{G}{A}\}$.
This set equals $\{h\}$ if and only if $(gk)|_X=g|_X$ for every $k\in\stab{G}{A}$.
Since $g$ is injective, this holds if and only if every $k\in\stab{G}{A}$ fixes every $x\in X$, that is, $X\subseteq\Fix(\stab{G}{A})=\cl(A)$.
The condition does not involve $g$, so all hypotheses have the same teaching sets, and $\TD(h)=\min\{|A| : A\text{ a relative base}\}=\bsz_X(G)$, which is $\infty$ if there is no relative base.
A teaching set is inclusion-minimal if and only if it is an inclusion-minimal relative base, so the supremum of the sizes of the inclusion-minimal teaching sets is $\Bsz_X(G)$.
\end{proof}

\section{Proofs for Section~\ref{sec:erm}: optimal ERM}
\label{app:erm}

\subsection{The class of closed sets and the optimality of ERM}

The high-probability bound uses the family of closed subsets of the domain as a binary concept class.

\begin{lemma}[Closed sets]
\label{lem:closed-sets}
Let $G\le\Sym(\Omega)$, let $X\subseteq\Omega$ be nonempty, and let $\mathcal{K}=\{C\subseteq X : \cl_X(C)=C\}$ be the family of closed subsets of $X$.
\begin{enumerate}[label=(\alph*)]
\item $X\in\mathcal{K}$, and $\mathcal{K}$ is closed under arbitrary intersections.
\item For $B\subseteq X$, the set $\cl_X(B)$ is the smallest member of $\mathcal{K}$ that contains $B$.
\item A finite $T\subseteq X$ is shattered by $\mathcal{K}$ if and only if $T$ is independent.
  Hence the VC dimension of $\mathcal{K}$ is $\hgt_X(G)$.
\end{enumerate}
\end{lemma}

\begin{proof}
(a) $\cl_X(X)=X$ because $\cl$ is extensive (\cref{lem:closure}(a)).
If $C_\alpha\in\mathcal{K}$ for all $\alpha$ and $C=\bigcap_\alpha C_\alpha$, then by monotonicity $\cl(C)\cap X\subseteq\cl(C_\alpha)\cap X=C_\alpha$ for every $\alpha$, so $\cl(C)\cap X\subseteq C\subseteq\cl(C)\cap X$.

(b) The set $\cl_X(B)$ contains $B$.
It is closed: by monotonicity and idempotence (\cref{lem:closure}(b,c)), $\cl(\cl(B)\cap X)\cap X\subseteq\cl(\cl(B))\cap X=\cl(B)\cap X$, and the reverse inclusion is extensivity.
If $C\in\mathcal{K}$ contains $B$, then $\cl(B)\cap X\subseteq\cl(C)\cap X=C$.

(c) ($\Leftarrow$) Let $T$ be independent and $U\subseteq T$.
If $t\in T\setminus U$, then $\cl(U)\subseteq\cl(T\setminus\{t\})$, which does not contain $t$.
So $\cl_X(U)\cap T=U$, and the closed set $\cl_X(U)$ cuts out $U$ from $T$.
($\Rightarrow$) Let $T$ be shattered and $t\in T$.
Pick $C\in\mathcal{K}$ with $C\cap T=T\setminus\{t\}$.
By (b), $\cl_X(T\setminus\{t\})\subseteq C$, and $t\in X\setminus C$, so $t\notin\cl(T\setminus\{t\})$.
The VC dimension is the supremum of the sizes of shattered finite sets, which is $\hgt_X(G)$.
\end{proof}

\ThmErm*

\begin{proof}
Throughout, $h^\star=g|_X$ is the target.
In (a), (b) and (d), $\Omega$ is countable, so all subsets of $\Omega$ and of its finite powers are measurable, and Fubini's theorem applies to the product measure $D^{m+1}$.

\emph{(a).}
Let $x_1,\dots,x_{m+1}$ be i.i.d.\ from $D$, and write $T_j=\{x_i : i\le m+1,\ i\neq j\}$, as a set.

Step 1 (consistent implies correct on the closure).
By \cref{lem:coset}(iii), every $f\in\VS(S_m)$ satisfies $f(x)=h^\star(x)$ for every $x\in\cl(\{x_1,\dots,x_m\})\cap X$.
Hence $\mathbf{1}[\mathcal{A}(S_m)(x_{m+1})\neq h^\star(x_{m+1})]\le\mathbf{1}[x_{m+1}\notin\cl(T_{m+1})]$.

Step 2 (Fubini).
$\E[\er_D(\mathcal{A}(S_m))]=\P[\mathcal{A}(S_m)(x_{m+1})\neq h^\star(x_{m+1})]\le\P[x_{m+1}\notin\cl(T_{m+1})]$.

Step 3 (exchangeability).
The joint law of $(x_1,\dots,x_{m+1})$ is invariant under permutations of the indices, and the event $\{x_j\notin\cl(T_j)\}$ is the image of the event $\{x_{m+1}\notin\cl(T_{m+1})\}$ under the transposition of $j$ and $m+1$.
So $\P[x_{m+1}\notin\cl(T_{m+1})]=\frac{1}{m+1}\E[Z]$, where $Z=|\{j\le m+1 : x_j\notin\cl(T_j)\}|$.

Step 4 (counting).
Let $S=\{x_1,\dots,x_{m+1}\}$, as a set.
If $j$ is counted in $Z$, then $x_j\neq x_i$ for all $i\neq j$, since otherwise $x_j\in T_j\subseteq\cl(T_j)$.
So $T_j=S\setminus\{x_j\}$ and $x_j\in E(S)$, the set of essential points of $S$.
Distinct counted indices give distinct points, because each counted point occurs exactly once in the sequence.
Hence $Z\le|E(S)|\le\hgt$ by \cref{lem:essential}.
Combining the steps, $\E[\er_D(\mathcal{A}(S_m))]\le\hgt/(m+1)$.

\emph{(b).}
If $\hgt=0$, no singleton $\{x\}$ with $x\in X$ is independent, that is, $X\subseteq\cl(\emptyset)=\Fix(G)$.
Then \cref{lem:coset}(iii) with $A=\emptyset$ forces every label, so every $f\in\VS(S_m)$ equals $h^\star$ and $\er_D(f)=0$.
Assume $\hgt\ge1$ from now on.
Consider the binary learning problem with concept class $\mathcal{K}$ of \cref{lem:closed-sets}, target concept $X\in\mathcal{K}$ and distribution $D$; every example is labeled positive.
The closure algorithm outputs the intersection of all concepts that are consistent with the sample, which by \cref{lem:closed-sets}(b) is $\cl_X(\{x_1,\dots,x_m\})$.
\citet{darnstadt2015optimal} analyzes this hypothesis for every concept class on a nonempty set that is closed under intersections of arbitrary subfamilies (Theorem~8 in the author's preprint, with the explicit constants of its proof): if the class has VC dimension $d$, then for every target concept, every distribution and every $m>0$, with probability at least $1-\delta$ the closure hypothesis has error at most $\frac{48}{m}(d\log880+\log(24/\delta))$.
By \cref{lem:closed-sets}, $\mathcal{K}$ satisfies the hypotheses with $d=\hgt$.
Since $\log_bx\le\log_2x$ for every $x\ge1$ and every base $b\ge2$, the same bound holds with $\log_2$, whichever base the analysis uses.
So with probability at least $1-\delta$,
\[
  D\big(X\setminus\cl_X(\{x_1,\dots,x_m\})\big)\le\frac{48}{m}\Big(\hgt\log_2880+\log_2\frac{24}{\delta}\Big).
\]
On this event, every $f\in\VS(S_m)$ agrees with $h^\star$ on $\cl_X(\{x_1,\dots,x_m\})$ by \cref{lem:coset}(iii), so $\er_D(f)$ is at most the left-hand side.
The event does not depend on $f$, so the bound holds simultaneously for all $f\in\VS(S_m)$.
For the second form, $\log_2(24/\delta)=\log_224+\log_2(1/\delta)$ and $\hgt\log_2880+\log_224\le\hgt(\log_2880+\log_224)\le14.37\,\hgt$ since $\hgt\ge1$; and $48\cdot14.37\le690$.

\emph{(c).}
If $\hgt\ge2$, take an independent set $\{a,b\}\subseteq X$; independence says that $\stab{G}{\{b\}}$ moves $a$, so $x_0=b$, $x_1=a$ and a suitable $k\in G_b$ witness nondegeneracy.
Our condition is exactly the nondegeneracy condition of \citet{hanneke2024improved}: there are $h_1,h_2\in\Hc$ and $x_0,x_1\in X$ with $h_1(x_0)=h_2(x_0)$ and $h_1(x_1)\neq h_2(x_1)$.
Indeed, for $h_1=g_1|_X$ and $h_2=g_2|_X$ this says that $k\defeq g_1^{-1}g_2\in G_{x_0}$ and $kx_1\neq x_1$, because $g_1$ is injective; conversely, take $g_1=\mathrm{id}$ and $g_2=k$.
By \cref{thm:dictionary}(a), $\dimDS(\Hc)=\hgt$.

\citet[Theorem~2.5, with $k=1$]{hanneke2024improved} use the following definitions.
Let $\mathrm{RE}(\Hc)$ be the set of distributions $P$ on $X\times Y$ with $\inf_{f\in\Hc}\er_P(f)=0$.
For a learner $\mathcal{A}'$, let $M_{\mathcal{A}',\Hc}(\eps,\delta)$ be the least $n$ such that $\P_{S\sim P^{m'}}[\er_P(\mathcal{A}'(S))>\eps]\le\delta$ for all $m'\ge n$ and all $P\in\mathrm{RE}(\Hc)$, and let $M^{\mathrm{HMZ}}_{\Hc}(\eps,\delta)$ be the infimum of $M_{\mathcal{A}',\Hc}(\eps,\delta)$ over all learners.
They prove that for every nondegenerate class of DS dimension $d\in\N$ and all $\eps\in(0,1/16)$, $\delta\in(0,1/8)$,
\[
  M^{\mathrm{HMZ}}_{\Hc}(\eps,\delta)\ge\frac{(d-1)\ln2+4\ln(1/\delta)}{32\,\eps}.
\]

Our learners succeed at one sample size $m$ and on exactly realizable pairs, so we need a bridge.
Let $\mathcal{A}$ be a learner as in (c), and fix $\eta>0$ with $\eps+\eta<1/16$ and $\delta+m\eta<1/8$.
Let $P\in\mathrm{RE}(\Hc)$ with marginal $P_X$ on $X$, and pick $f_\eta\in\Hc$ with $\er_P(f_\eta)<\eta$.
Draw $(x_i,y_i)\sim P$ i.i.d.\ for $i\le m$, and set $S=((x_i,y_i))_{i\le m}$ and $S'=((x_i,f_\eta(x_i)))_{i\le m}$.
Then $S'$ is a sample from the exactly realizable pair $(P_X,f_\eta)$, and $\P[S\neq S']\le\sum_{i\le m}\P[y_i\neq f_\eta(x_i)]\le m\eta$.
For every predictor $f$, writing $\er_{P_X}(f;f_\eta)$ for the error of $f$ with respect to the target $f_\eta$,
\[
  \er_P(f)\le\P[f(x)\neq f_\eta(x)]+\P[f_\eta(x)\neq y]<\er_{P_X}(f;f_\eta)+\eta .
\]
Hence
\[
  \P\big[\er_P(\mathcal{A}(S))>\eps+\eta\big]\le\P[S\neq S']+\P\big[\er_{P_X}(\mathcal{A}(S');f_\eta)>\eps\big]\le m\eta+\delta .
\]
Let $\mathcal{A}'$ be the learner that, on a sample of size $m'\ge m$, runs $\mathcal{A}$ on its first $m$ examples.
The bound holds for $\mathcal{A}'$ at every $m'\ge m$ and every $P\in\mathrm{RE}(\Hc)$, so $M^{\mathrm{HMZ}}_{\Hc}(\eps+\eta,\delta+m\eta)\le m$.
The theorem of \citet{hanneke2024improved} at $(\eps+\eta,\delta+m\eta)$, with $d=\hgt$, gives
\[
  m\ge\frac{(\hgt-1)\ln2+4\ln\big(1/(\delta+m\eta)\big)}{32(\eps+\eta)} .
\]
The right-hand side is continuous in $\eta$; letting $\eta\to0$ with $m$ fixed gives the bound of (c).
If $\hgt=\infty$, then $\dimDS(\Hc)=\infty$ by \cref{thm:dictionary}(a), and a class of infinite DS dimension is not PAC learnable \citep{brukhim2022characterization}.

\emph{(d).}
A nondegenerate class has $\hgt\ge1$: the point $x_1$ is moved by $k\in G$, so $x_1\notin\Fix(G)=\cl(\emptyset)$ and $\{x_1\}$ is independent.
Upper bound: by (b), every consistent learner reaches error $\eps$ with probability at least $1-\delta$ from $m=\lceil c(\hgt+\log_2(1/\delta))/\eps\rceil$ examples.
Lower bound: the bound of (c) equals $\frac{(\hgt-1)\ln2}{32\,\eps}+\frac{\ln(1/\delta)}{8\,\eps}$, and $\ln(1/\delta)=\ln2\cdot\log_2(1/\delta)$.
If $\hgt\ge2$, then $\hgt-1\ge\hgt/2$, so $\frac{(\hgt-1)\ln2}{32}\ge\frac{\hgt\ln2}{64}$; also $\frac{\ln(1/\delta)}{8}\ge\frac{\ln(1/\delta)}{64}=\frac{\ln2\cdot\log_2(1/\delta)}{64}$.
Adding the two inequalities, the bound of (c) is at least $\frac{\ln2}{64}(\hgt+\log_2(1/\delta))/\eps$.
If $\hgt=1$, the bound of (c) is $\ln(1/\delta)/(8\eps)$.
Since $\delta<1/8$, we have $\ln(1/\delta)>\ln8>\ln2$, so
\[
  \frac{\ln(1/\delta)}{8}=\frac{\ln(1/\delta)}{64}+\frac{7\ln(1/\delta)}{64}\ge\frac{\ln2\cdot\log_2(1/\delta)}{64}+\frac{\ln2}{64}=\frac{\ln2}{64}\big(\hgt+\log_2(1/\delta)\big),
\]
and the bound of (c) is again at least $\frac{\ln2}{64}(\hgt+\log_2(1/\delta))/\eps$.
Hence $M_{\Hc}(\eps,\delta)=\Theta((\hgt+\log(1/\delta))/\eps)$, and every consistent learner attains this rate.

\emph{(e).}
Let the realizable sequence be labeled by $h^\star=g|_X$.
If at round $t$ we have $x_t\in\cl(\{x_1,\dots,x_{t-1}\})$, then by \cref{lem:coset}(iii) every hypothesis consistent with the first $t-1$ examples predicts $gx_t$, so no mistake occurs at round $t$.
Let $t_1<t_2<\cdots$ be the mistake rounds.
For each $k$, $x_{t_k}\notin\cl(\{x_1,\dots,x_{t_k-1}\})\supseteq\cl(\{x_{t_1},\dots,x_{t_{k-1}}\})$, by monotonicity of $\cl$.
So the points of the mistake rounds, in order, form an irredundant sequence, and there are at most $\irr$ of them.
For the lower bound, let $\irr<\infty$ and take the shattered mistake tree of depth $\irr$ constructed in the proof of \cref{thm:dictionary}(c).
Against a deterministic learner, walk down the tree: at each node, present its point and reveal the label of an out-edge that differs from the learner's prediction (one of the two distinct labels does).
The resulting sequence is a root-to-leaf path, hence realizable, and every round is a mistake.
\end{proof}

\subsection{The expected-error lower bound}

\PropExpectedLower*

\begin{proof}
Since $d\le\hgt_X(G)$, there are distinct points $a_1,\dots,a_d\in X$ forming an independent set (a subset of an independent set is independent, by monotonicity of $\cl$).
By the proof of \cref{thm:dictionary}(a), $B=\Hc|_A=\{(ga_1,\dots,ga_d):g\in G\}$ is a $d$-dimensional pseudo-cube, and it is finite by~\eqref{eq:F}.
For each $b\in B$ fix $g_b\in G$ with $g_ba_i=b_i$ for all $i$, and let $h_b=g_b|_X$.
Let $D(a_1)=1-(d-1)/m$ and $D(a_i)=1/m$ for $2\le i\le d$; this is a distribution because $m\ge d-1$.
Draw $b$ uniformly from $B$, independently of the sample points and of the learner's randomness, and use the target $h_b$.

Fix $i\in\{2,\dots,d\}$ and let $U_i$ be the event that $a_i$ is not among $x_1,\dots,x_m$.
Condition on the sample points, with $U_i$ holding, and let $J\subseteq[d]\setminus\{i\}$ be the set of indices of the points that were drawn.
Given the observed labels $(b_j)_{j\in J}$, the target $b$ is uniform on $B_J=\{b'\in B:b'_j=b_j\text{ for }j\in J\}$.
Partition $B_J$ into classes of words that agree outside coordinate $i$.
Each class lies in $B_J$, because $i\notin J$, and is an $i$-panel of $B$.
So it has at least two words, and their $i$-th coordinates are pairwise distinct.
Hence for every label $y$ at most half of $B_J$ has $b'_i=y$.
The prediction at $a_i$ depends only on the observed sample and the learner's randomness, so it errs with conditional probability at least $1/2$.

Since $\er_D(f)\ge\sum_{i=2}^d\frac1m\mathbf 1[f(a_i)\neq b_i]$, averaging over $b$ gives
\[
\E_b\E\big[\er_D(\mathcal A(S_m))\big]\ge\sum_{i=2}^d\frac1m\cdot\P(U_i)\cdot\frac12\ge\frac{d-1}{8m},
\]
because $\P(U_i)=(1-1/m)^m\ge1/4$ for $m\ge2$.
Some $b$ attains at least the average; take $h^\star=h_b$.
For the comparison, take $d=\hgt$.
\Cref{thm:erm}(a) gives $\hgt/(m+1)$ for every consistent learner, and $\frac{\hgt/(m+1)}{(\hgt-1)/(8m)}<\frac{8\hgt}{\hgt-1}\le16$.
If $\hgt=\infty$, take $d=m+1$.
\end{proof}

\subsection{Linear, orthogonal and unitary maps}

\PropLinear*

\begin{proof}
(a) A linear map fixing $A$ fixes $\operatorname{span}(A)=W$.
Conversely, for $x\notin W$: in $\mathrm{GL}(d,\mathbb F)$, extend a basis of $W$ by $x$ and scale $x$ by~$2$; in $\mathrm O(d)$ or $\mathrm U(d)$, write $x=w+y$ with $0\ne y\in W^\perp$ and use the reflection $I-2yy^*/(y^*y)$, which fixes $W$ and sends $x$ to $w-y$.
So $\Fix(\stab{G}{A})=W$.

(b) If a sequence $(a_i)$ carries a pseudo-cube and $a_i=\sum_{j\ne i}c_ja_j$, then every $h\in\Hc$ is linear, so $h(a_i)$ is determined by the other coordinates, and no word has an $i$-neighbor; hence DS-shattered sequences are linearly independent and $\dimDS\le d$.
The diagonal matrices $\mathrm{diag}((-1)^{\varepsilon_i})$ Natarajan-shatter the standard basis with $f_0(e_i)=e_i$ and $f_1(e_i)=-e_i$, and $\dimN\le\dimDS$.

(c) A consistent output agrees with the target on $\operatorname{span}(x_1,\dots,x_m)$, so its error is at most $D(X\setminus\operatorname{span}(x_1,\dots,x_m))$.
The set of $(y_1,\dots,y_{m+1})$ with $y_{m+1}\notin\operatorname{span}(y_1,\dots,y_m)$ is Borel (a rank condition), so by Tonelli and exchangeability the expectation of this quantity equals $\frac1{m+1}\E[Z]$, where $Z$ counts the indices $j$ with $x_j\notin\operatorname{span}(x_i:i\ne j)$.
These $x_j$ are distinct and linearly independent, so $Z\le d$.

(d) Put mass $1/m$ on $e_2,\dots,e_d$ and the rest on $e_1$, and draw a uniformly random sign vector $s$ for the target $\mathrm{diag}(s)$.
At an unsampled $e_i$ the sign $s_i$ is independent of the observed labels, so every prediction errs with probability at least $1/2$; each $e_i$ ($i\ge2$) is unsampled with probability $(1-1/m)^m\ge1/4$.
Summing gives $(d-1)/(8m)$ on average over $s$, hence for some $s$.
\end{proof}

\subsection{Agnostic ERM and the graph dimension}

\PropAgnostic*

\begin{proof}
(a) If $(x_1,\dots,x_t)$ is Natarajan-shattered with witnesses $f_0,f_1$, the $x_i$ are distinct (\cref{thm:natarajan}(i)), and $\{x_1,\dots,x_t\}$ is graph-shattered with witness $f_0$; so $\dimN\le d_G$.
Let $A$ be graph-shattered with witness $f$ and let $a\in A$.
The sets $T=A$ and $T=A\setminus\{a\}$ give $g_1,g_2\in G$ with $g_1=f=g_2$ on $A\setminus\{a\}$, $g_1a=f(a)$ and $g_2a\neq f(a)$; then $g_1^{-1}g_2\in\stab{G}{A\setminus\{a\}}$ moves $a$, so $a\notin\cl(A\setminus\{a\})$, $A$ is independent and $d_G\le\hgt$.
Conversely, let $A$ be a finite independent set and $f$ the identity on $A$.
For each $a\in A$ pick $k_a\in\stab{G}{A\setminus\{a\}}$ with $k_aa\neq a$.
Given $T\subseteq A$, write $A\setminus T=\{b_1,\dots,b_q\}$ and put $P=k_{b_1}\cdots k_{b_q}$.
Every factor fixes $T$ pointwise, so $P$ does.
For each $j$, the factors after $k_{b_j}$ fix $b_j$, so $Pb_j=Q(z)$ with $Q=k_{b_1}\cdots k_{b_{j-1}}$ and $z=k_{b_j}b_j\neq b_j$; as $Q$ fixes $b_j$, $Q(z)\neq b_j$.
So $P|_X$ agrees with $f$ exactly on $T$, and $A$ is graph-shattered.
Subsets of independent sets are independent, so $d_G\ge\hgt$ also when $\hgt=\infty$.

(b) \Citet[Theorem~5]{daniely2015multiclass} bound the agnostic sample complexity of the worst ERM (a learner returning a minimizer of the empirical error in the class) by $C_2(d_G+\ln(1/\delta))/\eps^2$ for every class $\Hc\subseteq Y^X$ with $X$ and $Y$ countable, where $Y$ may be infinite; take $Y=\Omega$ and use (a). (If $\hgt=0$, then $G$ fixes $X$ pointwise, $\Hc$ has one element, and every ERM is exact with no examples.)

(c) Two distinct hypotheses differ at some point, which they Natarajan-shatter, so $\dimN\ge1$.
With $\lambda=\ln(1/\delta)>0$ and $\hgt\ge\dimN\ge1$, we have $\hgt+\lambda\le(\hgt/\dimN)(\dimN+\lambda)$, since the difference is $\lambda(\hgt-\dimN)/\dimN\ge0$; multiply by $C_2/\eps^2$.
\end{proof}

\section{Proofs for Section~\ref{sec:fields}: Galois groups}
\label{app:fields}

\ThmGalois*

\begin{proof}
Every orbit satisfies $|\Gamma\alpha|\le|\Gamma|<\infty$, so~\eqref{eq:F} holds.

(i) An automorphism $\sigma\in\Gamma$ fixes $A$ pointwise if and only if it fixes the field $K(A)$ pointwise.
So $\stab{\Gamma}{A}=\Gal(L/K(A))$, and $\cl(A)=\Fix(\Gal(L/K(A)))=K(A)$ by the fundamental theorem of Galois theory \citep[Chapter~VI, Theorems~1.1 and~1.2]{lang2002algebra}.

(ii) \Cref{fact:realization} with $U=1$ gives $\theta\in L$ with $\Gamma_\theta=1$.
Then $\cl(\{\theta\})=L$, so $\{\theta\}$ is a relative base.
The empty set is not a relative base, because $\cl(\emptyset)=K\neq L$ when $\Gamma\neq1$.
So $\bsz_L(\Gamma)=1$, and $\TD(\Hc)=1$ by \cref{thm:dictionary}(d).

(iii) Let $\Delta\le\Gamma$ act on $L$; then $\stab{\Delta}{B}=\Delta\cap\stab{\Gamma}{B}$ for $B\subseteq L$.
A finite $A\subseteq L$ is $\Delta$-independent if and only if $\stab{\Delta}{A\setminus\{a\}}=\bigcap_{b\in A\setminus\{a\}}\Delta_b$ is not contained in $\Delta_a$ for every $a\in A$.
We show that the largest size of such a set is $\mu(\Delta)$.
($\le$) Let $A$ be $\Delta$-independent, and pick $g_a\in\big(\bigcap_{b\neq a}\Delta_b\big)\setminus\Delta_a$ for each $a\in A$.
For $b\neq a$ we have $g_b\in\Delta_a$, so $\langle g_b : b\neq a\rangle\le\Delta_a$, which does not contain $g_a$.
The elements $g_a$ are distinct, since $g_a\notin\Delta_a\ni g_b$ for $b\neq a$.
So $\{g_a : a\in A\}$ is an independent subset of $\Delta$ of size $|A|$.
($\ge$) Let $\{g_1,\dots,g_h\}\subseteq\Delta$ be independent, and put $U_i=\langle g_j : j\neq i\rangle\le\Delta$.
By \cref{fact:realization} choose $\alpha_i\in L$ with $\Gamma_{\alpha_i}=U_i$; then $\Delta_{\alpha_i}=\Delta\cap U_i=U_i$.
The subgroups $U_i$ are pairwise distinct, because $g_i\in U_j\setminus U_i$ for $j\neq i$; so the points $\alpha_i$ are distinct.
Moreover $g_i\in\bigcap_{j\neq i}\Delta_{\alpha_j}$ and $g_i\notin U_i=\Delta_{\alpha_i}$, so $\{\alpha_1,\dots,\alpha_h\}$ is $\Delta$-independent.
Hence $\hgt_L(\Delta)=\mu(\Delta)$.
With $\Delta=\Gamma$ and \cref{thm:dictionary}(a), $\dimDS(\Hc)=\mu(\Gamma)$.

(iv) A sequence $(a_1,\dots,a_k)$ is irredundant if and only if the chain $\Gamma\ge\stab{\Gamma}{\{a_1\}}\ge\stab{\Gamma}{\{a_1,a_2\}}\ge\dots\ge\stab{\Gamma}{\{a_1,\dots,a_k\}}$ is strictly decreasing.
Indeed, $a_i\notin\cl(\{a_1,\dots,a_{i-1}\})$ if and only if $\stab{\Gamma}{\{a_1,\dots,a_{i-1}\}}$ moves $a_i$, that is, $\stab{\Gamma}{\{a_1,\dots,a_i\}}\neq\stab{\Gamma}{\{a_1,\dots,a_{i-1}\}}$.
A strictly decreasing chain of subgroups that starts at $\Gamma$ can be extended down to $1$, so its length is at most $\ell(\Gamma)$.
Conversely, for a chain $\Gamma=U_0>U_1>\dots>U_\ell=1$, choose $\alpha_i$ with $\Gamma_{\alpha_i}=U_i$ by \cref{fact:realization}.
Then $\stab{\Gamma}{\{\alpha_1,\dots,\alpha_i\}}=\bigcap_{j\le i}U_j=U_i$, so $(\alpha_1,\dots,\alpha_\ell)$ is irredundant.
Hence $\irr_L(\Gamma)=\ell(\Gamma)$, and $\Ldim(\Hc)=\ell(\Gamma)$ by \cref{thm:dictionary}(c).

(v) By \cref{lem:coset}(ii), the consistent labels of $\alpha$ are $\sigma(\stab{\Gamma}{A}\,\alpha)$ for the target $\sigma$, a set of size $|\stab{\Gamma}{A}\,\alpha|$.
By (i), $\stab{\Gamma}{A}=\Gal(L/K(A))$, and $L/K(A)$ is Galois.
So the orbit of $\alpha$ under $\Gal(L/K(A))$ is the set of roots of the minimal polynomial of $\alpha$ over $K(A)$, which is separable of degree $[K(A)(\alpha):K(A)]$ (\cref{fact:conjugates}).

(vi) By \cref{thm:natarajan}(iii) and by (iii) applied to $\Delta$, $\dimN(\Hc)\ge\hgt_L(\Delta)=\mu(\Delta)$ for every abelian $\Delta\le\Gamma$.
It remains to compute $\mu$ for a finite abelian group $\Delta$, which we write additively.
($\ge$) Write $\Delta=\bigoplus_{i=1}^rZ_i$ with $Z_i=\langle z_i\rangle$ cyclic of prime-power order, where $r=\sum_pd(\Delta_p)$.
Then $z_i\notin\bigoplus_{j\neq i}Z_j=\langle z_j : j\neq i\rangle$, so $\{z_1,\dots,z_r\}$ is independent.
($\le$) Let $\{h_1,\dots,h_s\}$ be independent, let $N=\langle h_1,\dots,h_s\rangle$ and $B_i=\langle h_j : j\neq i\rangle$.
The quotient $N/B_i$ is cyclic, generated by the image $\bar h_i$ of $h_i$, and nontrivial; choose a prime $p_i$ dividing $|N/B_i|$.
Fix a prime $p$ and let $I_p=\{i : p_i=p\}$.
Suppose that $\sum_{i\in I_p}c_ih_i\in pN$ for integers $0\le c_i<p$, not all zero, say $c_k\neq0$.
Reducing modulo $B_k$ gives $c_k\bar h_k=p\bar n$ for some $\bar n\in N/B_k=\langle\bar h_k\rangle$; writing $\bar n=t\bar h_k$, we get $(c_k-pt)\bar h_k=0$, so the order of $\bar h_k$ divides $c_k-pt$.
Since $p$ divides this order, $p$ divides $c_k$, a contradiction.
So the images of the $h_i$, $i\in I_p$, in $N/pN$ are linearly independent over $\mathbb{F}_p$, and $|I_p|\le\dim_{\mathbb{F}_p}N/pN=d(N_p)\le d(\Delta_p)$; the last inequality holds because a subgroup of a finite abelian $p$-group needs no more generators than the group.
Summing over $p$ gives $s\le\sum_pd(\Delta_p)$.
\end{proof}

\CorGaloisValues*

\begin{proof}
In each case $\Gamma\neq1$, so $\TD=1$ by \cref{thm:galois}(ii).

(a) \citet{whiston2000maximal} proved that every independent subset of $S_n$ has at most $n-1$ elements, with equality attained; this is stated for arbitrary (not necessarily generating) independent subsets by \citet[Section~2]{cameron2002independent}.
So $\mu(S_n)=n-1$ and $\dimDS=n-1$ by \cref{thm:galois}(iii).
By \citet{cameron1989chains}, $\ell(S_n)=\lceil3n/2\rceil-b(n)-1$, so \cref{thm:galois}(iv) gives $\Ldim$.
The $\lfloor n/2\rfloor$ disjoint transpositions $(1\,2),(3\,4),\dots$ generate an abelian subgroup isomorphic to $(\Z/2)^{\lfloor n/2\rfloor}$, whose $\mu$ is $\lfloor n/2\rfloor$ by the formula in \cref{thm:galois}(vi); so $\dimN\ge\lfloor n/2\rfloor$.
Finally $\dimN\le\dimDS=n-1$ by \cref{thm:natarajan}(i).

(b) $\Z/n$ is abelian, so $\dimN=\dimDS$ by \cref{thm:natarajan}(iii), and $\dimDS=\mu(\Z/n)=\sum_pd((\Z/n)_p)=\omega(n)$, because the Sylow $p$-subgroup of $\Z/n$ is cyclic and nontrivial exactly when $p$ divides $n$.
The subgroups of $\Z/n$ correspond to the divisors of $n$, ordered by divisibility, so a longest subgroup chain has length $\omega^{+}(n)$, and $\Ldim=\omega^{+}(n)$.

(c) $(\Z/2)^r$ is abelian with $\mu((\Z/2)^r)=r$ by the formula in \cref{thm:galois}(vi), so $\dimN=\dimDS=r$.
Every step of a strictly decreasing chain of subgroups at least halves the order, so $\ell((\Z/2)^r)\le r$, and the chain of coordinate subgroups has length $r$; so $\Ldim=r$.
\end{proof}

\section{Proofs for Section~\ref{sec:natarajan}: commuting symmetries and the Natarajan dimension}
\label{app:natarajan}

\subsection{Commuting moves}

\ThmNatarajan*

\begin{proof}
(i) Let $(t_1,\dots,t_s)$ be N-shattered with witnesses $f_0,f_1$.
Distinctness: if $t_i=t_j$ with $i\neq j$, the pattern with $\epsilon_i=0$ and $\epsilon_j=1$ requires some $h$ with $h(t_i)=f_0(t_i)$ and $h(t_i)=h(t_j)=f_1(t_j)=f_1(t_i)\neq f_0(t_i)$, which is absurd.
Independence: let $T=\{t_1,\dots,t_s\}$ and fix $i$.
Choose any $\epsilon$ with $\epsilon_i=0$, let $\epsilon'$ be $\epsilon$ with the $i$-th entry changed to $1$, and let $g,g'\in G$ realize $\epsilon$ and $\epsilon'$ on $T$.
Then $gt_j=g't_j$ for $j\neq i$, while $gt_i=f_0(t_i)\neq f_1(t_i)=g't_i$.
So $g^{-1}g'\in\stab{G}{T\setminus\{t_i\}}$ moves $t_i$, that is, $t_i\notin\cl(T\setminus\{t_i\})$.
Hence $T$ is independent and $s\le\hgt_X(G)$.
Under~\eqref{eq:F}, $\dimDS(\Hc)=\hgt_X(G)$ by \cref{thm:dictionary}(a).

(ii) For $\epsilon\in\{0,1\}^A$ let $g_\epsilon=\prod_{a:\epsilon_a=1}k_a$, which is well defined because the $k_a$ commute.
Fix $b\in A$.
If $\epsilon_b=0$, every factor $k_a$ with $a\neq b$ lies in $\stab{G}{A\setminus\{a\}}$ and therefore fixes $b$; so $g_\epsilon b=b=f_0(b)$.
If $\epsilon_b=1$, write $g_\epsilon=k_br$ with $r=\prod_{a\neq b,\ \epsilon_a=1}k_a$, using commutativity.
Since $r$ fixes $b$, $g_\epsilon b=k_b(rb)=k_bb=f_1(b)$.
As $k_bb\neq b$, we have $f_0(b)\neq f_1(b)$.
So every pattern $\epsilon$ is realized by $g_\epsilon|_X\in\Hc$, and $A$ is N-shattered.

(iii) Let $G$ be abelian and let $A\subseteq X$ be finite and independent.
Choose $k_a\in\stab{G}{A\setminus\{a\}}$ moving $a$, for each $a\in A$; these elements commute, so $A$ is N-shattered by (ii).
Together with (i), $\dimN(\Hc)=\hgt_X(G)$, including the value $+\infty$.
Now let $K\le G$ be abelian and let $A\subseteq X$ be finite and $K$-independent, that is, $a\notin\Fix(\stab{K}{A\setminus\{a\}})$ for all $a\in A$.
Choose $k_a\in\stab{K}{A\setminus\{a\}}\subseteq\stab{G}{A\setminus\{a\}}$ moving $a$.
These elements commute, so by (ii), applied to $K$, the set $A$ is N-shattered by $\HG{K}{X}\subseteq\HG{G}{X}$.
The Natarajan dimension is monotone under inclusion of classes, so $\dimN(\HG{G}{X})\ge\hgt_X(K)$.
\end{proof}

\subsection{The factor one half}

\PropSymmetric*

\begin{proof}
(a) Let $(t_1,\dots,t_s)$ be N-shattered by $\HG{G}{X}$ with witnesses $f_0,f_1$, where $s\ge1$.
If $s=1$, then $f_0(t_1)\neq f_1(t_1)$ requires $n\ge2$, so $s=1\le\lfloor n/2\rfloor$.
If $s\ge2$, we claim that the $2s$ labels $f_e(t_i)$, for $e\in\{0,1\}$ and $i\le s$, are pairwise distinct.
First, $f_0(t_i)\neq f_1(t_i)$ by definition.
Second, for $i\neq j$ and $e,e'\in\{0,1\}$, some pattern is realized by an element $g$ with $gt_i=f_e(t_i)$ and $gt_j=f_{e'}(t_j)$; since $t_i\neq t_j$ by \cref{thm:natarajan}(i) and $g$ is injective, $f_e(t_i)\neq f_{e'}(t_j)$.
So $2s\le n$.

For the height, note that $n\ge1$ since $X\neq\emptyset$.
If $\Omega$ itself were independent, then for $a\in\Omega$ the group $\stab{G}{\Omega\setminus\{a\}}$ would fix the $n-1$ points of $\Omega\setminus\{a\}$, hence also $a$, because a permutation that fixes all points but one fixes that one too.
This contradicts independence.
So every independent set is a proper subset of $\Omega$, and $\hgt(G)\le n-1$.

(b) The set $\{1,\dots,n-1\}$ is independent for $S_n$: the pointwise stabilizer of $\{1,\dots,n-1\}\setminus\{i\}$ contains the transposition $(i\,n)$, which moves $i$.
With (a), $\hgt(S_n)=n-1$.
For the Natarajan dimension, let $K=\langle(1\,2),(3\,4),\dots,(2\lfloor n/2\rfloor-1\ \ 2\lfloor n/2\rfloor)\rangle$, an abelian subgroup.
The set $A'=\{1,3,\dots,2\lfloor n/2\rfloor-1\}$ is $K$-independent: the transposition $(2i-1\ \ 2i)\in K$ fixes $A'\setminus\{2i-1\}$ and moves $2i-1$.
\Cref{thm:natarajan}(iii) gives $\dimN\ge|A'|=\lfloor n/2\rfloor$, and (a) gives equality.

(c) Every $g\in G$ is a pair $(g_1,g_2)$.
For $A=A_1\sqcup A_2$ with $A_r\subseteq\Omega_r$, we have $\stab{G}{A}=\stab{(G_1)}{A_1}\times\stab{(G_2)}{A_2}$.
For $a\in A_1$, the group $\stab{G}{A\setminus\{a\}}=\stab{(G_1)}{A_1\setminus\{a\}}\times\stab{(G_2)}{A_2}$ moves $a$ if and only if $\stab{(G_1)}{A_1\setminus\{a\}}$ does, because $G_2$ acts trivially on $\Omega_1$; symmetrically for $a\in A_2$.
So $A$ is independent if and only if $A_1$ and $A_2$ are, and the height is additive.
For the Natarajan dimension, $\HG{G}{\Omega_1\sqcup\Omega_2}$ is the product class $\{(h_1,h_2)\}$ of $\HG{G_1}{\Omega_1}$ and $\HG{G_2}{\Omega_2}$ on disjoint domains.
If $T=T_1\sqcup T_2$ (with $T_r\subseteq\Omega_r$) is N-shattered with witnesses $f_0,f_1$, then restricting the patterns to $T_r$ shows that $T_r$ is N-shattered by $\HG{G_r}{\Omega_r}$, because the $T_r$-coordinates of a realizing pair $(h_1,h_2)$ are given by $h_r$; so the Natarajan dimension is subadditive.
Conversely, if $T_r$ is N-shattered by $\HG{G_r}{\Omega_r}$ for $r=1,2$, then the witnesses combine on $T_1\sqcup T_2$, and every pattern is realized by the pair of elements that realize its two restrictions; so it is superadditive.

(d) For $S_3$ acting on $3$ points, (b) gives $\hgt=2$ and $\dimN=\lfloor3/2\rfloor=1$.
Applying (c) $k-1$ times gives $\hgt=2k$ and $\dimN=k$ for $S_3^k$ acting on $3k$ points.
The ratio is then $1/2$ at every even height $2k$, which gives the final statement.
\end{proof}

\subsection{Matroid closures}

\LemTower*

\begin{proof}
For $\epsilon\in\{0,1\}^t$ let $g_\epsilon=k_t^{\epsilon_t}\cdots k_2^{\epsilon_2}k_1^{\epsilon_1}$, so that $k_1$ acts first.
Fix $i$.
The factors $k_j^{\epsilon_j}$ with $j<i$ act first and fix $x_i$ by (T2).
Then $k_i^{\epsilon_i}$ sends $x_i$ to $x_i$ if $\epsilon_i=0$ and to $x_i'\defeq k_ix_i$ if $\epsilon_i=1$, and $x_i'\neq x_i$ by (T1).
The remaining factors $k_j^{\epsilon_j}$ with $j>i$ fix $x_i$ by (T2) and fix $x_i'$ by (T3), applied to the pair $i<j$.
Hence $g_\epsilon x_i=f_{\epsilon_i}(x_i)$ for every $i$, with $f_0(x_i)\neq f_1(x_i)$.
Since the $x_i$ are distinct, every pattern $\epsilon$ is realized by $g_\epsilon|_X\in\HG{G}{X}$.
If the $k_j$ commute and satisfy (T2), then for $i<j$ we have $k_jk_ix_i=k_ik_jx_i=k_ix_i$, which is (T3).
\end{proof}

\ThmMatroid*

\begin{proof}
\emph{The matroid.}
The operator $\cl$ is extensive, monotone and idempotent (\cref{lem:closure}).
Together with exchange on the finite set $\Omega$, these are the closure axioms of a matroid on $\Omega$ whose closure operator is $\cl$, and whose independent sets are the sets $I$ with $x\notin\cl(I\setminus\{x\})$ for all $x\in I$ \citep[Section~1.4]{oxley2011matroid}; these are exactly the independent sets of \cref{sec:prelim}.
Let $r$ be its rank function.
We use the standard facts \citep[Sections~1.3--1.4]{oxley2011matroid}: (M0) $r(\cl S)=r(S)$; (M1) $r(S)\le r(S\cup T)\le r(S)+|T|$; (M2) $s\in\cl(T)$ if and only if $r(T\cup\{s\})=r(T)$.
Say that $S$ is \emph{independent over} $F$ if $s\notin\cl(F\cup(S\setminus\{s\}))$ for every $s\in S$.
(M3) If $S$ is independent over $F$, then $r(F\cup S)=r(F)+|S|$: order $S$ as $s_1,\dots,s_k$; by monotonicity $s_i\notin\cl(F\cup\{s_1,\dots,s_{i-1}\})$, so by (M2) each step raises the rank by one.
(M4) For all $F,T$ there is $S\subseteq T$ independent over $F$ with $|S|=r(F\cup T)-r(F)$: scan $T$ and keep an element exactly when it is not in the closure of $F$ together with the elements kept before it.
Every discarded element lies in $\cl(F\cup S)$, so $\cl(F\cup S)=\cl(F\cup T)$ and, by (M0), $r(F\cup S)=r(F\cup T)$.
Each kept element raises the rank by one, so $r(F\cup S)=r(F)+|S|$.
For $s\in S$, (M1) gives $r(F\cup(S\setminus\{s\}))\le r(F)+|S|-1<r(F\cup S)$, so $s\notin\cl(F\cup(S\setminus\{s\}))$ by (M2).

\emph{The recursion.}
Let $A$ be independent with $|A|=h=\hgt(G)$, and put $F_0=\emptyset$ and $R_1=A$, which is independent over $\emptyset$.
For $j=1,2,\dots$, while $R_j\neq\emptyset$: pick $x_j\in R_j$.
Since $R_j$ is independent over $F_{j-1}$, we have $x_j\notin\cl(F_{j-1}\cup(R_j\setminus\{x_j\}))$, so some $k_j\in\stab{G}{F_{j-1}\cup(R_j\setminus\{x_j\})}$ moves $x_j$.
Put $x_j'=k_jx_j$ and $F_j=F_{j-1}\cup\{x_j,x_j'\}$.
Since $F_j\cup(R_j\setminus\{x_j\})\supseteq F_{j-1}\cup R_j$, (M1) and (M3) give
\[
r\big(F_j\cup(R_j\setminus\{x_j\})\big)\ \ge\ r(F_{j-1}\cup R_j)\ =\ r(F_{j-1})+|R_j|\ \ge\ r(F_j)-2+|R_j|,
\]
where the last step uses $r(F_j)\le r(F_{j-1})+2$.
By (M4), choose $R_{j+1}\subseteq R_j\setminus\{x_j\}$ independent over $F_j$ with $|R_{j+1}|\ge|R_j|-2$.

\emph{The tower.}
The point $x_j$ is not in $\cl(F_{j-1}\cup(R_j\setminus\{x_j\}))$, a set that contains every earlier point $x_i$ and every later point $x_i$ (the later points lie in $R_{j+1}\subseteq R_j\setminus\{x_j\}$); so the points are distinct.
(T1) holds by the choice of $k_j$.
For $i<j$, both $x_i$ and $x_i'$ lie in $F_{j-1}$, which $k_j$ fixes pointwise: this gives (T2) for $i<j$ and (T3).
For $i>j$, $x_i\in R_i\subseteq R_{j+1}\subseteq R_j\setminus\{x_j\}$, which $k_j$ fixes pointwise: this gives (T2) for $i>j$.
Since $|R_j|\ge h-2(j-1)\ge1$ for all $j\le\lceil h/2\rceil$, the recursion produces at least $\lceil h/2\rceil$ points, and \cref{lem:tower} gives $\dimN(\HG{G}{\Omega})\ge\lceil h/2\rceil$.

\emph{Tightness.}
For $S_n$ acting on $[n]$, $n\ge2$, the pointwise stabilizer of $A$ is $\Sym([n]\setminus A)$, so $\cl(A)=A$ if $|A|\le n-2$ and $\cl(A)=[n]$ otherwise.
Exchange holds: if $q\in\cl(S\cup\{p\})\setminus\cl(S)$, then $|S|\le n-2$ and $q\notin S$; either $|S\cup\{p\}|\le n-2$, in which case $q\in S\cup\{p\}$ forces $q=p$; or $|S|=n-2<|S\cup\{p\}|$, in which case $|S\cup\{q\}|=n-1$ and $p\in[n]=\cl(S\cup\{q\})$.
For $n=2k+1$, \cref{prop:symmetric}(b) gives $\hgt=2k$ and $\dimN=k$.
\end{proof}

\subsection{Natarajan squares}

\PropSquare*

\begin{proof}
($\Leftarrow$) Let $w\in(PQ\cap QP)\setminus(P\cup Q)$, and write $w=pq=q'p'$ with $p,p'\in P$ and $q,q'\in Q$.
Take $f_0(x)=x$, $f_0(y)=y$, $f_1(x)=wx$ and $f_1(y)=wy$; then $f_1(x)\neq x$ because $w\notin P=G_x$, and $f_1(y)\neq y$ because $w\notin Q$.
The pattern $(0,0)$ is realized by the identity.
The pattern $(1,0)$ is realized by $q'$: it fixes $y$, and $q'x=q'p'x=wx$.
The pattern $(0,1)$ is realized by $p$: it fixes $x$, and $py=pqy=wy$.
The pattern $(1,1)$ is realized by $w$.

($\Rightarrow$) Let $f_0,f_1$ be witnesses and $g_{ij}$ ($i,j\in\{0,1\}$) realizers, with $g_{ij}x=f_i(x)$ and $g_{ij}y=f_j(y)$.
Put $k_{ij}=g_{00}^{-1}g_{ij}$.
Then $k_{ij}x=g_{00}^{-1}f_i(x)$, which equals $g_{00}^{-1}f_0(x)=x$ exactly when $i=0$; similarly at $y$.
So $k_{10}\in Q$, $k_{01}\in P$, and $w\defeq k_{11}$ moves both $x$ and $y$, so $w\notin P\cup Q$.
Moreover $wx=k_{10}x$, so $k_{10}^{-1}w\in P$ and $w\in k_{10}P\subseteq QP$; and $wy=k_{01}y$, so $w\in k_{01}Q\subseteq PQ$.

(a) Note $u\neq v$, since $h\in G_v$ moves $u$.
Put $x=u$ and $y=v$, so $P=G_u$ and $Q=G_v$, pick $m\in\stab{G}{\{u,hu\}}\setminus G_v$, and let $w=mh$.
Then $w\in PQ$ because $m\in P$ and $h\in Q$.
Also $w=h(h^{-1}mh)$, and $h^{-1}mh$ fixes $u$ because $m$ fixes $hu$; so $w\in QP$.
Finally $wu=m(hu)=hu\neq u$, so $w\notin P$, and $wv=m(hv)=mv\neq v$, so $w\notin Q$.

(b) Let $R=P\cap Q$, $r_P=[P:R]$ and $r_Q=[Q:R]$.
Independence of $\{x,y\}$ says that $Q$ moves $x$ and $P$ moves $y$, so neither subgroup contains the other and $r_P,r_Q\ge2$.
If $PQ=QP$, then $|PQ\cap QP|=|PQ|=r_Pr_Q|R|$, while $|P\cup Q|=(r_P+r_Q-1)|R|<r_Pr_Q|R|$ because $(r_P-1)(r_Q-1)\ge1$.

(c) Pick $h\in G_c$ with $ha=b$.
Independence gives $\stab{G}{\{a,b\}}\not\subseteq G_c$, so (a) with $u=a$ and $v=c$ shows that $(a,c)$ is N-shattered.
If $G$ is $2$-transitive and $\hgt(G)\ge3$, take three points $a,b,c$ of an independent set; subsets of independent sets are independent because $\cl$ is monotone, and $G_c$ is transitive on $\Omega\setminus\{c\}$, so $b\in G_c\,a$.
\end{proof}

\paragraph{Computations for \cref{conj:half}.}
Before the run we fixed a list of permutation groups, all of degree at most $20$: the cyclic groups $C_3,\dots,C_7$ and the dihedral groups $D_3,\dots,D_7$ ($D_n$ of order $2n$ on $n$ points), $S_3,\dots,S_7$, $A_4,\dots,A_7$, $\mathrm{AGL}(1,5)$, $\mathrm{AGL}(1,7)$, $\mathrm{PGL}(2,q)$ for $q\in\{3,5,7\}$, $\mathrm{PSL}(2,q)$ for $q\in\{5,7\}$, $S_3\times S_3$ and $S_4\times C_3$ (on $6$ and $7$ points), the imprimitive wreath products $S_2\wr S_3$, $S_2\wr S_4$ and $S_3\wr S_2$ (on $6$, $8$ and $6$ points), $\mathrm{GL}(3,2)$ on the $7$ nonzero vectors of $\mathbb F_2^3$, $\mathrm{AGL}(3,2)$ on $8$ points, $S_5$ on $2$-sets ($10$ points), and $S_6$ on $2$-sets and on $3$-sets ($15$ and $20$ points).
We computed $\hgt$ and $\dimN$ exactly for every group of the list up to $\mathrm{GL}(3,2)$; these groups all have degree at most $8$.
The last four groups of the list, $\mathrm{AGL}(3,2)$ and the actions of $S_5$ and $S_6$ on $k$-sets, were not reached within the time limit.
The second sweep, run by an independent script, covered all coset actions and two-orbit unions of degree at most $40$ of $33$ groups (including $S_4$--$S_6$, $A_5$, $A_6$, $\mathrm{PSL}(2,q)$ and $\mathrm{PGL}(2,q)$ for small $q$, $\mathrm{GL}(2,3)$, $\mathrm{AGL}(2,3)$, $\mathrm{AGL}(3,2)$, the Coxeter groups of types $B_3$, $B_4$ and $D_4$, wreath products and direct products, restricted to subgroups generated by at most two elements), quotients of affine Coxeter groups of types $\tilde A_2$, $\tilde A_3$, $\tilde A_4$, and $870$ actions of $p$-groups: $7{,}292$ actions in all.

\section{Proofs for Section~\ref{sec:solvable}: solvability and the Natarajan dimension}
\label{app:solvable}

\subsection{Coset cubes}

\ThmCosetCube*

\begin{proof}
(i) ($\le$) Let $x_1,\dots,x_t\in L$ be Natarajan-shattered with witnesses $f_0,f_1$, and for each $\varepsilon\in\{0,1\}^t$ let $\sigma_\varepsilon\in\Gamma$ realize $\varepsilon$.
Put $\tau_\varepsilon=\sigma_0^{-1}\sigma_\varepsilon$, where $0$ is the all-zero pattern.
If $\varepsilon_i=0$ then $\tau_\varepsilon x_i=\sigma_0^{-1}f_0(x_i)=x_i$.
If $\varepsilon_i=1$ then $\tau_\varepsilon x_i=\sigma_0^{-1}f_1(x_i)=:y_i$, and $y_i\neq x_i$ because $f_1(x_i)\neq f_0(x_i)=\sigma_0x_i$.
Let $V_i=\Gamma_{x_i}$ and $k_i=\tau_{e_i}$, where $e_i$ is the $i$-th unit pattern.
Then $k_ix_i=y_i\neq x_i$, so $k_i\notin V_i$.
For every $\varepsilon$: if $\varepsilon_i=0$ then $\tau_\varepsilon\in V_i$; if $\varepsilon_i=1$ then $\tau_\varepsilon x_i=k_ix_i$, so $\tau_\varepsilon\in k_iV_i$.
Hence $\tau_\varepsilon\in\bigcap_ik_i^{\varepsilon_i}V_i$, and $(V_i,k_i)$ is a coset cube.

($\ge$) Let $(V_i,k_i)_{i\le t}$ be a coset cube with $t\ge1$.
By \cref{fact:realization} choose $\alpha_i\in L$ with $\Gamma_{\alpha_i}=V_i$.
The $\alpha_i$ are distinct: if $\alpha_i=\alpha_j$ with $i\neq j$, then $V_i=V_j$, and a pattern with $\varepsilon_i=0$ and $\varepsilon_j=1$ would need an element of $V_j\cap k_jV_j=\emptyset$.
Let $f_0(\alpha_i)=\alpha_i$ and $f_1(\alpha_i)=k_i\alpha_i\neq\alpha_i$.
For a pattern $\varepsilon$ pick $\sigma\in\bigcap_ik_i^{\varepsilon_i}V_i$.
If $\varepsilon_i=0$, $\sigma$ fixes $\alpha_i$; if $\varepsilon_i=1$, $\sigma=k_iv$ with $v\in V_i$, so $\sigma\alpha_i=k_i\alpha_i$.
So $(\alpha_1,\dots,\alpha_t)$ is Natarajan-shattered.

(ii) A cube in a subgroup $\Delta$ is a cube in $\Gamma$.
For a quotient $\pi:\Gamma\to\Gamma/M$, let $(\bar V_i,\bar k_i)$ be a cube in $\Gamma/M$, put $V_i=\pi^{-1}(\bar V_i)$, and choose $k_i\in\pi^{-1}(\bar k_i)$.
Then $k_i\notin V_i$, and $\pi^{-1}(\bar k_i^{\varepsilon}\bar V_i)=k_i^{\varepsilon}V_i$, because $g\in\pi^{-1}(\bar k\bar V)$ if and only if $\pi(k^{-1}g)\in\bar V$.
Preimages commute with intersections, and the preimage of a nonempty set under a surjection is nonempty.

(iii) Let $(V_i,k_i)_{i\le t}$ be a cube in $Q$ with $t=N(Q)$, and fix $k_0\in M\setminus\{1\}$.
Set $V_0=Q$, and $V_i'=MV_i$ with $k_i'=k_i$ for $i\ge1$; each $MV_i$ is a subgroup because $M$ is normal.
First, $k_0\notin Q$ because $M\cap Q=1$.
Second, by Dedekind's rule, $Q\cap MV_i=(Q\cap M)V_i=V_i$: if $mv\in Q$ with $v\in V_i\subseteq Q$, then $m=(mv)v^{-1}\in Q\cap M$.
Hence $k_i\in Q\setminus V_i$ gives $k_i\notin MV_i$.
Third, given a pattern $(\varepsilon_0,\varepsilon)$, pick $g\in\bigcap_{i\ge1}k_i^{\varepsilon_i}V_i\subseteq Q$ (for $t=0$ take $g=1$) and put $w=k_0^{\varepsilon_0}g$.
Then $w\in k_0^{\varepsilon_0}Q$.
For $i\ge1$, the set $k_i^{\varepsilon_i}MV_i=Mk_i^{\varepsilon_i}V_i$ contains $g$ and is closed under left multiplication by elements of $M$, so it contains $w$.

(iv) By (ii) it suffices to show, for every finite group $\Delta$ and every chief series $1=\Delta_0<\Delta_1<\dots<\Delta_r=\Delta$ of $\Delta$, that $N(\Delta)\ge c$, where $c$ is the number of complemented factors of the series; we induct on $|\Delta|$.
For $|\Delta|=1$ both sides are $0$.
Otherwise let $M=\Delta_1$, a minimal normal subgroup, and $\bar\Delta=\Delta/M$.
By the correspondence theorem, $(\Delta_{j+1}/M)_{0\le j<r}$ is a chief series of $\bar\Delta$ whose factors correspond to $\Delta_{j+1}/\Delta_j$.
For $i\ge2$, the factor $\Delta_i/\Delta_{i-1}$ is complemented in $\Delta$ if and only if its image is complemented in $\bar\Delta$.
Indeed, the subgroups $C\supseteq\Delta_{i-1}\supseteq M$ correspond to $C/M$, with $(C/M)(\Delta_i/M)=C\Delta_i/M$ and $(C/M)\cap(\Delta_i/M)=(C\cap\Delta_i)/M$.
So the induced series of $\bar\Delta$ has $\bar c=c-[\Delta_1\text{ complemented}]$ complemented factors, where $[\phi]=1$ if the statement $\phi$ holds and $[\phi]=0$ otherwise, and $N(\bar\Delta)\ge\bar c$ by induction.
If $M$ has a complement $C$, then $C\cong\Delta/M$, and (iii) gives $N(\Delta)\ge N(C)+1=N(\bar\Delta)+1\ge c$.
Otherwise (ii) gives $N(\Delta)\ge N(\bar\Delta)\ge\bar c=c$.
Combined with (i) and (ii), this proves the last statement.
\end{proof}

\subsection{Solvable groups and the groups of the quintic}

\ThmSolvable*

\begin{proof}
(a) By \cref{thm:natarajan}(i), $\dimN(\Hc)\le\dimDS(\Hc)=\hgt_L(\Gamma)$, and $\hgt_L(\Gamma)=\mu(\Gamma)$ by \cref{thm:galois}(iii).
For the lower bound, let $S\subseteq\Gamma$ be independent with $|S|=\mu(\Gamma)$, and let $\Delta=\langle S\rangle$.
No proper subset $T\subsetneq S$ generates $\Delta$: for $s\in S\setminus T$ we have $\langle T\rangle\le\langle S\setminus\{s\}\rangle\not\ni s$.
So $S$ is a minimal generating set of $\Delta$.
The group $\Delta$ is solvable, so all its chief factors are abelian, and the following lemma (the inequality half of \citealp[Theorem~2]{lucchini2013largest}) gives $\mu(\Gamma)=|S|\le c$ for the number $c$ of complemented factors of any chief series of $\Delta$.
Then \cref{thm:cosetcube}(iv) gives $\dimN(\Hc)\ge c\ge\mu(\Gamma)$.

\begin{lemma}
\label{lem:mingen}
Let $G$ be a finite group with a chief series $1=G_0<G_1<\dots<G_r=G$ whose factors are all abelian, and let $c$ be its number of complemented factors.
Then every minimal generating set $T$ of $G$ has $|T|\le c$.
\end{lemma}

\begin{proof}
Induct on $|G|$; for $|G|=1$, $T=\emptyset$.
Let $N=G_1$, an abelian minimal normal subgroup, and $\bar G=G/N$ with the chief series $(G_{j+1}/N)_j$.
For $i\ge2$, the factor $G_i/G_{i-1}$ is complemented in $G$ if and only if its image is complemented in $\bar G$, through $C\leftrightarrow C/N$.
So the series of $\bar G$ has $\bar c=c-[N\text{ complemented}]$ complemented factors.
Choose $T'\subseteq T$ minimal with $\langle T'\rangle N=G$.
The images of $T'$ are distinct and form a minimal generating set of $\bar G$, so $|T'|\le\bar c$ by induction.
For $z\in T\setminus T'$, put $M_z=\langle T\setminus\{z\}\rangle$.
Then $M_zN=G$ and $M_z\ne G$.
The intersection $M_z\cap N$ is normalized by $M_z$ and centralized by the abelian $N$, so it is normal in $G$; it is not $N$, hence $M_z\cap N=1$, and $M_z$ complements $N$.
Suppose $x\ne y$ both lie in $T\setminus T'$, and let $K=\langle T\setminus\{x,y\}\rangle$.
Then $KN=G$ and $K\cap N\le M_x\cap N=1$, so $|K|=|M_x|$.
Hence $K=M_x\ni y$, so $T\setminus\{y\}$ generates $G$, contradicting minimality.
So $|T\setminus T'|\le1$, with equality only if $N$ is complemented, and $|T|\le\bar c+[N\text{ complemented}]=c$.
\end{proof}

(c) The upper bound is \cref{cor:galois-values}(a) together with $\dimN\le\dimDS$ (for $n=1$ both dimensions are~$0$).
For the lower bound, first $N(\Gamma_1\times\Gamma_2)\ge N(\Gamma_1)+N(\Gamma_2)$.
Indeed, cubes $(V_i,k_i)$ in $\Gamma_1$ and $(U_j,l_j)$ in $\Gamma_2$ give the cube consisting of $(V_i\times\Gamma_2,(k_i,1))$ and $(\Gamma_1\times U_j,(1,l_j))$, whose intersections are products of nonempty sets.
We have $N(S_2)=1$.
By \cref{thm:cosetcube}(iv), $N(S_3)\ge2$, using the series $1<A_3<S_3$, where $A_3$ is complemented by a transposition.
Likewise $N(S_4)\ge3$, using the series $1<V_4<A_4<S_4$: $V_4$ is complemented by a point stabilizer $S_3$, and $A_4/V_4$ by $D_4$, since $D_4\cap A_4=V_4$ and $|D_4||A_4|/|V_4|=24$.
In both series the top factor is complemented by the previous term.
Write $n=4a+\rho$ with $0\le\rho\le3$.
Disjoint blocks give a subgroup $S_4^a\times S_3^{[\rho=3]}\times S_2^{[\rho=2]}\le S_n$.
So $N(S_n)\ge3a+[\rho=2]+2[\rho=3]=\lfloor3n/4\rfloor$, and \cref{thm:cosetcube}(i) converts this to $\dimN$.

(b) Here $\mu(A_5)\ge3$: the elements $u_1=(2\,3\,4)$, $u_2=(1\,2)(3\,4)$ and $u_3=(0\,1)(3\,4)$ of $A_5$ on $\{0,\dots,4\}$ are independent.
Indeed, $\langle u_2,u_3\rangle$ preserves the partition $\{0,1,2\}\mid\{3,4\}$ but $u_1$ does not; $\langle u_1,u_3\rangle$ preserves $\{0,1\}\mid\{2,3,4\}$ but $u_2$ does not; and $\langle u_1,u_2\rangle$ fixes $0$ but $u_3$ does not.
For $S_5$, the transpositions $(1\,2),(2\,3),(3\,4),(4\,5)$ are independent, and $\dimDS=4$ by \cref{cor:galois-values}(a).
Next, $N(A_5)\ge2$.
Take $V_0=\langle(0\,1\,2)\rangle$ with $k_0=(0\,1)(2\,3)$, and $V_1=\{1,(0\,1)(2\,3),(0\,2)(1\,3),(0\,3)(1\,2)\}$ with $k_1=(0\,1\,2)$.
The four patterns contain $1$, $k_0$, $k_1$ and $k_0k_1$ respectively; for the last, $k_1^{-1}k_0k_1\in V_1$ because $V_1$ is normal in the stabilizer $A_4$ of the point $4$.
Also $N(S_5)\ge3$: in $S_5$ acting on $\{0,\dots,4\}$, the subgroups $V_1=\langle(1\,2),(3\,4)\rangle$, $V_2=\langle(0\,1\,2),(3\,4)\rangle$ and $V_3=\mathrm{Sym}\{0,1,2\}$, with $k_1=(0\,1)$, $k_2=(1\,2)$ and $k_3=(3\,4)$, form a coset cube; this is also part~(c).
The upper bounds $N(A_5)\le2$, $\mu(A_5)\le3$ and $N(S_5)\le3$ are \cref{lem:quintic} below.
By (i) of \cref{thm:cosetcube} and \cref{thm:galois}(iii), this gives $\dimN=2<3=\dimDS$ for $A_5$ and $\dimN=3<4=\dimDS$ for $S_5$.
\end{proof}

\begin{lemma}
\label{lem:quintic}
$N(A_5)\le2$, $\mu(A_5)\le3$ and $N(S_5)\le3$.
\end{lemma}

\begin{proof}
Groups act on the left.
Restricting a cube to some of its coordinates gives a cube, and subsets of independent sets are independent.
So it suffices to exclude a $3$-cube in $A_5$, an independent $4$-set in $A_5$ and a $4$-cube in $S_5$.
For a $t$-cube $(V_i;k_i)$ pick $g_\varepsilon$ in the cell of $\varepsilon$, with $g_0=1$.
We use three facts.
\begin{itemize}
\item[(Q1)] The $V_i$ are pairwise incomparable: if $V_i\le V_j$, the pattern with $\varepsilon_i=0$ and $\varepsilon_j=1$ needs an element of $V_j\cap k_jV_j=\emptyset$.
\item[(Q2)] \emph{Face lemma.} For fixed $i$, the subgroups $W_j=V_i\cap V_j$ ($j\ne i$) with the elements $g_{e_j}\in V_i$ form a $(t-1)$-cube inside $V_i$.
Indeed, $V_i\cap k_jV_j=g_{e_j}(V_i\cap V_j)$, because it contains $g_{e_j}$.
By (Q1) inside $V_i$, the $W_j$ are pairwise incomparable, hence nontrivial and proper when $t-1\ge2$.
\item[(Q3)] Two distinct subgroups $\langle s\rangle,\langle s'\rangle$ of order $2$ carry a $2$-cube only if $ss'=s's$.
Indeed, the cells force $g_{10}=s'$ and $g_{01}=s$, and $g_{11}\in\{s',s's\}\cap\{s,ss'\}$.
In $A_4$, an involution subgroup $\langle a\rangle$ and a subgroup $\langle c\rangle$ of order~$3$ carry no $2$-cube, since the cells would force $a\in N_{A_4}(\langle c\rangle)=\langle c\rangle$, where $N_H(U)$ denotes the normalizer of a subgroup $U$ in a group $H$.
\end{itemize}

\emph{No $3$-cube in $A_5$.}
Each $V_i$ is proper and, by (Q2), contains two incomparable nontrivial proper subgroups, so it is $V_4$, $S_3$, $D_5$ or $A_4$.
Call the $V_i$ the \emph{vertices} of the cube and the subgroups $W_{ij}=V_i\cap V_j$ its \emph{edges}; the edges at each vertex carry a $2$-cube.
Two edges at a vertex are distinct subgroups, and distinct reflections of $S_3$ and $D_5$ do not commute.
So by (Q3) the edge types at a vertex are as follows:
\begin{itemize}
\item $\{C_2,C_2\}$ at $V_4$;
\item $\{C_2,C_3\}$ at $S_3$;
\item $\{C_2,C_5\}$ at $D_5$;
\item $\{C_2,C_2\}$, $\{C_3,C_3\}$ or $\{V_4,C_3\}$ at $A_4$.
\end{itemize}
In $A_5$ there is no $C_5$ edge, since a $C_5$ lies only in $N_{A_5}(C_5)\cong D_5$; so no vertex is $D_5$.
There is no $V_4$ edge, since the subgroups containing a $V_4$ are $V_4$ and $N_{A_5}(V_4)\cong A_4$, which are nested.
No $C_3$ edge joins two $S_3$ vertices, since $N_{A_5}(C_3)$ is the only $S_3$ containing a given $C_3$.
Every $V_4$ is the set $V_4^{(q)}$ of double transpositions fixing a point $q$, together with $1$.
Every $A_4$ is a point stabilizer $\mathrm{Stab}(q)=(A_5)_q$; we also write $\mathrm{Stab}(p,q)=\stab{(A_5)}{\{p,q\}}$ for the pointwise stabilizer of two points.
Since $V_4^{(q)}$ meets every other $V_4$ and every $\mathrm{Stab}(p)$ with $p\ne q$ trivially, a $V_4$ vertex would need two $S_3$ neighbours.
Each of those would then need a $C_3$ edge to the other, which is impossible.
So every vertex is $S_3$ or $\mathrm{Stab}(p)$; let $s$ be the number of $S_3$ vertices.
\begin{itemize}
\item $s=3$ is impossible, since an $S_3$ needs a $C_3$ edge to a vertex that is not an $S_3$.
\item If $s=2$, the $A_4$ vertex $\mathrm{Stab}(p)$ has edges $\mathrm{Stab}(p,q)$ and $\mathrm{Stab}(p,r)$ with $q\ne r$, and the two $S_3$ vertices are their normalizers.
Every involution of $N_{A_5}(\mathrm{Stab}(p,q))$ contains the transposition $(p\,q)$, and every involution of $N_{A_5}(\mathrm{Stab}(p,r))$ contains $(p\,r)$.
So the common edge of the two $S_3$ vertices contains no involution, contradicting its type $C_2$.
\item If $s=1$, two point stabilizers meet in a $C_3$.
This forces both of their edges to the $S_3$ vertex to be $C_3$, again impossible.
\item If $s=0$, then $V_i=\mathrm{Stab}(p_i)$; put $q_i=k_ip_i$.
The pattern $\varepsilon$ requires $g_\varepsilon(p_i)\in\{p_i,q_i\}$ according to $\varepsilon_i$.
Injectivity of $g_\varepsilon$ forces the pairs $\{p_i,q_i\}$ to be pairwise disjoint, which needs $6$ points.
\end{itemize}

\emph{No independent $4$-set in $A_5$.}
An independent $k$-set gives a strict chain of subgroups of length $k$.
A chain in $H$ has length at most $\omega^{+}(|H|)$, the number of prime factors of $|H|$ counted with multiplicity (\cref{sec:fields}).
Suppose $S\subseteq A_5$ is independent with $|S|=4$.
Then each $3$-subset generates a proper subgroup with a chain of length $3$, which must be an $A_4$.
Every such chain in $A_4$ passes through its $V_4$, since $A_4$ has no subgroup of order~$6$.
So every pair from the $3$-subset generates $V_4$, which puts the whole $3$-subset inside $V_4$, a contradiction.

\emph{No $4$-cube in $S_5$.}
By (Q2), each $V_i$ carries a $3$-cube.
So $|V_i|\ge8$, and $V_i$ has three pairwise incomparable proper subgroups, each carrying a $2$-cube.
We list the proper subgroups $H$ of $S_5$ with $|H|\ge8$.
\begin{itemize}
\item If $H$ is intransitive with orbits $4+1$, it lies in a point stabilizer and is $D_4$, $A_4$ or $S_4$.
\item If $H$ is intransitive with orbits $3+2$, it is the stabilizer $S_3\times S_2$ of a $2$-set.
\item Finer orbit partitions give $|H|\le6$.
\item If $H$ is transitive, then $5$ divides $|H|$. A normal Sylow $5$-subgroup puts $H$ inside its normalizer, the Frobenius group $F_{20}=\mathrm{AGL}(1,5)$ of order $20$, so $H$ is $D_5$ or $F_{20}$. Otherwise $30$ divides $|H|$, and $H=A_5$, because $A_5$ has no subgroup of order $15$ or $30$.
\end{itemize}
Of these, $D_4$ has only its two Klein subgroups carrying a $2$-cube, $D_5$ has none, $A_4$ has one and $F_{20}$ has one.
Also $N(A_5)\le2$.
So each $V_i$ is the stabilizer of an object $x_i$, which is a point or a $2$-set.
Put $y_i=k_ix_i\ne x_i$.
The pattern $\varepsilon$ requires $g_\varepsilon(x_i)=x_i$ or $y_i$ according to $\varepsilon_i$.
Comparing the patterns $(1,0)$, $(0,1)$ and $(1,1)$ on two coordinates gives three rules.
\begin{itemize}
\item[(S1)] Two point objects have disjoint pairs $\{x_i,y_i\}$ and $\{x_j,y_j\}$.
\item[(S2)] A point object lies outside every $2$-set object. Moreover, the point and its image both avoid the $2$-set and its image.
\item[(S3)] Two $2$-set objects meet in exactly one point. If they were disjoint, the fifth point would lie in both images.
\end{itemize}
Let $\nu$ be the number of point objects among the four.
\begin{itemize}
\item $\nu\ge3$ needs $6$ points.
\item $\nu=2$ leaves a single point for any $2$-set object.
\item $\nu=1$ confines two $2$-set objects and their two images, four distinct $2$-sets, to a $3$-set.
\item $\nu=0$ gives four pairwise-intersecting $2$-sets, which form a star $\{v,a_1\},\dots,\{v,a_4\}$.
The element realizing the first unit pattern fixes the other three sets setwise.
So it fixes $v,a_2,a_3,a_4$ and hence $a_1$, so it is the identity, contradicting $y_1\ne x_1$.
\end{itemize}
\end{proof}

All three bounds were also confirmed by exhaustive search with several implementations that share no code.
They agree on all intermediate counts:
\begin{itemize}
\item $59$ subgroups of $A_5$;
\item $1{,}240$ unordered independent triples (independent $3$-element subsets) of $A_5$, and no independent $4$-set;
\item $8{,}451$ unordered pairs of subgroups of $S_5$ carrying a $2$-cube;
\item $10{,}200$ unordered coset-level $3$-cubes in $S_5$ (sets of three pairs $(V_i,k_iV_i)$ forming a coset cube), none of which extends.
\end{itemize}

\subsection{Beyond chief factors: \texorpdfstring{$\mathrm{P\Gamma L}(2,8)$}{PGammaL(2,8)}}

\PropPGaL*

\begin{proof}
The group $K=\mathrm{PSL}(2,8)=\mathrm{PGL}(2,8)$ acts sharply $3$-transitively on the $9$ points of the projective line over $\mathbb F_8$ (\cref{ext:pgl}), and $\Gamma=K\rtimes\langle\varphi\rangle$ acts $3$-transitively.
Choose six distinct points $p_1,p_2,p_3,q_1,q_2,q_3$.
For $i\le3$ let $V_i=\Gamma_{p_i}$, and choose $k_i\in K$ with $k_ip_i=q_i$.
Let $V_4=K$ and $k_4=\varphi$.
Fix $\varepsilon\in\{0,1\}^4$, and let $r_i$ be $p_i$ or $q_i$ according to $\varepsilon_i$, for $i\le3$; the $r_i$ are distinct.
An element $g$ lies in $k_i^{\varepsilon_i}V_i$ if and only if $gp_i=r_i$.
By sharp $3$-transitivity there is $\kappa\in K$ with $\kappa p_i=r_i$ for $i\le3$.
The pointwise stabilizer $P$ of $p_1,p_2,p_3$ in $\Gamma$ has order $1512/(9\cdot8\cdot7)=3$ and meets $K$ trivially, so $P$ is a complement of $K$.
Hence the coset $\kappa P$, which consists of the elements $g$ with $gp_i=r_i$ for $i\le3$, meets every coset of $K$, in particular $\varphi^{\varepsilon_4}K$.
An element of the intersection lies in all four sets $k_i^{\varepsilon_i}V_i$, so all $16$ cells are nonempty.
The bound on complemented chief factors was obtained by exhaustive computation over all $1154$ subgroups of $\Gamma$, with independent implementations.
\end{proof}

\subsection{Saturation without solvability: \texorpdfstring{$\mathrm{PSL}(2,8)$}{PSL(2,8)}}

\PropPSL*

\begin{proof}
The translations $z\mapsto z+b$, $b\in\mathbb F_8$, form a subgroup $E\cong C_2^3$ of $\mathrm{PSL}(2,8)$.
A basis $t_1,t_2,t_3$ of $E$ is an independent set, so $\mu\ge3$.
The chief series $1<\langle t_1\rangle<\langle t_1,t_2\rangle<E$ has three complemented factors, so $N\ge N(E)\ge3$ by \cref{thm:cosetcube}(ii) and~(iv).
Moreover $N\le\mu$ for every finite group: given a coset cube $(V_i,k_i)_{i\le t}$, pick $g_i$ in the cell of the $i$-th unit pattern; then $g_i\notin V_i$ while $g_j\in V_i$ for $j\neq i$, so $\langle g_j:j\neq i\rangle\le V_i\not\ni g_i$, the $g_i$ are distinct, and $\{g_1,\dots,g_t\}$ is independent.
Finally $\mu\le3$; we argue by hand, using Dickson's list of maximal subgroups and sharp $3$-transitivity (an exhaustive computer search, with several independent programs, confirms the value).
(For comparison, \citet{whiston2002psl} bound the size of independent generating sets of $\mathrm{PSL}_2(q)$; our $\mu$ counts all independent subsets.)
Write $T=\mathrm{PSL}(2,8)$, acting on the projective line $\mathbb F_8\cup\{\infty\}$.
Since $8$ is even, every element of $\mathbb F_8^\times$ is a square, so $T$, the group of M\"obius maps $z\mapsto(az+b)/(cz+d)$ with $ad-bc$ a nonzero square, is all of $\mathrm{PGL}(2,8)$ \citep[Section~2]{cameron2006designs} \citep[see also][p.~3]{leemans2009pgl}.
It acts $3$-transitively on the $9$ points and has order $504=9\cdot8\cdot7$ \citep[Section~2]{cameron2006designs}, so the action is sharply $3$-transitive, and a nonidentity element fixes at most two points.
Every maximal subgroup of $T$ is isomorphic to $C_2^3\rtimes C_7$, $D_7$ or $D_9$, of orders $56$, $14$ and $18$ \citep[Theorem~2.1]{giudici2007maximal}.
Independence of a set does not depend on the ambient group, and subsets of independent sets are independent.
\emph{Dihedral groups.} In $D_m$ with $m\in\{7,9\}$, two rotations are dependent, since the subgroups of a cyclic group of prime-power order form a chain; two reflections $u,v$ generate $\langle uv,u\rangle$, which is all of $D_m$ unless $m=9$ and $uv$ has order $3$; in that case a third element outside $\langle uv,u\rangle$ generates $D_m$ together with $u$. So no $3$-subset is independent, and $\mu(D_7)=\mu(D_9)=2$.
\emph{Point stabilizers.} $T_\infty=\{z\mapsto az+b\}$, with the normal subgroup $E$ of translations, a Sylow $2$-subgroup of $T$ (as $|T|=504=8\cdot63$).
A nonidentity translation fixes only $\infty$, so an element $g$ that normalizes $E$ satisfies $g\infty=\infty$ (the nonidentity elements of $gEg^{-1}$ fix $g\infty$); since $E$ is normal in $T_\infty$, the normalizer of $E$ is $T_\infty$.
A subgroup of $T$ of order $56$ isomorphic to $C_2^3\rtimes C_7$ normalizes its normal subgroup of order $8$, a conjugate $gEg^{-1}$, so it lies in $gT_\infty g^{-1}=T_{g\infty}$ and, having the same order, equals it: it is a point stabilizer.
An element $x\colon z\mapsto az+b$ with $a\neq1$ has order $7$, and conjugation by $x$ multiplies translations by $a$; since $\mathbb F_2[a]=\mathbb F_8$, the only $x$-invariant subgroups of $E$ are $1$ and $E$.
Hence, if a subgroup $K\le T_\infty$ is not contained in $E$, it contains such an $x$, and $K\cap E$, being $x$-invariant, is $1$ or $E$. If $K\cap E=E$, then $K\supseteq\langle E,x\rangle=T_\infty$, since $8$ and $7$ divide $|\langle E,x\rangle|$. If $K\cap E=1$, then $|K|=|KE|/|E|\le7$, so $K=\langle x\rangle$. So every proper subgroup of $T_\infty$ lies in $E$ or is cyclic of order $7$.
If $S\subseteq T_\infty$ were independent with $|S|=4$, then $S\not\subseteq E$ (four vectors in $\mathbb F_2^3$ are dependent); for $s_1\in S\setminus E$ and two other elements $s_2,s_3$, the proper subgroups $\langle S\setminus\{s_2\}\rangle$ and $\langle S\setminus\{s_3\}\rangle$ contain $s_1$, so both equal $\langle s_1\rangle$, and $s_2\in\langle s_1\rangle\subseteq\langle S\setminus\{s_2\}\rangle$, a contradiction. So $\mu(T_\infty)=3$, and hence $\mu(T_p)=3$ for every point $p$, as all point stabilizers are conjugate.
\emph{The group $T$.} Let $S=\{s_1,\dots,s_4\}\subseteq T$ be independent and $M_i=\langle S\setminus\{s_i\}\rangle$.
Each $M_i$ is proper, so it lies in a maximal subgroup of $T$; that subgroup contains the independent $3$-set $S\setminus\{s_i\}$, so by the dihedral paragraph it is not isomorphic to $D_7$ or $D_9$, so it has order $56$ and is a point stabilizer $T_{p_i}$, by the paragraph on point stabilizers.
If the $p_i$ are pairwise distinct, each $s_k$ lies in the three groups $M_i$ with $i\neq k$ and fixes three points, so $s_k=1$, which no independent set contains.
If $p_i=p_j$ for some $i\neq j$, then $S\subseteq M_i\cup M_j\subseteq T_{p_i}$, contradicting $\mu(T_{p_i})=3$.
The statement about $\HG{\Gamma}{L}$ follows from \cref{thm:cosetcube}(i) and \cref{thm:galois}(iii).
\end{proof}

\subsection{The Abel--Ruffini threshold}

\ThmAbelRuffini*

Permutations compose as functions, $(gh)(x)=g(h(x))$, and cosets are left cosets, as in the definition of coset cubes.
We write $[n]=\{1,\dots,n\}$, $\mathbf 1_A\in\Z^n$ for the indicator vector of $A\subseteq[n]$, and $\langle\cdot,\cdot\rangle$ for the standard inner product, so that $\langle\mathbf 1_A,\mathbf 1_B\rangle=|A\cap B|$.
Call a family $(V_i,k_i)_{i\le t}$ of subgroups $V_i$ of a group and elements $k_i\notin V_i$ \emph{$2$-full} if every cell $C_\varepsilon=\bigcap_ik_i^{\varepsilon_i}V_i$ with at most two coordinates of $\varepsilon$ equal to~$1$ is nonempty.
Let $e_i$ be the $i$-th unit pattern.

\begin{lemma}[Unit cells]
\label{lem:unitcells}
Let $(V_i,k_i)_{i\le t}$ be $2$-full, and pick $g_i\in C_{e_i}$ for each~$i$.
Then $k_iV_i=g_iV_i$, $g_i\notin V_i$ and $\langle g_j:j\neq i\rangle\le V_i$; in particular $\{g_1,\dots,g_t\}$ is an independent set of size~$t$.
\end{lemma}

\begin{proof}
As $g_i\in k_iV_i$, we have $k_iV_i=g_iV_i$, and $g_i\in V_i$ would give $k_i\in V_i$.
For $j\neq i$, $g_j\in C_{e_j}\subseteq V_i$.
So $g_i\notin\langle g_j:j\neq i\rangle$, and the $g_i$ are distinct.
\end{proof}

\begin{lemma}[Orthogonality]
\label{lem:orthogonality}
Let $(V_i,k_i)_{i\le t}$ be a $2$-full family in $S_n$, and let $g_i$ be as in \cref{lem:unitcells}.
Let $i\neq j$, and let $A,A'\subseteq[n]$ satisfy $V_iA=A$ and $V_jA'=A'$.
Then $\langle\mathbf 1_{g_iA}-\mathbf 1_A,\ \mathbf 1_{g_jA'}-\mathbf 1_{A'}\rangle=0$.
\end{lemma}

\begin{proof}
For $\alpha,\beta\in\{0,1\}$ pick $h\in C_{\alpha e_i+\beta e_j}$.
Then $h\in k_i^\alpha V_i=g_i^\alpha V_i$, so $hA=g_i^\alpha A$; likewise $hA'=g_j^\beta A'$.
Hence $|g_i^\alpha A\cap g_j^\beta A'|=|h(A\cap A')|=|A\cap A'|$ does not depend on $(\alpha,\beta)$.
Expanding, the inner product is $|g_iA\cap g_jA'|-|g_iA\cap A'|-|A\cap g_jA'|+|A\cap A'|=0$.
\end{proof}

\begin{lemma}[Overgroups of Young subgroups]
\label{lem:young}
Let $[n]=A\sqcup B$ with $|A|>|B|\ge1$, and let $V\le S_n$ contain $\Sym(A)\times\Sym(B)$.
If $V\neq S_n$, then $VA=A$.
\end{lemma}

\begin{proof}
Recall that the transpositions along the edges of a graph on $[n]$ generate a group containing every transposition inside a connected component, since $(x\,y)(y\,z)(x\,y)=(x\,z)$.
Suppose $v\in V$ and $a\in A$ have $b=v(a)\in B$.
Since $|A|+|v(A)|>n$, some $x$ lies in $A\cap v(A)$; then $x\neq b$ and $(x\,b)\in\Sym(v(A))=v\,\Sym(A)\,v^{-1}\le V$.
So $V$ contains all transpositions inside $A$, all inside $B$, and $(x\,b)$, which connect $[n]$; hence $V=S_n$.
\end{proof}

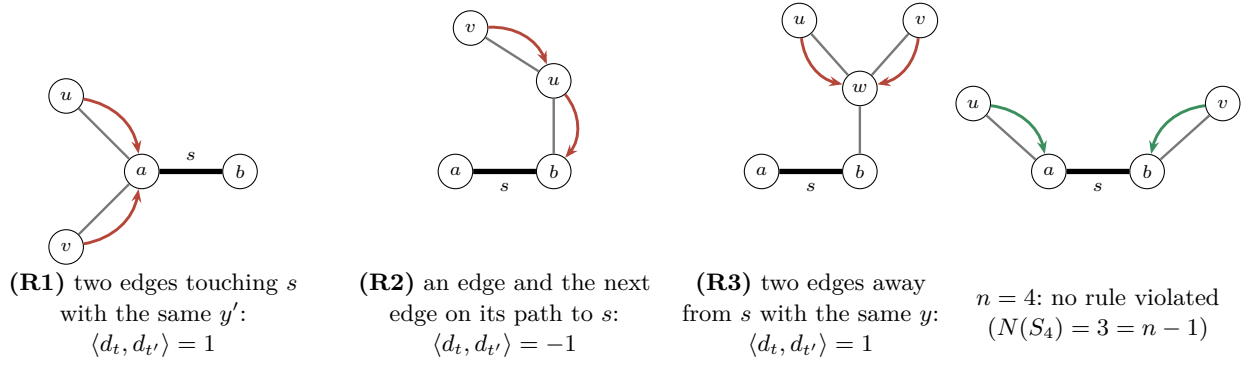
\begin{figure}[t]
\centering
\begin{tikzpicture}[font=\small,
  v/.style={circle, draw, fill=white, inner sep=1.2pt, minimum size=13pt, font=\scriptsize},
  sedge/.style={line width=2pt},
  tedge/.style={FigGray, line width=0.9pt},
  d/.style={-{Stealth[length=5pt]}, line width=1.1pt, bend left=35}]
\begin{scope}
\node[v] (a) at (0,0) {$a$}; \node[v] (b) at (1.3,0) {$b$};
\node[v] (x) at (-1.0,1.0) {$u$}; \node[v] (z) at (-1.0,-1.0) {$v$};
\draw[sedge] (a) -- node[above,font=\scriptsize]{$s$} (b);
\draw[tedge] (x) -- (a); \draw[tedge] (z) -- (a);
\draw[d,FigRed] (x) to (a); \draw[d,FigRed,bend right=35] (z) to (a);
\node[align=center] at (0.15,-1.9) {\textbf{(R1)} two edges touching $s$\\ with the same $y'$:\\ $\langle d_t,d_{t'}\rangle=1$};
\end{scope}
\begin{scope}[shift={(4.15,0)}]
\node[v] (a) at (0,0) {$a$}; \node[v] (b) at (1.3,0) {$b$};
\node[v] (x) at (1.3,1.2) {$u$}; \node[v] (z) at (0.2,1.9) {$v$};
\draw[sedge] (a) -- node[below,font=\scriptsize]{$s$} (b);
\draw[tedge] (b) -- (x); \draw[tedge] (x) -- (z);
\draw[d,FigRed,bend left=40] (x) to (b);
\draw[d,FigRed] (z) to (x);
\node[align=center] at (0.65,-1.9) {\textbf{(R2)} an edge and the next\\ edge on its path to $s$:\\ $\langle d_t,d_{t'}\rangle=-1$};
\end{scope}
\begin{scope}[shift={(8.2,0)}]
\node[v] (a) at (0,0) {$a$}; \node[v] (b) at (1.3,0) {$b$};
\node[v] (w) at (1.3,1.1) {$w$}; \node[v] (x) at (0.5,2.0) {$u$}; \node[v] (z) at (2.1,2.0) {$v$};
\draw[sedge] (a) -- node[below,font=\scriptsize]{$s$} (b);
\draw[tedge] (b) -- (w); \draw[tedge] (w) -- (x); \draw[tedge] (w) -- (z);
\draw[d,FigRed,bend right=35] (x) to (w); \draw[d,FigRed] (z) to (w);
\node[align=center] at (0.65,-1.9) {\textbf{(R3)} two edges away\\ from $s$ with the same $y$:\\ $\langle d_t,d_{t'}\rangle=1$};
\end{scope}
\begin{scope}[shift={(12.0,0)}]
\node[v] (x) at (-1.0,0.9) {$u$}; \node[v] (a) at (0,0) {$a$}; \node[v] (b) at (1.3,0) {$b$}; \node[v] (z) at (2.3,0.9) {$v$};
\draw[sedge] (a) -- node[below,font=\scriptsize]{$s$} (b);
\draw[tedge] (x) -- (a); \draw[tedge] (b) -- (z);
\draw[d,FigGreen] (x) to (a); \draw[d,FigGreen,bend right=35] (z) to (b);
\node[align=center] at (0.65,-1.9) {$n=4$: no rule violated\\ ($N(S_4)=3=n-1$)};
\end{scope}
\end{tikzpicture}
\caption{The orthogonality rules (R1)--(R3) of the proof of \cref{thm:abelruffini}(a), drawn in the notation of that proof.
The thick edge is $s=(a\,b)$ and the thin edges are other edges of the tree~$T$.
An arrow from $x_t$ to $y'_t$ depicts the vector $d_t=\mathbf 1_{y'_t}-\mathbf 1_{x_t}$ of an unbalanced edge $t\neq s$, and \cref{lem:orthogonality} requires $\langle d_t,d_{t'}\rangle=0$ for distinct such edges.
The three configurations in red violate this.
For $n\ge5$ no tree survives: by (R2) every edge other than $s$ touches~$s$, by (R1) at most two do, and there are $n-2\ge3$ (a balanced edge is excluded separately).
For $n=4$ the path on the right satisfies all three rules, consistent with $N(S_4)=3$.}
\label{fig:orthogonality}
\end{figure}

\begin{proof}[Proof of \cref{thm:abelruffini}]
Let $n\ge5$ and suppose that $(V_i,k_i)_{i\le n-1}$ is a $2$-full family in $S_n$.
By \cref{lem:unitcells}, $S=\{g_1,\dots,g_{n-1}\}$ is an independent subset of $S_n$ of size $n-1$.
By the theorem of \citet{whiston2000maximal}, an independent subset of $S_n$ of size $n-1$ generates $S_n$ (see also \citealp[Section~2]{cameron2002independent}); so $S$ generates $S_n$.
By \citet[Theorem~2.1]{cameron2002independent}, for $n\ge7$, there are a tree $T$ on $[n]$ and an edge $s=(a\,b)$ of~$T$ such that, writing $S(T)$ for the transpositions along the edges of~$T$, either $S=S(T)$ or $S=\{s\}\cup\{(st)^{\pm1}:t\in S(T)\setminus\{s\}\}$.
By the remark following their Corollary~2.4, the same holds for $n=5$, and for $n=6$ it holds after applying an automorphism of $S_6$; an automorphism maps cosets to cosets and preserves intersections and non-membership, so we may apply it to the whole family.
If $t$ is disjoint from $s$ then $st=ts$, and otherwise $(st)^{-1}=ts$.
So in all cases $S=\{s\}\cup\{g_t:t\in S(T)\setminus\{s\}\}$ with $g_t\in\{t,st,ts\}$ (for $S=S(T)$, take any edge as $s$ and $g_t=t$).
The $n-1$ elements of $S$ are distinct (\cref{lem:unitcells}) and $T$ has $n-1$ edges, so $s$ and the $g_t$ ($t\neq s$) are distinct and exhaust~$S$.
We index the family by the edges of~$T$, so that $g_s=s$ and $g_t$ is the element chosen from the unit cell of the coordinate indexed by~$t$.

For an edge $t\neq s$, the endpoints $a,b$ of $s$ lie in one component $B_t$ of $T-t$; let $A_t$ be the other component, $x_t\in A_t$ the endpoint of $t$ in $A_t$, and $y_t\in B_t$ its other endpoint.
Every generator in $S\setminus\{g_t\}$ preserves $A_t$ and $B_t$, and the group they generate contains $s$, hence every $t'\in S(T)\setminus\{t\}$; so $\langle S\setminus\{g_t\}\rangle=\Sym(A_t)\times\Sym(B_t)$.
By \cref{lem:unitcells} this group lies in $V_t$, and $V_t\neq S_n$.
Call $t$ \emph{balanced} if $|A_t|=|B_t|$.
If $t$ is unbalanced, \cref{lem:young}, applied to the larger of $A_t,B_t$, gives $V_tA_t=A_t$.
Since $s$ fixes $A_t$ pointwise and $t$ fixes $A_t\setminus\{x_t\}$ pointwise, $g_tA_t=(A_t\setminus\{x_t\})\cup\{y'_t\}$ with $y'_t=g_t(x_t)\in B_t$.
If $y_t\notin\{a,b\}$ then $y'_t=y_t$, and if $y_t\in\{a,b\}$ (we say that $t$ \emph{touches}~$s$) then $y'_t\in\{a,b\}$.
Put $d_t=\mathbf 1_{y'_t}-\mathbf 1_{x_t}$.
By \cref{lem:orthogonality}, $\langle d_t,d_{t'}\rangle=0$ for distinct unbalanced edges $t,t'\neq s$.
This has three consequences (\cref{fig:orthogonality}).
\begin{itemize}
\item[(R1)] Let $t,t'$ be distinct unbalanced edges touching~$s$. Their endpoints $x_t,x_{t'}$ are distinct (otherwise $T$ would contain a cycle through~$s$) and lie outside $\{a,b\}$, so $\langle d_t,d_{t'}\rangle=[y'_t=y'_{t'}]$ and hence $y'_t\neq y'_{t'}$. As $y'_t\in\{a,b\}$, at most two unbalanced edges touch~$s$.
\item[(R2)] If $t'$ does not touch $s$, let $t$ be the first edge on the path from $y_{t'}$ to~$a$; then $t\neq s$, $x_t=y_{t'}$ and $A_{t'}\subseteq A_t$. Now $d_{t'}=\mathbf 1_{x_t}-\mathbf 1_{x_{t'}}$, and $y'_t\in B_t$ differs from $x_t$ and from $x_{t'}\in A_t$, while $x_t=y_{t'}\neq x_{t'}$, so $\langle d_t,d_{t'}\rangle=-1$. Hence $t$ and $t'$ are not both unbalanced.
\item[(R3)] Two distinct edges $t,t'\neq s$ that do not touch $s$ and have $y_t=y_{t'}=w$ have $\langle d_t,d_{t'}\rangle=\langle\mathbf 1_w-\mathbf 1_{x_t},\mathbf 1_w-\mathbf 1_{x_{t'}}\rangle=1$, so they are not both unbalanced.
\end{itemize}
A tree has at most one balanced edge: for two balanced edges $e,f$, the side of $e$ away from $f$ and the side of $f$ away from $e$ are disjoint, have $n/2$ points each, and miss an endpoint of~$e$.
Suppose an edge $\beta\neq s$ is balanced.
Then every other edge is unbalanced, so by (R2) every edge inside $A_\beta$ has $x_\beta$ as its endpoint nearer to~$s$ (such an edge does not touch $s$, and the first edge on its path to $a$ is $\beta$ or another, unbalanced, edge inside $A_\beta$), and by (R3) there is at most one such edge.
As $A_\beta$ spans a subtree, $|A_\beta|\le2$, so $n=2|A_\beta|\le4$, a contradiction.
So every edge other than $s$ is unbalanced, by (R2) every such edge touches~$s$, and by (R1) there are at most two of them.
But $T$ has $n-2\ge3$ edges other than~$s$.
This contradiction proves the first statement of~(a).
If $N(S_n)\ge n-1$, the first $n-1$ members of a cube of dimension at least $n-1$ would form a family whose cells contain cells of the cube; so $N(S_n)\le n-2$.

For (b), $S_n$ is the Galois group of $\Q(x_1,\dots,x_n)$ over its fixed field, a countable field, by Artin's theorem \citep[Corollary~3.11]{milne2022fields}, so \cref{thm:cosetcube}(i) and \cref{thm:solvable}(c) give $N(S_n)\ge\lfloor3n/4\rfloor$, which equals $n-2$ for $5\le n\le8$.
For (c), $\dimDS(\Hc)=n-1$ by \cref{thm:solvable}(c); if $n\le4$, then $S_n$ is solvable and \cref{thm:solvable}(a) applies; if $n\ge5$, then $\dimN(\Hc)=N(S_n)\le n-2$ by \cref{thm:cosetcube}(i) and~(a).
For $X_0$, use \cref{thm:solvable}(d).
\end{proof}

\subsection{Computations}
\Cref{tab:cosetcube} lists the coset-cube numbers we computed exhaustively.
In every case $N(\Gamma)=\max_{\Delta\le\Gamma}c(\Delta)$, where $c(\Delta)$ is the largest number of complemented factors of a chief series of $\Delta$.

\begin{table}[h]
\centering
\small
\begin{tabular}{lccccc}
\toprule
group & solvable & $\mu$ & $\max_\Delta c(\Delta)$ & $N=\dimN$ & $\dimN=\dimDS$?\\
\midrule
$C_2,\ C_4$ & yes & 1 & 1 & 1 & yes\\
$C_6,\ S_3,\ D_4,\ D_5,\ D_8,\ Q_8,\ A_4,\ \mathrm{SL}(2,3)$ & yes & 2 & 2 & 2 & yes\\
$C_2^3,\ D_6,\ C_3\times S_3,\ S_4,\ D_4\times C_2,\ A_4\times C_2$ & yes & 3 & 3 & 3 & yes\\
$C_2^4,\ S_3\times S_3,\ C_2\times S_4$ & yes & 4 & 4 & 4 & yes\\
$A_5$ & no & 3 & 2 & 2 & no\\
$S_5,\ A_5\times C_2,\ \mathrm{PSL}(2,7)$ & no & 4 & 3 & 3 & no\\
\bottomrule
\end{tabular}
\caption{Coset-cube numbers of small groups, computed by exhaustive search with independent programs; $C_2^2$ (with $\mu=N=2$) is omitted.}
\label{tab:cosetcube}
\end{table}

\section{Proofs for Section~\ref{sec:monodromy}: monodromy of pseudo-cubes}
\label{app:monodromy}

\ThmMonodromy*

\begin{proof}
(a) The word $s_i(b)$ is an $i$-neighbor of $b$, and $b$ is an $i$-neighbor of $s_i(b)$, since the two words differ exactly in coordinate $i$.
By uniqueness of $i$-neighbors, $s_i(s_ib)=b$; and $s_ib\neq b$.
For $i\neq j$, the word $s_ib$ agrees with $b$ in coordinate $j$, so $\pi_j\circ s_i=\pi_j$.
Hence $\pi_j$ is invariant under the group $K_j$ generated by the $s_i$ with $i\neq j$, that is, $\pi_j$ is constant on $K_j$-orbits.
Further, $\pi_j(s_jb)\neq\pi_j(b)$ by the definition of a $j$-neighbor.
If $s_jb\in K_j\cdot b$, then $\pi_j(s_jb)=\pi_j(b)$, a contradiction; so $s_jb\notin K_j\cdot b$, and in particular $s_j\notin K_j$.

(b) If $\iota(b)=\iota(b')$, then $b'\in K_j\cdot b$ for every $j$, so $\pi_j(b')=\pi_j(b)$ for every $j$ by (a), and $b=b'$ because $B\subseteq Y^d$.
The set $B^\sharp=\iota(B)$ is finite and nonempty.
For $j\neq i$ we have $K_j\cdot(s_ib)=K_j\cdot b$, since $s_i\in K_j$; and $K_i\cdot(s_ib)\neq K_i\cdot b$ by separation.
So $\iota(s_ib)$ is an $i$-neighbor of $\iota(b)$, and $B^\sharp$ is a pseudo-cube.
The maps $K_j\cdot b\mapsto\pi_j(b)$ are well defined by (a), and together they send $\iota(b)$ to $b$.

(c) For $m\in M$ write $w(m)=(K_1m,\dots,K_dm)$.
The word $w(s_im)$ agrees with $w(m)$ at every $j\neq i$, because $s_i\in K_j$ gives $K_js_im=K_jm$.
It differs from $w(m)$ at $i$, because $s_im\in K_im$ would mean $s_i\in K_i$, which is false by (a).
The set $C$ is finite, since $M\le\Sym(B)$ is finite, and nonempty; so $C$ is a $d$-dimensional pseudo-cube.
Under $m\mapsto m^{-1}$, right cosets $K_jm$ become left cosets $m^{-1}K_j$, which gives the coset complex $\mathcal{C}_M(K_1,\dots,K_d)$; and $s_i\in\big(\bigcap_{j\neq i}K_j\big)\setminus K_i$ is condition~(1) of \citet[Theorem~45]{brukhim2022characterization}.

The map $K_jm\mapsto K_j\cdot(mb_0)$ is well defined: if $K_jm=K_jm'$, then $m'=km$ with $k\in K_j$, so $m'b_0=k(mb_0)\in K_j\cdot(mb_0)$ and the two orbits coincide.
Applied coordinatewise, these maps send $w(m)$ to $(K_1\cdot(mb_0),\dots,K_d\cdot(mb_0))=\iota(mb_0)$.
Since $M$ is transitive, every $\iota(b)$ is hit, so $C$ is mapped onto $B^\sharp$.

Now let $M$ act regularly, so that $m\mapsto mb_0$ is a bijection from $M$ onto $B$.
Each coordinate map is injective: if $K_j\cdot(mb_0)=K_j\cdot(m'b_0)$, then $m'b_0=kmb_0$ for some $k\in K_j$, so $m'=km$ by regularity, and $K_jm'=K_jm$.
The map on words is injective as well: if $\iota(mb_0)=\iota(m'b_0)$, then $mb_0=m'b_0$ because $\iota$ is injective, so $m=m'$.
Hence the coordinatewise maps are bijections and $C\cong B^\sharp$.

(d) ($\Rightarrow$) Let $M$ be abelian and $b\in B$.
Since $M$ is generated by the commuting involutions $s_i$, we have $M=\{s^\epsilon : \epsilon\in\{0,1\}^d\}$, where $s^\epsilon=\prod_is_i^{\epsilon_i}$.
The map $\epsilon\mapsto s^\epsilon b$ is injective.
Indeed, suppose $s^\epsilon b=s^{\epsilon'}b$ with $\epsilon\neq\epsilon'$, let $\eta=\epsilon+\epsilon'$ modulo $2$, and pick $i$ with $\eta_i=1$.
Then $s^\eta=s^{\epsilon'}s^\epsilon$, so $s^\eta b=b$.
Write $s^\eta=s_ir$ with $r\in K_i$; then $rb=s_ib$, which contradicts separation.
Hence $|Mb|=2^d$.
For $j\in[d]$ write $s^\epsilon=r's_j^{\epsilon_j}$ with $r'\in K_j$; then $\pi_j(s^\epsilon b)=\pi_j(s_j^{\epsilon_j}b)\in\{\pi_j(b),\pi_j(s_jb)\}$, and these two values are distinct by (a).
So the injective map $b'\mapsto(\pi_1(b'),\dots,\pi_d(b'))$ sends $Mb$ into $\prod_j\{\pi_j(b),\pi_j(s_jb)\}$, a set of size $2^d=|Mb|$, and $Mb$ is this product.
Since $B$ is the disjoint union of its $M$-orbits, $B$ is a disjoint union of full Natarajan cubes, and these are the $M$-orbits.
The group $M$ is generated by commuting involutions, so it is an elementary abelian $2$-group.

($\Leftarrow$) Let $B=\bigsqcup_rQ_r$ with $Q_r=\prod_j\{u^r_j,v^r_j\}$ and $u^r_j\neq v^r_j$.
For $b\in Q_r$, the element of $Q_r$ obtained by flipping coordinate $i$ is an $i$-neighbor of $b$ in $B$; since $B$ is two-fibered, it equals $s_ib$.
So each $s_i$ preserves every $Q_r$ and acts on it by flipping coordinate $i$.
These actions commute, so $M$ is abelian, and every element of $M$ has order at most $2$.
The sets $Q_r$ are the $M$-orbits, since $M$ acts transitively on each of them by flips.

(e) Let $M_T=\langle s_i : i\in T\rangle$; it preserves the orbit $O=M_T\cdot b$.
Since the $s_i$ with $i\in T$ commute as maps on $O$, every element of $M_T$ agrees on $O$ with some $s^\epsilon=\prod_{i\in T}s_i^{\epsilon_i}$, $\epsilon\in\{0,1\}^T$.
If $s^\epsilon b=s^{\epsilon'}b$ with $\epsilon\neq\epsilon'$, the argument of (d), evaluated on $O$ and using $K_i\supseteq\langle s_j : j\in T\setminus\{i\}\rangle$, contradicts separation; so $|O|=2^{|T|}$.
For $j\in T$, write $s^\epsilon=r's_j^{\epsilon_j}$ on $O$, where $r'$ is a product of the $s_i$ with $i\in T\setminus\{j\}$; each such $s_i$ preserves $\pi_j$, so $\pi_j(s^\epsilon b)=\pi_j(s_j^{\epsilon_j}b)$.
Hence the projection of $s^\epsilon b$ to the coordinates in $T$ is the element of $\prod_{j\in T}\{\pi_j(b),\pi_j(s_jb)\}$ indexed by $\epsilon$, and the projection of $O$ is this whole product, whose factors have two distinct entries by (a).
So the coordinates in $T$ are N-shattered by $B$, with $f_0(j)=\pi_j(b)$ and $f_1(j)=\pi_j(s_jb)$, and $\dimN(B)\ge|T|$.
\end{proof}

\end{document}